\documentclass[table]{opendatalab}
\usepackage{xcolor}
\usepackage{longtable}
\usepackage{listings}
\usepackage[numbers]{natbib}
\definecolor{codegreen}{rgb}{0,0.6,0}
\definecolor{codegray}{rgb}{0.5,0.5,0.5}
\definecolor{codepurple}{rgb}{0.58,0,0.82}
\definecolor{backcolour}{rgb}{0.95,0.95,0.92}
\definecolor{promptcolor}{HTML}{D1D0F2}
\definecolor{promptcolorheader}{HTML}{bdbcec}
\newcommand{\promptbox}[2]{
\begin{tcolorbox}[
top=0.3em,bottom=0.3em,left=0.5em,right=0.5em,
toptitle=0.3em,bottomtitle=0.2em,boxsep=0pt,
colframe=promptcolorheader,colback=promptcolor!50,boxrule=0.5pt,
]
\footnotesize
\end{tcolorbox}
}
\lstdefinestyle{mystyle}{
    backgroundcolor=\color{backcolour},   
    commentstyle=\color{codegreen},
    keywordstyle=\color{magenta},
    numberstyle=\tiny\color{codegray},
    stringstyle=\color{codepurple},
    basicstyle=\ttfamily\footnotesize,
    breakatwhitespace=false,         
    breaklines=true,                 
    captionpos=b,                    
    keepspaces=true,                 
    numbers=left,                    
    numbersep=5pt,                  
    showspaces=false,                
    showstringspaces=false,
    showtabs=false,                  
    tabsize=2
}

\usepackage[utf8]{inputenc} 
\usepackage[T1]{fontenc}    
\usepackage{hyperref}       
\usepackage{url}            
\usepackage{booktabs}       
\usepackage{amsfonts}       
\usepackage{nicefrac}       
\usepackage{microtype}      

\usepackage{amsmath}
\newcommand{\sysname}{{REST}}

\usepackage{hyperref}
\usepackage{url}
\usepackage{booktabs}
\usepackage{multirow} 
\usepackage{subcaption}
\usepackage{arydshln}
\usepackage{todonotes}
\usepackage{utfsym}
\usepackage{algorithm}
\usepackage{algpseudocode}
\usepackage{amsmath}
\usepackage{setspace}
\usepackage{wrapfig}
\usepackage{makecell}
\usepackage{tcolorbox}
\usepackage{diagbox}
\usepackage{arydshln}
\usepackage{enumitem}
\usepackage{placeins}
\usepackage{tcolorbox}
\tcbuselibrary{theorems}
\usepackage{fancyvrb}
\usepackage{fvextra}

\newcommand{\zhuoshi}[1]{\textcolor{purple}{#1}}
\newtcbtheorem[]{prompt}{Prompt}%
{colback=green2!5,colframe=blue1,fonttitle=\bfseries, left=.02in, right=.02in,bottom=.02in, top=.02in,before upper={\linespread{1.5}\selectfont}}{prompt}
\newtcbtheorem[number within=section]{exmp}{Example}%
{beforeafter skip balanced=.01in,colback=green2!5,colframe=blue1,fonttitle=\small \bfseries, left=.02in, right=.02in,bottom=.02in, top=.02in}{exmp}

\tcbuselibrary{most}

\definecolor{bluex}{rgb}{0.27, 0.42, 0.81}
\definecolor{purplex}{HTML}{9564bf}
\definecolor{red3}{HTML}{C52A20}
\definecolor{red2}{HTML}{B36A6F}
\definecolor{red1}{HTML}{FFb5b5}
\definecolor{purple}{HTML}{B36A6F}
\definecolor{darkyellow}{HTML}{D5BA82}
\definecolor{blue1}{HTML}{508AB2}
\definecolor{blue2}{HTML}{C4E4E3}
\definecolor{green1}{HTML}{A1D0C7}
\definecolor{green2}{HTML}{BFF6BA}
\definecolor{green3}{HTML}{028100}
\definecolor{teal}{HTML}{508AB2}
\definecolor{purple1}{HTML}{8d3a94}

\definecolor{lightgreen}{RGB}{200,255,200}
\definecolor{lightpink}{rgb}{1.0, 0.85, 0.9} 
\definecolor{lightblue}{rgb}{0.529, 0.808, 0.922} 
\definecolor{lightgray}{gray}{0.85}

\newtcolorbox{mybox}[2][]
{colback = black!5!white, colframe = black!75!black, fonttitle = \bfseries\small,
colbacktitle = black!100!black, enhanced,
attach boxed title to top left={yshift=-2.2mm,xshift=4mm},
title=#2,#1}

\title{REST: Stress Testing Large Reasoning Models by Asking Multiple Problems at Once}

\author{Zhuoshi Pan$^{1,2,\dagger,\ddagger}$ ~~Qizhi Pei$^{2,3,\dagger,\ddagger}$ ~~Yu Li$^{2}$  ~~Qiyao Sun$^2$  ~~Zinan Tang$^2$ \\ \textbf{H. Vicky Zhao$^{1,*}$} ~~\textbf{Conghui He$^{2,*}$} ~~\textbf{Lijun Wu$^{2,*}$}
}

\affiliation[1]{Tsinghua University}
\affiliation[2]{OpenDataLab, Shanghai Artificial Intelligence Laboratory}
\affiliation[3]{Renmin University of China}

\abstract{
Recent Large Reasoning Models (LRMs) have achieved remarkable progress on task-specific benchmarks, yet their evaluation methods remain constrained by isolated problem-solving paradigms. Existing benchmarks predominantly assess single-question reasoning through sequential testing, resulting critical limitations: (1) vulnerability to data contamination and less challenging (e.g., DeepSeek-R1 achieves 97.0\% on MATH500), forcing costly and perpetual creation of new questions with large human efforts, (2) failure to evaluate models under multi-context pressure, a key requirement for real-world deployment. 
To bridge this gap, we present \textbf{REST} (Reasoning Evaluation through Simultaneous Testing), a stress-testing framework that exposes LRMs to multiple problems simultaneously. 
Beyond basic reasoning, REST evaluates several under-tested capabilities: \textit{contextual priority allocation}, \textit{cross-problem interference resistance}, and \textit{dynamic cognitive load management}.
Our evaluation across \textbf{34} advanced reasoning models on \textbf{7} reasoning benchmarks reveals several striking findings:
\textit{Even state-of-the-art (SOTA) models like DeepSeek-R1 exhibit substantial performance degradation under stress testing}, challenging the prevailing assumption that ``LLMs are multi-problem solvers''.
Crucially, \textit{REST demonstrates stronger discriminative power than existing benchmarks}, revealing pronounced performance differences among models that exhibit similar, near-ceiling performance under single-question evaluations. 
Some key insights emerge from our analysis: \textit{(1) the ``overthinking trap'' is a critical factor contributing to the performance degradation}; \textit{(2) the models trained with ``long2short'' technique preserve more accuracy} of their single-problem performance under REST, outperforming standard-trained counterparts. 
These results establish REST as a cost-efficient, future-proof evaluation paradigm that better reflects real-world reasoning demands while reducing reliance on continuous human annotation.
}

\date{\today}
\correspondence{H. Vicky Zhao, \email{vzhao@tsinghua.edu.cn}; Conghui He, \email{heconghui@pjlab.org.cn}; Lijun Wu, \email{wulijun@pjlab.org.cn}}

\metadata[Code]{\url{https://github.com/opendatalab/REST}}

\begin{document}

\maketitle
\footnotetext[1]{Corresponding Authors.}
\footnotetext[2]{Equal Contribution.}
\footnotetext[3]{Work during internship at Shanghai Artificial Intelligence Laboratory.}

\begin{figure*}[!t]
    \centering
    \includegraphics[width=\linewidth]{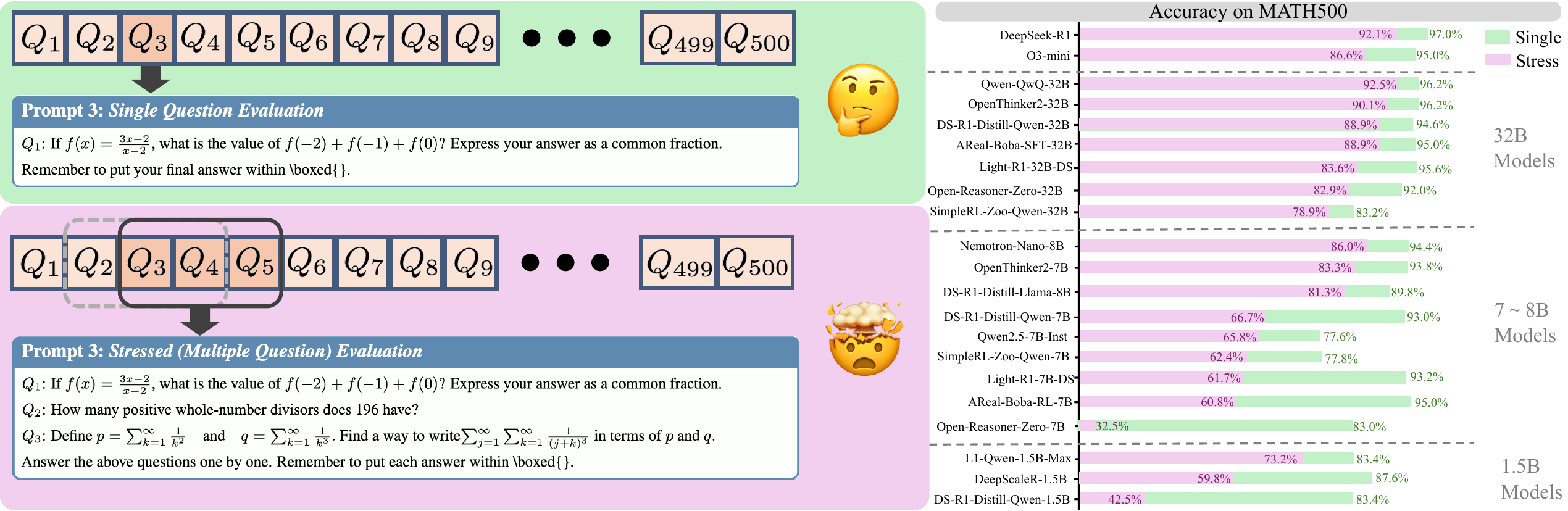}
    \caption{
    Illustration of \sysname\ evaluation compared with single-question evaluation. \sysname\ concatenates multiple questions into a single prompt, stress testing LRMs' ability in handling increasing reasoning loads within a single reasoning process.
    }
    \label{fig:REST}
    \vspace{-5mm}
\end{figure*}

\section{Introduction}

Recent years have witnessed significant advancements in the reasoning capabilities of Large Reasoning Models (LRMs), which have demonstrated impressive performance across a variety of reasoning tasks, including mathematical problem solving~\citep{yang2024qwen25mathtechnicalreportmathematical,pan2025lemma,shao2024deepseekmathpushinglimitsmathematical,pei2025mathfusion}, code generation~\citep{hui2024qwen25codertechnicalreport, rozière2024codellamaopenfoundation}, and complex concept understanding~\citep{rein2024gpqa}. Researchers have proposed various benchmarks~\citep{cobbe2021trainingverifierssolvemath,hendrycks2021measuringmathematicalproblemsolving,rein2024gpqa,jain2024livecodebench,li2025cipherbank} to evaluate the reasoning capabilities of LRMs.
However, the rapid advancement of LRMs has exposed critical limitations in current evaluation paradigms. Most benchmarks rely on single-question testing, where models process and answer questions in isolation. While this approach is effective for early model development, it now faces two fundamental challenges.

First, many benchmarks, such as GSM8K~\citep{cobbe2021trainingverifierssolvemath} and MATH~\citep{hendrycks2021measuringmathematicalproblemsolving}, are becoming less effective at distinguishing model performance, as current LRMs have achieved nearly saturated performance on them (e.g., \textit{DeepSeek-R1} achieves 97.0\% on MATH500), the community is forced into a costly cycle of benchmark obsolescence - continuously developing new, more difficult datasets (e.g., AIME24~\citep{ye2025aimepreview}) while discarding still-valuable existing ones. This raises our first key question: \textit\textbf{{Can we enhance the utility of current benchmarks by making them more challenging without complete replacement?}}

Second, single-question evaluation fails to assess how models perform in real-world, multi-context scenarios where reasoning must occur across multiple, potentially interfering questions. For instance, in educational tutoring systems, an AI might need to simultaneously address a student’s follow-up questions while correcting previous misconceptions; also, in technical support, it may process multiple user-reported issues within a single context window. Though a few prior studies~\citep{cheng2023batch,wang2024evaluating,son2024multi} have investigated multi-question prompting, they only focus on simple tasks such as text classification~\citep{semwal2018practitioners} or commonsense QA~\citep{talmor2019commonsenseqa}, which are insufficient to assess the reasoning capabilities of current LRMs. This leads to our second critical question: \textit{\textbf{How well can current LRMs handle multiple questions simultaneously, and what factors affect their performance in such settings?}} 

To answer the aforementioned research questions, we propose \textbf{\sysname}~(Reasoning Evaluation through Simultaneous Testing), a simple yet powerful method that repurposes existing benchmarks into more challenging variants. 
Specifically, \sysname~transforms existing benchmarks by concatenating multiple questions into a single instruction to evaluate these questions at once. 
Based on \textbf{7} representative reasoning benchmarks (\textit{e.g.,} GSM8K, MATH500, and AMC23), we reconstruct them into a multi-question format (see Fig.~\ref{fig:REST} for illustration). Through a comprehensive evaluation under our REST of \textbf{more than 30} state-of-the-art (SOTA) LRMS, we have obtained some valuable findings.
\begin{itemize}[leftmargin=0.6cm]
    \item Even SOTA LRMs like DeepSeek-R1~\citep{guo2025deepseek} \textbf{exhibit significant performance degradation} under \sysname, such as 29.1\% accuracy drop on AIME24, revealing a critical limitation in their reasoning robustness, challenging that ``LLMs are inherently multi-problem solvers''~\citep{cheng2023batch,wang2024evaluating,son2024multi}.
    \item Despite excelling in the single-question evaluation, several LRMs \textbf{struggle to maintain their advantage} under \sysname\ compared to their non-reasoning counterparts.
    \item \sysname~provides \textbf{enhanced discriminative power}. While many LRMs exhibit similar performance on single-question evaluations, they display notable differences in accuracy under \sysname.
    \item Effectively \textbf{revitalizes existing benchmarks}, making them challenging again for top-tier models.
\end{itemize}

Through further analysis, we identify a key factor behind this performance drop: the overthinking phenomenon~\citep{sui2025stop,chen2024not,wang2025thoughts}, where LRMs tend to generate unnecessarily redundant reasoning even for relatively simple problems.
Besides, LRMs trained with ``Long2Short'' training~\citep{sui2025stop,aggarwal2025l1,arora2025training}, which encourages concise reasoning, perform better in \sysname, suggesting a promising direction for developing multi-question-capable LRMs.
These findings have immediate implications for both evaluation practices and model development. REST offers a cost-effective, scalable alternative to constant benchmark replacement, while providing insights for building more robust reasoning systems capable of handling real-world, multi-context scenarios.

\section{Related Work}

\subsection{Compositional Instruction Training and Evaluation}
Some recent works use multi-instruction or problem combinations for training data augmentation~\citep{lin2025metaladder,pei2025mathfusion,li2024mosaic}.
\mbox{MathFusion}~\citep{pei2025mathfusion} fuses semantically similar problem pairs to construct a more difficult one for enhanced mathematical problem solving.
Mosaic-IT~\citep{li2024mosaic} concatenates randomly sampled instruction data with meta-instructions to reduce training cost and improve performance.
To evaluate the ability of LLMs to follow compositional instructions simultaneously, several benchmarks have been proposed. 
Batch prompting~\citep{cheng2023batch}, multi-problem prompting~\citep{ wang2024evaluating}, and multi-task inference benchmark~\citep{son2024multi} investigate concatenating multiple independent tasks, primarily for inference efficiency on simple tasks like text classification. 
Compound-QA~\citep{hou2024compound} focuses on increasing complexity within a single instruction, requiring models to follow multiple constraints, compositional directives, or sequential steps.
These methods, however, do not focus on the evaluation of complex reasoning tasks.

\subsection{Overthinking and Controlled Reasoning Length in LRMs}
Excessive verbosity in the reasoning processes of LRMs can hinder their inference efficiency.
The phenomenon of ``overthinking'', where models generate excessively long and complex reasoning steps without proportional gains in accuracy, has been identified~\citep{sui2025stop,chen2024not}.
In response, several ``long2short'' or controlled reasoning length methods emerge.
A prominent approach involves fine-tuning LRMs using reinforcement learning (RL) with penalties for excessive token generation~\citep{aggarwal2025l1,arora2025training,luo2025o1}, which effectively reduces the reasoning length while preserving or improving accuracy. 
Other strategies to promote conciseness include constrained prompting~\citep{nayab2024concise} and dynamic token budget allocation~\citep{han2024token}. 
Importantly, the validation of these length-control techniques occur in single-problem evaluation settings. 
Our \sysname{} provides a new dimension for assessing LRMs under multi-problem scenarios.

\subsection{Reasoning Benchmarks}
The rigorous assessment of the reasoning ability of LLMs relies heavily on specialized benchmarks. 
In mathematical reasoning, GSM8K~\citep{cobbe2021gsm8k} and MATH~\citep{hendrycks2021math} are most commonly used. 
As model capabilities surge, harder benchmarks like OlympiadBench~\citep{he2024olympiadbench}, AIME~\citep{aimo_validation_aime,aime2025}, and Omni-Math~\citep{gao2024omni} have emerged, further pushing the boundaries of mathematical evaluation. 
Beyond mathematics, reasoning in other domains like code generation~\citep{jain2024livecodebench,quan2025codeelo,swebench}, cryptographic decryption~\citep{li2025cipherbank}, and scientific problem-solving~\citep{rein2024gpqa}, is also critical.
Despite the increasing difficulty and diversity of these benchmarks, the rapid improvement of models means that their differentiating power can diminish when evaluated solely through standard single-instance problem solving.
Our \sysname{} builds on these existing benchmarks and repurposes them into more challenge variants.
\section{\sysname: Reasoning Evaluation through Simultaneous Testing }

Despite the significant progress in large reasoning models (LRMs), many existing benchmarks, such as MATH500~\cite{hendrycks2021measuringmathematicalproblemsolving}, have reached near-saturation performance levels, limiting their utility in distinguishing between the capabilities of increasingly powerful models. Our introduced REST aims to address this limitation, which is a new evaluation protocol designed to systematically increase the cognitive load on LRMs by aggregating multiple questions into a single prompt. This design enables a finer-grained assessment of a model’s ability to handle multi-step, continuous reasoning under stress. Below are the details of the benchmark reconstruction and evaluation under REST.

\subsection{REST Benchmark Reconstruction}
Let the original benchmark be denoted as $\mathcal{Q} = \{q_1, q_2, \dots q_N\}$, where $q_i$ represents an individual question and $N$ is the total number of questions. In REST, we transform this benchmark into a new prompt set $\mathcal{P}^s$ by concatenating $s$ consecutive questions into each prompt. We refer to the parameter $s \in \mathbb{Z}^+$ as the stress level, since a larger $s$ imposes a greater reasoning burden on the model.
Formally, for each $i \in \{1, 2, \dots, N\}$,  we define the stress-level-$s$ prompt $p_i^s$ as:
\begin{equation*}
    p_{i}^{s} = \texttt{Compose}(q_{i}, q_{i+1}, \dots, q_{[(i+s-1)\bmod N]}), \quad i \in \{1, 2, \dots, N\}.
\end{equation*}

To ensure continuity and full coverage, we apply cyclic indexing when the end of the benchmark is reached. The function $\texttt{Compose}()$ formats multiple questions into a single prompt:
\begin{equation*}
    \texttt{Compose}(q_1, \dots, q_s) = ``Q1:  \{q_1\}, \dots, Qs:  \{q_s\}. \text{ Answer the above questions one by one.''}
\end{equation*}
This transformation yields a new prompt set $\mathcal{P}^s = \{p_1^s, p_2^s,...,p_N^s\}$, where each prompt contains $s$ questions. Importantly, $\|P^s\|=N$, matching the size of the original benchmark. 
However, unlike the original benchmark, every prompt $p_i^s$ in $\mathcal{P}^s$ comprises $s$ consecutive questions, spanning from $q_i$ to $q_{[(i+s-1) \bmod N]}$ rather than just a single question $q_i$. 
Moreover, each original question $q_i$ appears exactly $s$ times across all the prompts and exactly once in each of the $s$ possible positions within the concatenated prompts. This design mitigates positional biases and ensures comprehensive coverage across stress levels.

\subsection{REST Evaluation}
Under REST, given a LRM model $f$, we evaluate it on each prompt $p_i^s$ to obtain an output response $o_i^s=f(p_i^s)$. From this response, we extract the individual predicted answers $\{\hat{a}_i^{s}, \hat{a}_{i+1}^{s}, \dots, \hat{a}_{[(i+s-1)\bmod N]}^{s}\}$ using a function $\texttt{Extract}()$:
\begin{equation*}
    \{\hat{a}_i^{s}, \hat{a}_{i+1}^{s}, \dots, \hat{a}_{[(i+s-1)\bmod N]}^{s}\} = \texttt{Extract}(o_i^s).
\end{equation*}

To facilitate accurate extraction, we instruct LRMs to format their answers in task-specific ways. For Mathematical and GPQA problems, we instruct the model to put each answer within ``\texttt{\textbackslash boxed\{\}}'', while for code generation tasks, answers must be wrapped in Python code blocks ``\texttt{\textasciigrave\textasciigrave\textasciigrave python \textasciigrave\textasciigrave\textasciigrave}''. 
Notably, although a more structured format method like JSON output could offer convenience for output parsing, we deliberately avoid imposing such strict formatting constraints, as prior studies~\citep {tam2024let, shorten2024structuredrag} indicate that such requirements will degrade performance.
We employ both rule-based and LLM-based extraction methods: (a) The rule-based method uses regular expressions to extract answers from predefined markers (e.g., ``\texttt{\textbackslash boxed\{\}}''); (b) The LLM-based method prompts a model to retrieve the predicted answer from the response for each question. Further implementation details can be found in Appendix~\ref{sec:answer_extraction}.
Finally, we define the model’s accuracy at stress level $s$ by comparing the predicted answer $\hat{a}$ and the ground truth answer $a$: 
\begin{equation*}
\label{eqn:acc}
    \textit{Acc}(\mathcal{P}^{s}) = \frac{1}{N}\sum_{i=1}^{N} \textit{Acc}(p_{i}^{s}) = \frac{1}{N}\sum_{i=1}^{N} \frac{1}{s} \sum_{j=1}^{s} \delta(\hat{a}_{[(i+j-1)\bmod N]}^{s}, a_{[(i+j-1)\bmod N]}^{s}),
\end{equation*}
where $\delta$ is the Kronecker delta function such that $\delta(i,j) = 1 $ if and only if $i=j$ and 0 otherwise.

\section{Experiment}

\subsection{Evaluation Setup}
\label{sec:evaluation_setup}

We evaluate a total of 34 LRMs, spanning a parameter size range from 1.5B to 671B. The temperature and top\_p parameters are configured according to the corresponding official guidelines for each model. We set a maximum output token length of 32K for reasoning models and 8K for non-reasoning models. In addition, we conduct experiments with an extended 128K token limit, but observe negligible performance differences, as shown in Tab.~\ref{tab:output_128k} of Appendix~\ref{sec:impact_of_max-out-len}. Our evaluation is based on OpenCompass\footnote{\url{https://github.com/open-compass/opencompass}} toolkit. 
To ensure consistency, we adopt the official prompt for each task, with a minor format adjustment for multi-question responses: ``Answer the above questions one by one. Remember to put each answer within \texttt{\textbackslash boxed\{\}} (or \texttt{\textasciigrave\textasciigrave\textasciigrave python \textasciigrave\textasciigrave\textasciigrave} for code generation).''

We select $7$ representative benchmarks for evaluation, and in \sysname, we set different stress levels $s$ for each benchmark. Specifically, for relatively simple benchmarks like GSM8K, the stress levels are $s \in \{1, 3, 6, 9, 12\}$; for medium-difficulty benchmarks, including MATH500 and AMC23, they are set as $s \in \{1, 3, 5, 7, 9\}$; and for more challenging benchmarks, such as AIME24, AIME25, GPQA, and LiveCodeBench (v5), the stress levels are $s \in \{1, 2, 3, 4, 5\}$.  
To ensure clarity and consistency in evaluation, we report the performance of \sysname~as the average accuracy across stress levels greater than 1. 
Detailed accuracy statistics for each stress level are provided in Appendix~\ref{sec:detailed_result}. For relatively small benchmarks like AIME24, AIME25, and AMC23, we conduct 8 sampling runs and report the average results to reduce variance.

\subsection{Evaluation Result}

\begin{table}[t!]
    \centering
    \caption{Evaluation results on various math benchmarks. The ``Stress'' column reports the average performance across four stress levels, as described in Sec.~\ref{sec:evaluation_setup}. Accuracy on other tasks are in Tab.~\ref{tab:main_other}. Detailed accuracy statistics for each stress level can be found in Tab.~\ref{tab:main_detail1} of the Appendix. Models that perform best under ``Stress'' are highlighted in \textcolor{gray}{gray}.}
    \def\arraystretch{1,2}
    \resizebox{\textwidth}{!}{
    \begin{tabular}{c|c>{\cellcolor{cyan!5}}c|c>{\cellcolor{cyan!5}}c|c>{\cellcolor{cyan!5}}c|c>{\cellcolor{cyan!5}}c|c>{\cellcolor{cyan!5}}c|c>{\cellcolor{cyan!5}}c}
    \toprule[0.8pt]
       \multirow{2}{*}{\diagbox{Model}{Bench}} & \multicolumn{2}{c}{GSM8K} & \multicolumn{2}{c}{MATH500} & \multicolumn{2}{c}{AMC23} & \multicolumn{2}{c}{AIME24} & \multicolumn{2}{c}{AIME25} & \multicolumn{2}{c}{Avg.} \\
       & Single & Stress & Single & Stress & Single & Stress & Single & Stress & Single & Stress & Single & Stress \\
       \hline
       \rowcolor{blue!15}
       \multicolumn{13}{c}{\textit{1.5B LRMs}} \\

       DS-R1-Distill-Qwen-1.5B~\citep{guo2025deepseek} & 84.62 & 70.21 & 83.40 & 42.47 & 62.50 & 13.98 & 29.17 & 4.97 & 25.00 & 5.91 & 56.94 & 27.51 \\
       DeepScaleR-1.5B~\citep{deepscaler2025} & 84.84 & 66.58 & \textbf{87.60} & 59.77 & 76.25 & 32.05 & \textbf{38.75} & 12.82 & \textbf{31.25} & 14.23 & \textbf{63.74} & 37.09 \\
       L1-Qwen-1.5B-Exact~\citep{aggarwal2025l1} & 84.87 & 79.01 & 84.00 & 72.07 & 71.25 & 47.37 & 21.25 & 12.62 & 18.33 & 12.96 & 55.94 & 44.81 \\
       \cellcolor{lightgray}L1-Qwen-1.5B-Max~\citep{aggarwal2025l1} & 84.17 & \cellcolor{lightgray}\textbf{78.29} & 83.40 & \cellcolor{lightgray}\textbf{73.23} & \textbf{77.50} & \cellcolor{lightgray}\textbf{48.37} & 20.00 & \cellcolor{lightgray}\textbf{15.13} & 22.92 & \cellcolor{lightgray}\textbf{14.95} & 57.60 & \cellcolor{lightgray}\textbf{45.99} \\
       Qwen2.5-Math-1.5B-Inst~\citep{yang2024qwen2} & \textbf{85.37} & 67.49 & 73.00 & 53.94 & 57.50 & 22.22 & 10.83 & 6.17 & 10.83 & 2.83 & 47.51 & 30.53 \\
       \rowcolor{blue!15}
       \multicolumn{13}{c}{\textit{$7\sim8$B LRMs}} \\
       DS-R1-Distill-Qwen-7B~\citep{guo2025deepseek} & 89.49 & 89.06 & 93.00 & 66.75 & 87.50 & 36.06 & 54.17 & 16.53 & 35.42 & 11.37 & 71.92 & 43.95 \\
       DS-R1-Distill-LLaMA-8B~\citep{guo2025deepseek} & 90.45 & 85.18 & 89.80 & 81.34 & 87.50 & 70.75 & 50.42 & 31.23 & 28.33 & 22.66 & 69.30 & 58.23 \\
       Efficient-R1-7B ($\alpha=0.1$)~\citep{arora2025training} & 88.63 & 84.76 & 90.00 & 74.99 & 87.50 & 44.25 & 54.58 & 21.45 & 35.42 & 15.75 & 71.23 & 48.24 \\
       Efficient-R1-7B ($\alpha=0.2$)~\citep{arora2025training} & 87.95 & 80.38 & 88.20 & 76.41 & 85.00 & 48.05 & 50.42 & 22.12 & 33.75 & 17.25 & 69.06 & 48.84 \\
       \cellcolor{lightgray}Nemotron-Nano-8B~\citep{bercovich2025llamanemotronefficientreasoningmodels} & 91.36 & 70.52 & 94.40 & \cellcolor{lightgray}\textbf{86.04} & 90.00 & \cellcolor{lightgray}\textbf{76.24} & \textbf{63.33} & \cellcolor{lightgray}\textbf{43.55} & \textbf{50.00} & \cellcolor{lightgray}\textbf{32.28} & \textbf{77.82} & \cellcolor{lightgray}\textbf{61.71} \\
       AReaL-boba-RL-7B~\citep{mei2025real} & 91.66 & 77.80 & \textbf{95.00} & 60.77 & \textbf{91.25} & 32.94 & 61.25 & 21.43 & 45.83 & 12.33 & 77.00 & 41.05 \\
       Light-R1-7B-DS~\citep{wen2025light} & 88.05 & 82.69 & 93.20 & 61.73 & 90.00 & 34.91 & 55.83 & 16.63 & 45.83 & 12.96 & 74.58 & 41.78 \\
       OpenR1-Qwen-7B~\citep{openr1} & \textbf{95.60} & 90.22 & 92.20 & 81.64 & 83.75 & 54.11 & 47.50 & 26.77 & 32.92 & 21.19 & 70.39 & 54.79 \\
       OpenThinker2-7B~\citep{OpenThoughts2025} & 94.39 & \cellcolor{lightgray}\textbf{91.99} & 93.80 & 83.30 & 85.00 & 63.23 & 54.58 & 34.50 & 41.67 & 23.66 & 73.89 & 59.33 \\
       SimpleRL-Zoo-Qwen-7B~\citep{zeng2025simplerlzooinvestigatingtamingzero} & 90.52 & 84.01 & 77.80 & 62.41 & 68.50 & 16.16 & 26.67 & 7.55 & 10.00 & 6.47 & 54.70 & 35.32 \\
       Open-Reasoner-Zero-7B~\citep{hu2025open} & 92.87 & 65.14 & 83.00 & 32.51 & 60.00 & 31.23 & 17.92 & 6.13 & 16.25 & 3.89 & 54.01 & 27.78 \\
       Marco-O1-7B~\citep{zhao2024marcoo1openreasoningmodels} & 89.08 & 79.56 & 72.40 & 48.19 & 47.50 & 17.23 & 10.00 & 4.35 & 10.83 & 3.64 & 45.96 & 30.59 \\
       MathFusion-Qwen-7B~\citep{pei2025mathfusion} & 89.46 & 83.78 & 74.00 & 68.15 & 52.50 & 36.24 & 9.58 & 7.89 & 5.83 & 2.35 & 46.27 & 39.68 \\
       Eurus-2-7B-PRIME~\citep{cui2025process} & 92.72 & 88.01 & 81.40 & 64.69 & 62.50 & 38.58 & 20.83 & 10.84 & 14.58 & 4.49 & 54.41 & 41.32 \\
       Qwen2.5-Math-7B-Inst~\citep{yang2024qwen25mathtechnicalreportmathematical} & 95.53 & 78.53 & 83.60 & 56.59 & 60.00 & 28.46 & 14.17 & 6.40 & 11.67 & 5.33 & 52.99 & 35.06 \\
       Qwen2.5-7B-Inst~\citep{yang2024qwen25} & 92.27 & 85.12 & 77.60 & 65.78 & 42.50 & 34.46 & 10.00 & 7.02 & 3.75 & 3.32 & 45.22 & 39.14 \\
       \rowcolor{blue!15}
       \multicolumn{13}{c}{\textit{32B LRMs}} \\
       DS-R1-Distill-Qwen-32B~\citep{guo2025deepseek} & 95.54 & 95.50 & 94.60 & 88.97 & 94.75 & \cellcolor{lightgray}\textbf{86.24} & 72.92 & 52.51 & 51.67 & 33.83 & 81.90 & 71.41 \\
       \cellcolor{lightgray}Qwen-QwQ-32B~\citep{qwq32b} & 95.83 & \cellcolor{lightgray}\textbf{95.78} & \textbf{96.20} & \cellcolor{lightgray}\textbf{92.49} & 95.00 & 82.89 & \textbf{78.75} & \cellcolor{lightgray}\textbf{54.79} & \textbf{69.58} & \cellcolor{lightgray}\textbf{41.53} & \textbf{87.07} & \cellcolor{lightgray}\textbf{73.49} \\
       AReaL-boba-SFT-32B~\citep{mei2025real} & 95.01 & 94.75 & 95.00 & 88.92 & \textbf{97.50} & 78.96 & 77.50 & 45.79 & 60.00 & 33.55 & 85.00 & 68.39 \\
       Light-R1-32B-DS~\citep{wen2025light} & 95.83 & 94.79 & 95.60 & 83.66 & 96.25 & 68.80 & 77.50 & 41.26 & 60.00 & 33.80 & 85.04 & 64.46 \\
       S1.1-32B~\citep{muennighoff2025s1simpletesttimescaling} & 89.84 & 61.10 & 90.40 & 53.85 & 90.00 & 32.26 & 55.83 & 24.42 & 45.42 & 19.13 & 74.30 & 38.15 \\
       OpenThinker2-32B~\citep{OpenThoughts2025} & \textbf{96.44} & 95.17 & \textbf{96.20} & 90.10 & 95.00 & 81.00 & 68.33 & 53.01 & 52.50 & 38.20 & 81.69 & 71.49 \\
       SimpleRL-Zoo-Qwen-32B~\citep{zeng2025simplerlzooinvestigatingtamingzero} & 96.06 & 93.49 & 83.20 & 78.90 & 67.50 & 57.02 & 27.20 & 16.80 & 16.67 & 8.87 & 58.13 & 51.01 \\ 
       Open-Reasoner-Zero-32B~\citep{hu2025open} & 95.83 & 91.80 &  92.00 & 82.90 & 83.75 & 70.04 & 46.67 & 31.65 & 36.67 & 23.63 & 70.98 & 60.00 \\
       Qwen2.5-32B-Inst~\citep{yang2024qwen25} & 95.53 & 93.77 & 82.20 & 73.39 & 60.00 & 49.72 & 20.00 & 9.61 & 16.67 & 6.73 & 54.88 & 46.64 \\

       \rowcolor{blue!15}
       \multicolumn{13}{c}{\textit{API-based LRMs}} \\
        \cellcolor{lightgray}DeepSeek-R1~\citep{guo2025deepseek} & \textbf{96.20} & \cellcolor{lightgray}\textbf{96.16} & 97.00 & \cellcolor{lightgray}\textbf{92.09} & 93.75 & 81.80 & \textbf{81.66} & \cellcolor{lightgray}\textbf{52.49} & 68.75 & 37.17 & \textbf{87.47} & \cellcolor{lightgray}\textbf{71.94} \\
        O3-mini~\citep{openai2025o3mini} & 95.83 & 93.85 & 95.00 & 86.62 & 90.00 & 59.17 & 79.16 & 34.07 & 71.66 & 20.63 & 86.25 & 55.51 \\
        O4-mini~\citep{openai2025o3mini} & 93.71 & 93.07 & 90.00 & 82.40 & 96.25 & \cellcolor{lightgray}\bf 82.79 & 73.33 & 49.69 & \bf 80.00 & \cellcolor{lightgray}\bf 41.42 & 86.66 & 69.87 \\
        Gemini-2.5-Flash-Thinking~\citep{team2023gemini} & 89.23 & 91.28 & \bf 97.20 & 69.92 & \bf 97.50 & 47.63 & 76.67 & 26.60 & 71.67 & 16.54 & 86.45 & 50.39 \\
    \bottomrule[0.8pt]
    \end{tabular}
    }
    \vspace{1mm}
    \label{tab:main}
\end{table}

Tab.~\ref{tab:main} and Tab.~\ref{tab:main_other} present the performance of various LRMs on the original reasoning benchmarks (Single) and their transformed counterparts by \sysname~(Stress). Our key findings from the experimental results can be summarized as follows:

\paragraph{LRMs can handle multiple simple questions at once, but struggle with challenging ones.}
LRMs can address several relatively simple problems within a single reasoning process. As shown in Tab.~\ref{tab:main}, on GSM8K, the accuracy drops under REST for DeepSeek-R1-Distill-Qwen-7B (R1-7B) and DeepSeek-R1-Distill-Qwen-32B (R1-32B) are just 0.43\% and 0.04\%, respectively.
However, this robustness diminishes significantly when the models face more challenging problems. 
In Tab.~\ref{tab:main} and Tab.~\ref{tab:main_other}, even SOTA LRMs (e.g., DeepSeek-R1) encounter substantial difficulties when tasked with multiple Olympiad-level competition problems within a single reasoning process. 
For instance, under \sysname, the accuracy of DeepSeek-R1 on AIME24 and AIME25 decreases by 29.17\% and 31.58\%, respectively, compared to the single-question setting. 
This performance gap is even more pronounced for smaller models, such as R1-7B, whose accuracy on AIME24 drops sharply from 54.17\% to 16.53\% under \sysname. Moreover, despite strong performance in the single-question setting, several LRMs fail to maintain their advantage under REST compared to their non-reasoning counterparts. For example, DeepSeek-R1-Distill-Qwen-1.5B achieves an accuracy of 83.40\% on MATH500 in the single-question evaluation, surpassing the 73.00\% of the Qwen2.5-Math-1.5B-Instruct model. However, under REST, its accuracy drops sharply to 42.47\%, falling significantly behind Qwen2.5-Math-1.5B-Instruct, which achieves 53.94\%.

\begin{figure*}[!t]
    \centering
    \begin{subfigure}{\textwidth}
    \includegraphics[width=\linewidth]{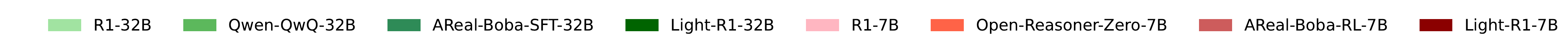}
    \end{subfigure}\\
    \vspace{-1mm}
    \begin{subfigure}{0.3\textwidth}
    \includegraphics[width=\linewidth]{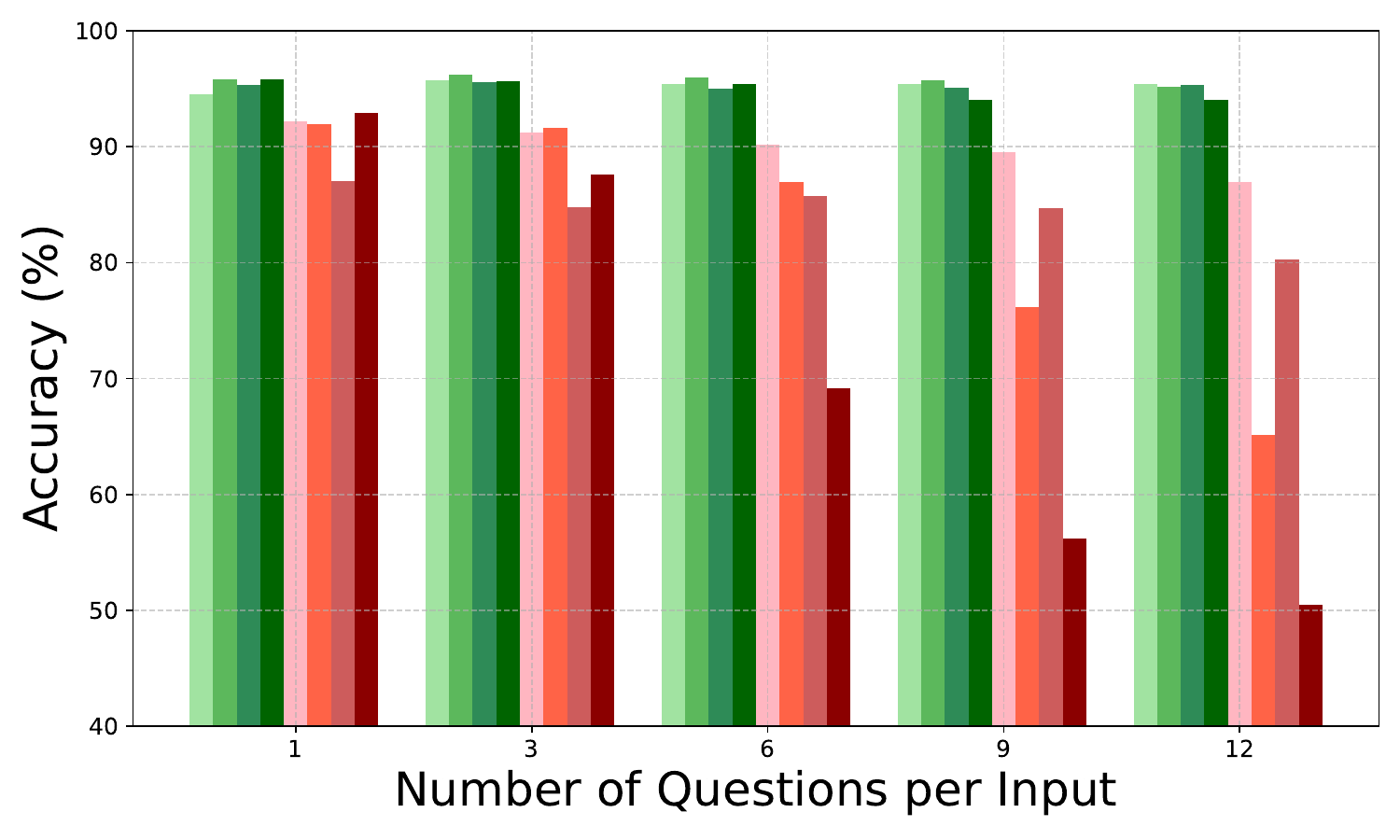}
    \vspace{-5mm}
    \caption{GSM8K}
    \label{fig:bar_gsm8k_size}
    \end{subfigure}
    \hfill
    \begin{subfigure}{0.3\textwidth}
    \includegraphics[width=\linewidth]{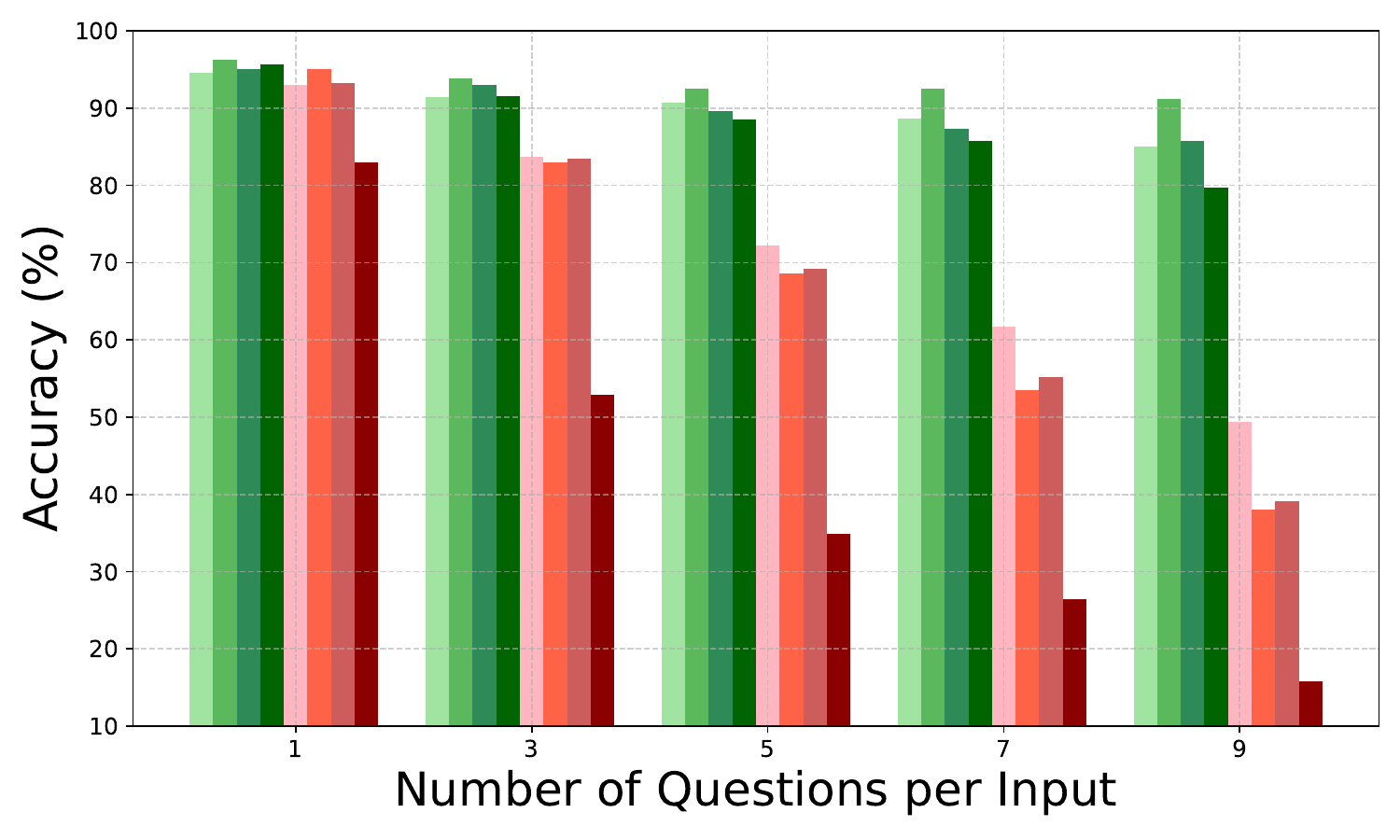}
    \vspace{-5mm}
    \caption{MATH500}
    \label{fig:bar_math500_size}
    \end{subfigure}
    \hfill
    \begin{subfigure}{0.3\textwidth}
    \includegraphics[width=\linewidth]{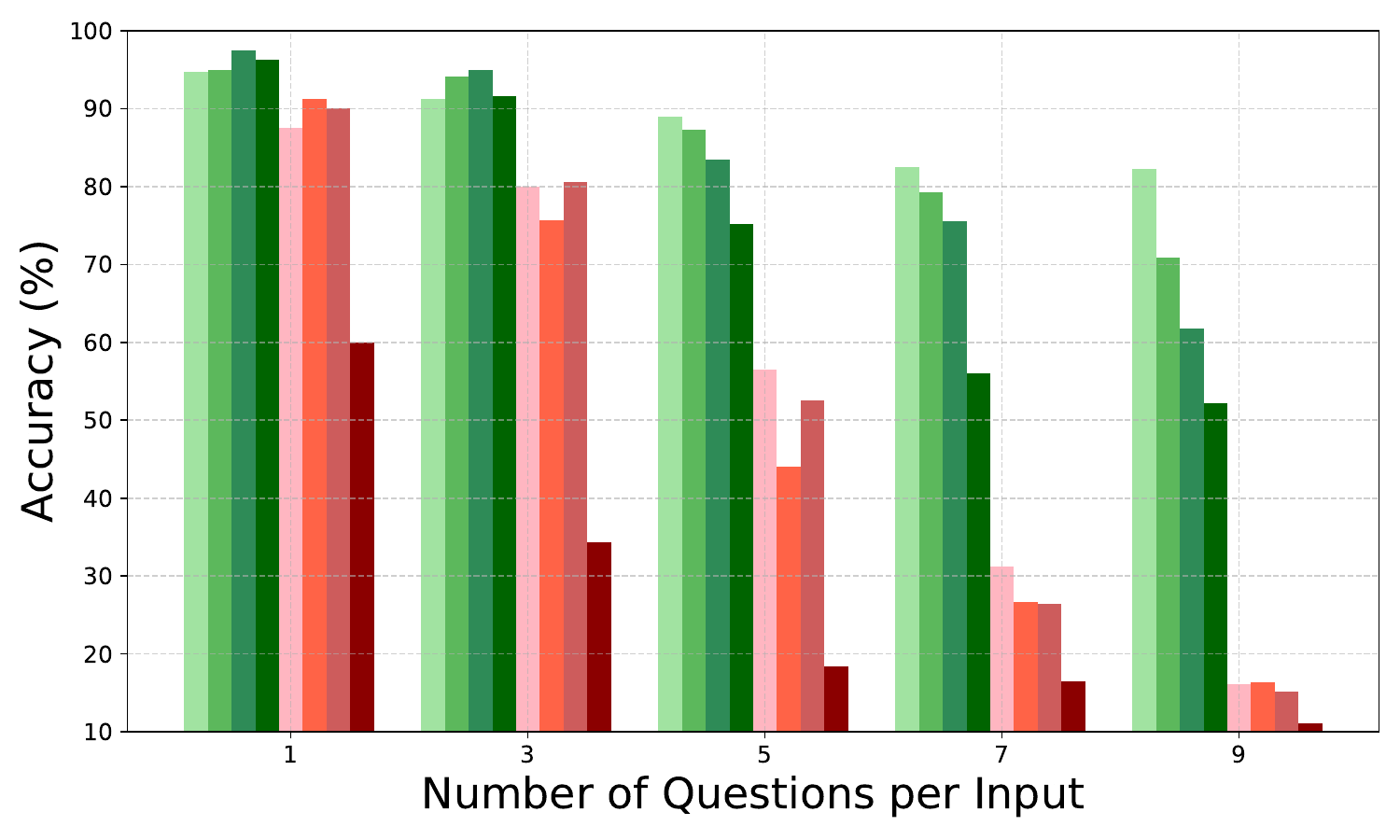}
    \vspace{-5mm}
    \caption{AMC23}
    \label{fig:bar_amc23_size}
    \end{subfigure}
    \caption{
    Performance comparison of LRMs of different sizes under various stress levels. \sysname\ reveals significant performance disparities among models of different sizes, although they demonstrate comparable, near-ceiling performance in traditional single-question evaluations.
    }
    \label{fig:bar_acc_size}
    \vspace{-2mm}
\end{figure*}

\begin{figure*}[!t]
    \centering
    \begin{subfigure}{0.3\textwidth}
    \includegraphics[width=\linewidth]{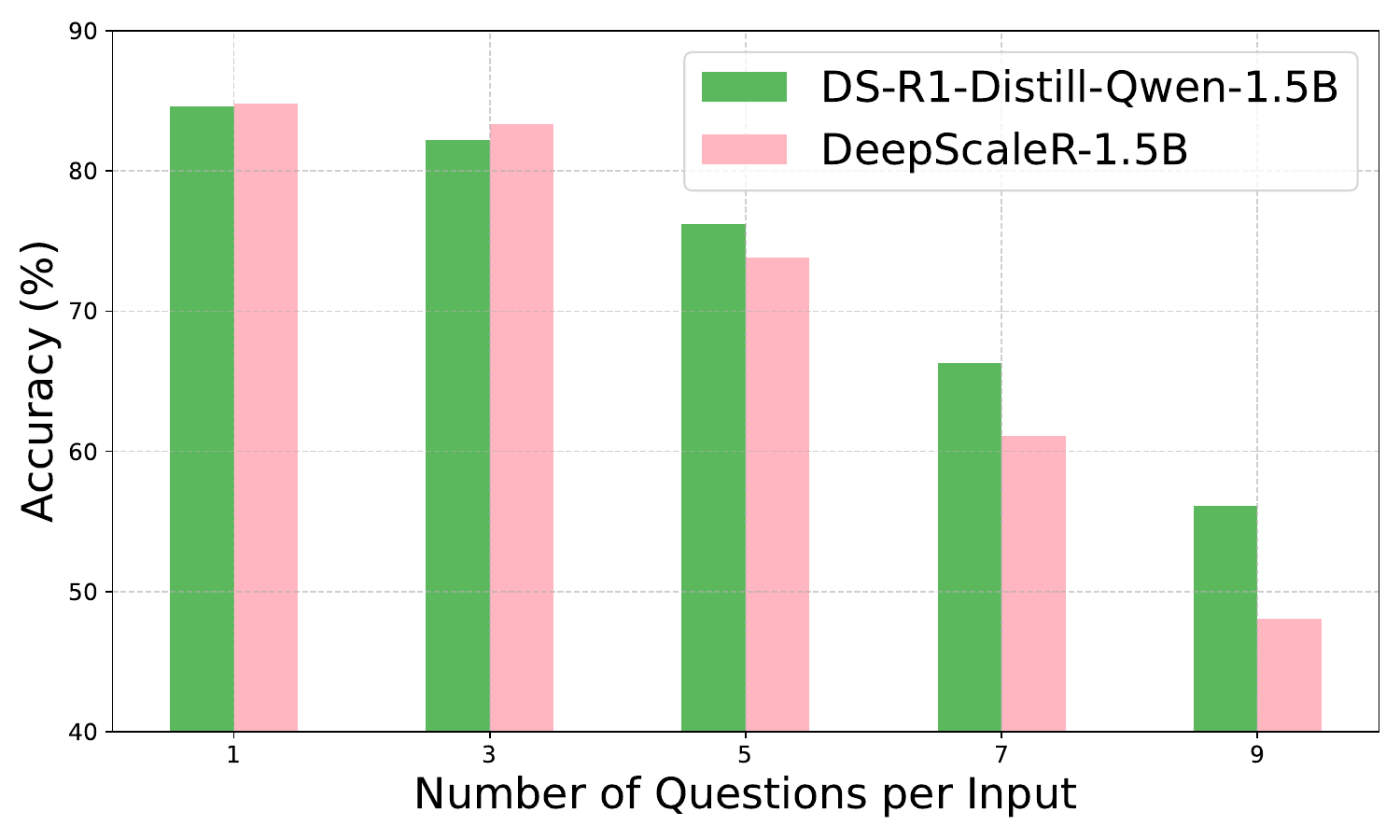}
    \vspace{-5mm}
    \caption{1.5B Model on GSM8K}
    \label{fig:bar_gsm8k_post}
    \end{subfigure}
    \hfill
    \begin{subfigure}{0.3\textwidth}
    \includegraphics[width=\linewidth]{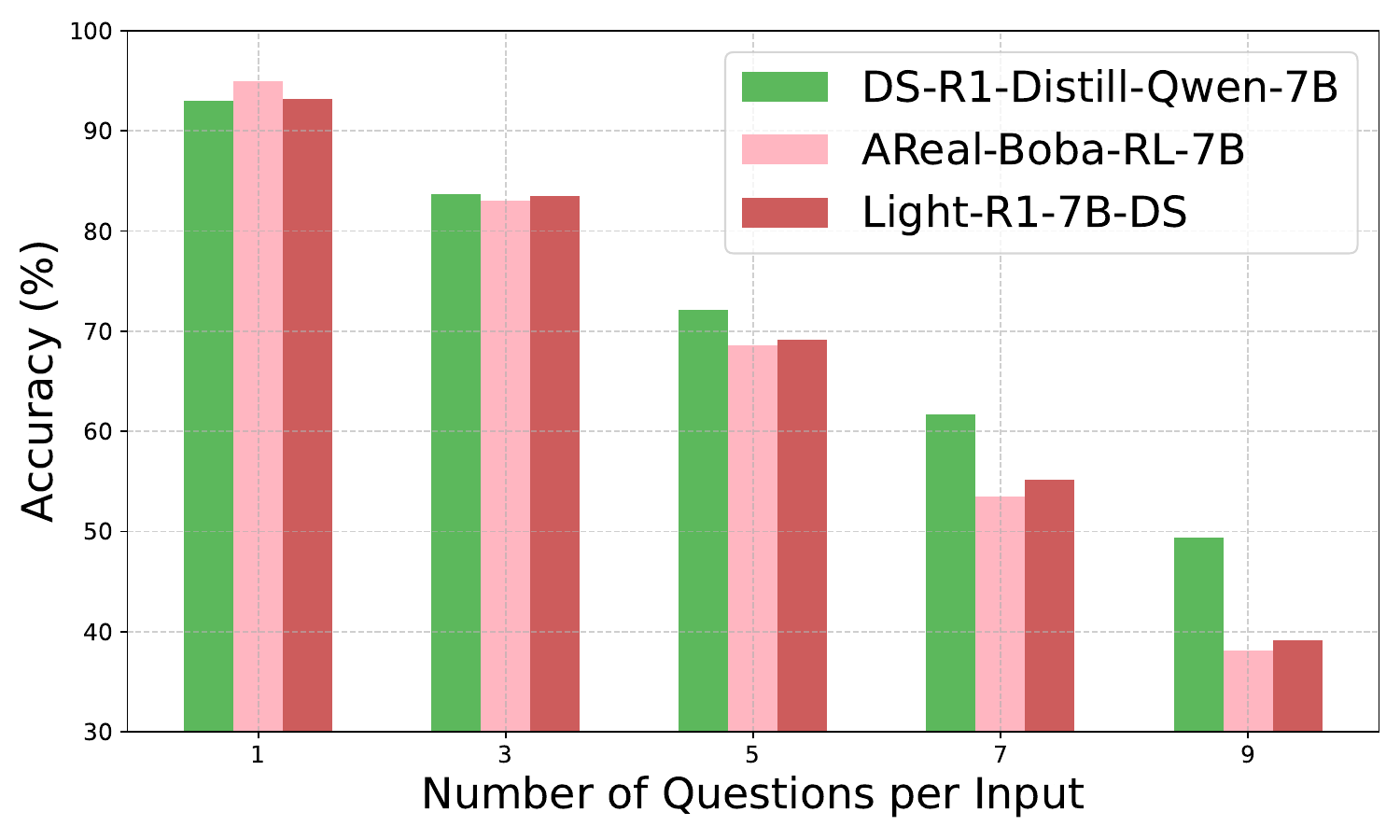}
    \vspace{-5mm}
    \caption{7B Model on MATH500}
    \label{fig:bar_math500_post}
    \end{subfigure}
    \hfill
    \begin{subfigure}{0.3\textwidth}
    \includegraphics[width=\linewidth]{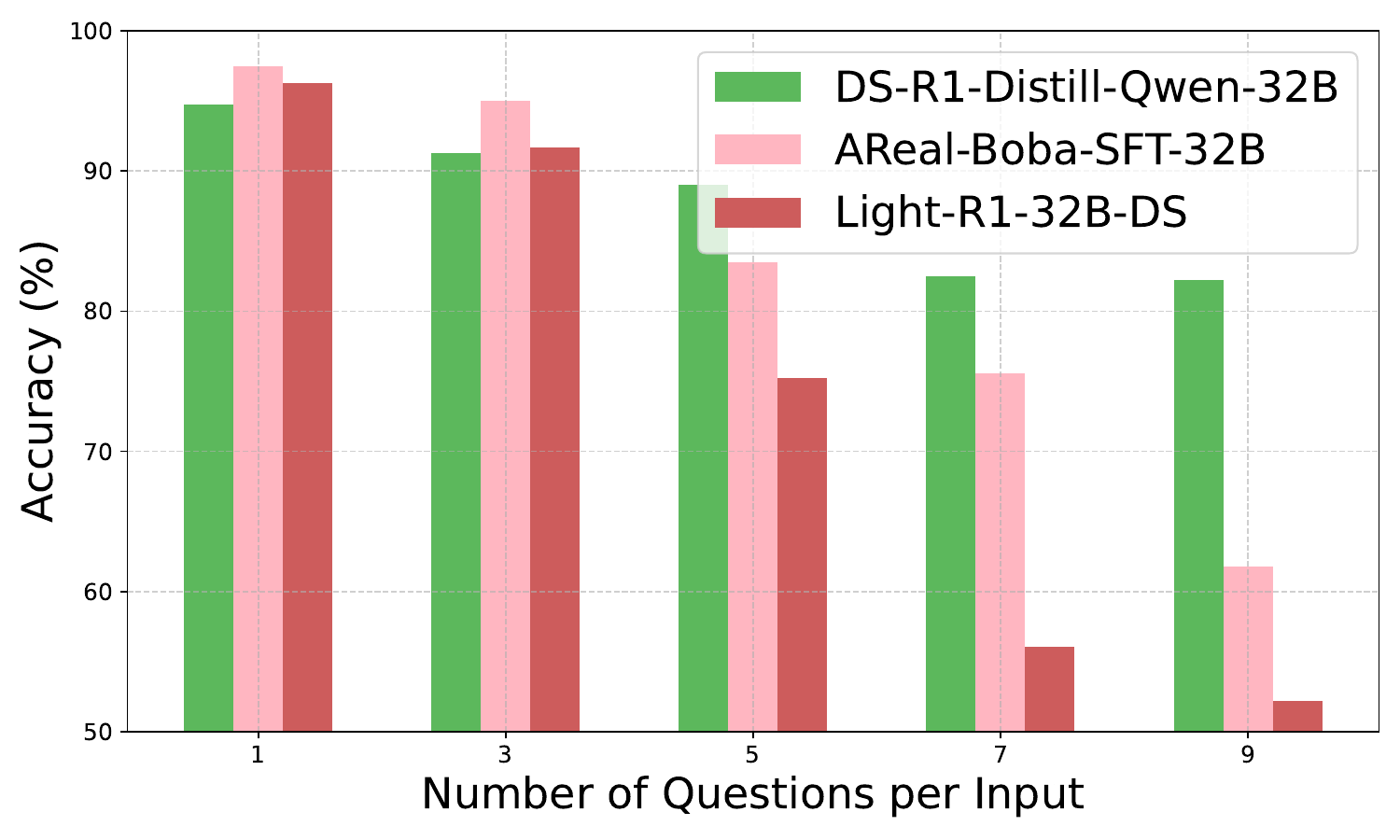}
    \vspace{-5mm}
    \caption{32B Model on AMC23}
    \label{fig:bar_amc23_post}
    \end{subfigure}
    \caption{
    Limitation of post training on \sysname. Post training exacerbates performance degradation under higher stress levels compared to the original DeepSeek distilled models.
    }
    \label{fig:limitation_of_RLVR}
    \vspace{-5mm}
\end{figure*}

\paragraph{\sysname\ enhances the discriminative power of existing benchmarks.}
By concatenating multiple problems into a single prompt, \sysname\ significantly amplifies performance differences among models that exhibit comparable, near-ceiling accuracy in traditional single-question evaluations.
For example, on MATH500, R1-7B and R1-32B achieve single-question accuracies of 93.0\% and 94.6\%, respectively, with R1-7B trailing by just 1.6\%. However, when evaluated under \sysname, the accuracy of R1-7B plummets to 66.75\%, while R1-32B maintains a substantially higher 88.97\%, revealing a pronounced 22.22\% performance gap.
This contrast is similarly evident among models of the same size. 
For instance, AReaL-boba-RL-7B and OpenThinker2-7B achieve similar, near-ceiling single-question accuracies of 95.0\% and 93.8\% on MATH500, respectively. 
Yet, under \sysname, the accuracy of AReaL-boba-RL-7B drops sharply to 60.77\%, while OpenThinker2-7B preserves a significantly higher 83.30\% accuracy. Similar trends are observed on AMC23, validating that \sysname\ can enhance the discriminative power of existing medium-difficulty benchmarks through problem concatenation.

\paragraph{Limitations of post training.}
The recent success of DeepSeek-R1~\citep{guo2025deepseek} has inspired numerous follow-up studies~\citep{deepscaler2025,mei2025real,wen2025light,openr1,hu2025open,zeng2025simplerlzooinvestigatingtamingzero}, some of which have adopted post RL training with verifiable rewards (RLVR)~\citep{deepscaler2025,mei2025real} or supervised fune-tuning (SFT)~\citep{wen2025light} based on DeepSeek-distilled models. Although such post training often yields improved single-question accuracy, our findings suggest that they can not maintain their performance advantage on more challenging, multi-question scenarios as in \sysname.
For instance, models like AReaL-boba-RL-7B and Light-R1-7B-DS, both post-trained on DeepSeek-R1-Distill-Qwen-7B, achieve single-question accuracies of 95.00\% and 93.20\% on MATH500, slightly higher than the 93.00\% accuracy of R1-7B. However, under \sysname, their accuracies drop significantly to 60.77\% and 61.73\%, respectively, both falling behind the 66.75\% achieved by R1-7B. This discrepancy becomes even more pronounced at higher stress levels. As detailed in Fig.~\ref{fig:bar_math500_post} and Tab.~\ref{tab:main_detail2} (in Appendix), \textbf{at stress level $s = 9$}, the accuracies of AReaL-boba-RL-7B and Light-R1-7B-DS decline sharply to 38.07\% and 39.11\%, respectively, more than 10.00\% below the 49.42\% achieved by R1-7B. This performance degradation is not limited to smaller models but also extends to larger ones, such as Light-R1-32B-DS and R1-32B, and is consistently observed across other benchmarks like AMC23.
These findings reinforce recent calls~\citep{gandhi2025cognitive,yue2025does,shah2025rethinking,hochlehnert2025sober} for a more cautious assessment of the benefits of post RL training in enhancing reasoning capabilities.

\begin{figure*}[!h]
    \centering
    \begin{subfigure}{0.3\textwidth}
    \includegraphics[width=\linewidth]{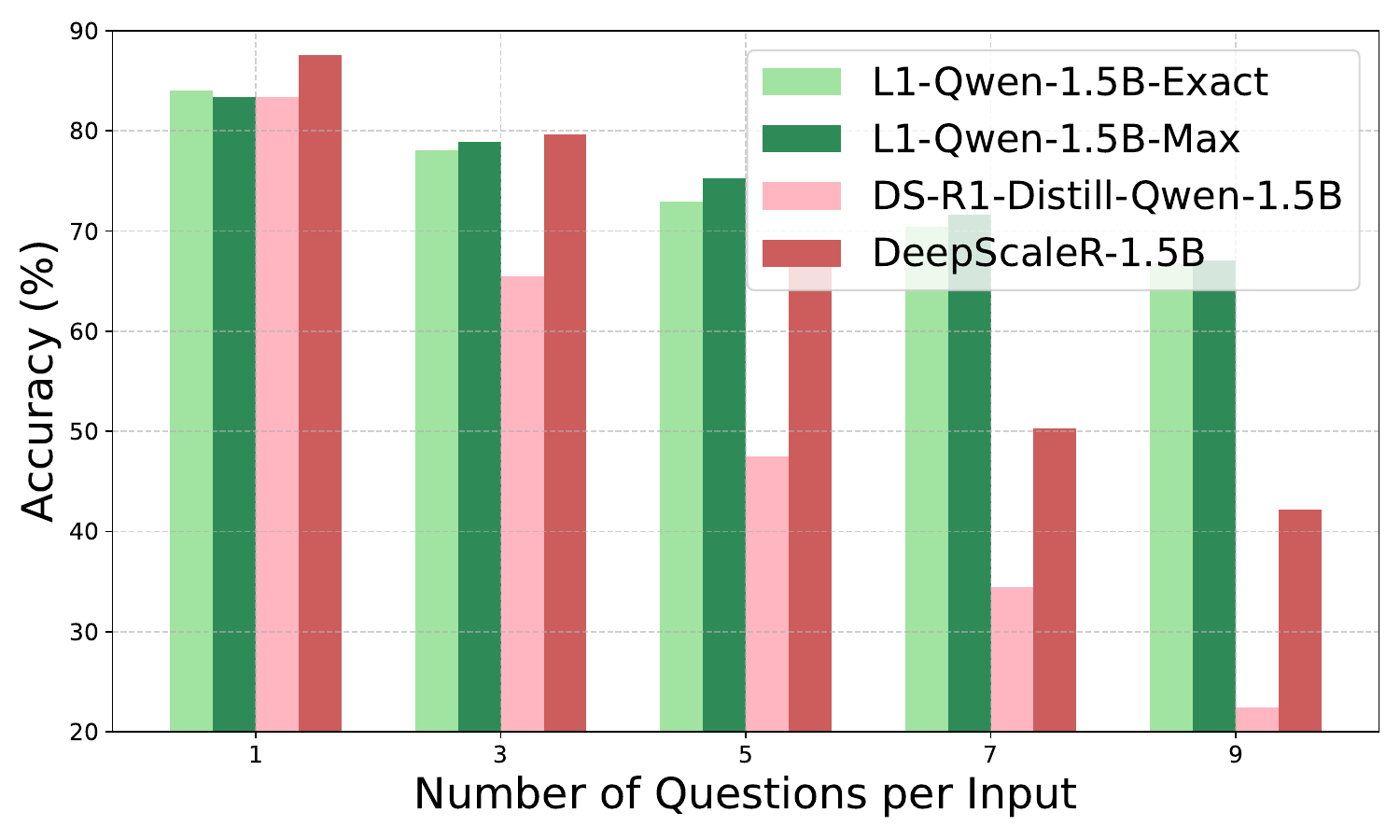}
    \vspace{-6mm}
    \caption{1.5B Model on MATH500}
    \label{fig:bar_math500_1p5b_long2short-acc}
    \end{subfigure}
    \hfill
    \begin{subfigure}{0.3\textwidth}
    \includegraphics[width=\linewidth]{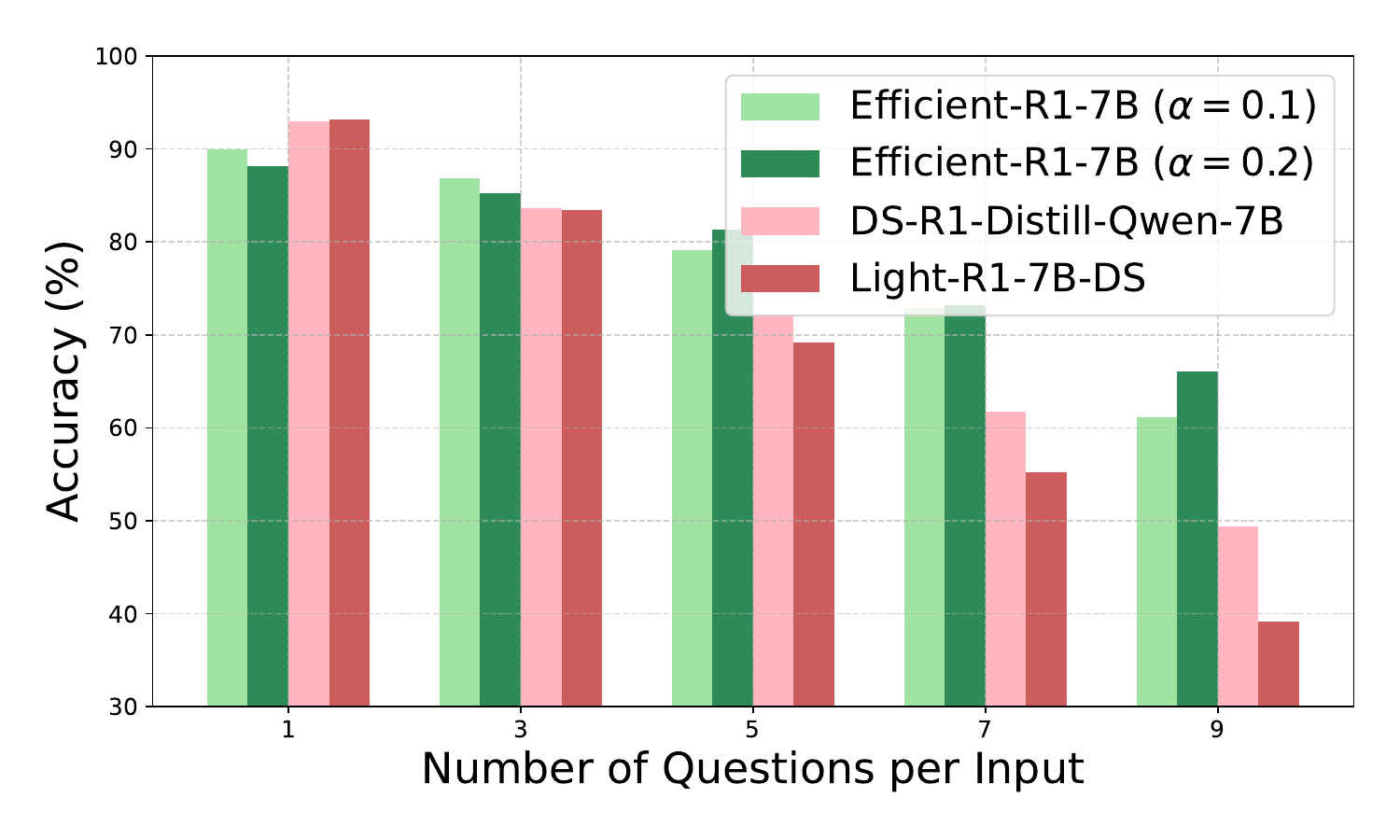}
    \vspace{-6mm}
    \caption{7B Model on MATH500}
    \label{fig:bar_math500_7b_long2short-acc}
    \end{subfigure}
    \hfill
    \begin{subfigure}{0.3\textwidth}
    \includegraphics[width=\linewidth]{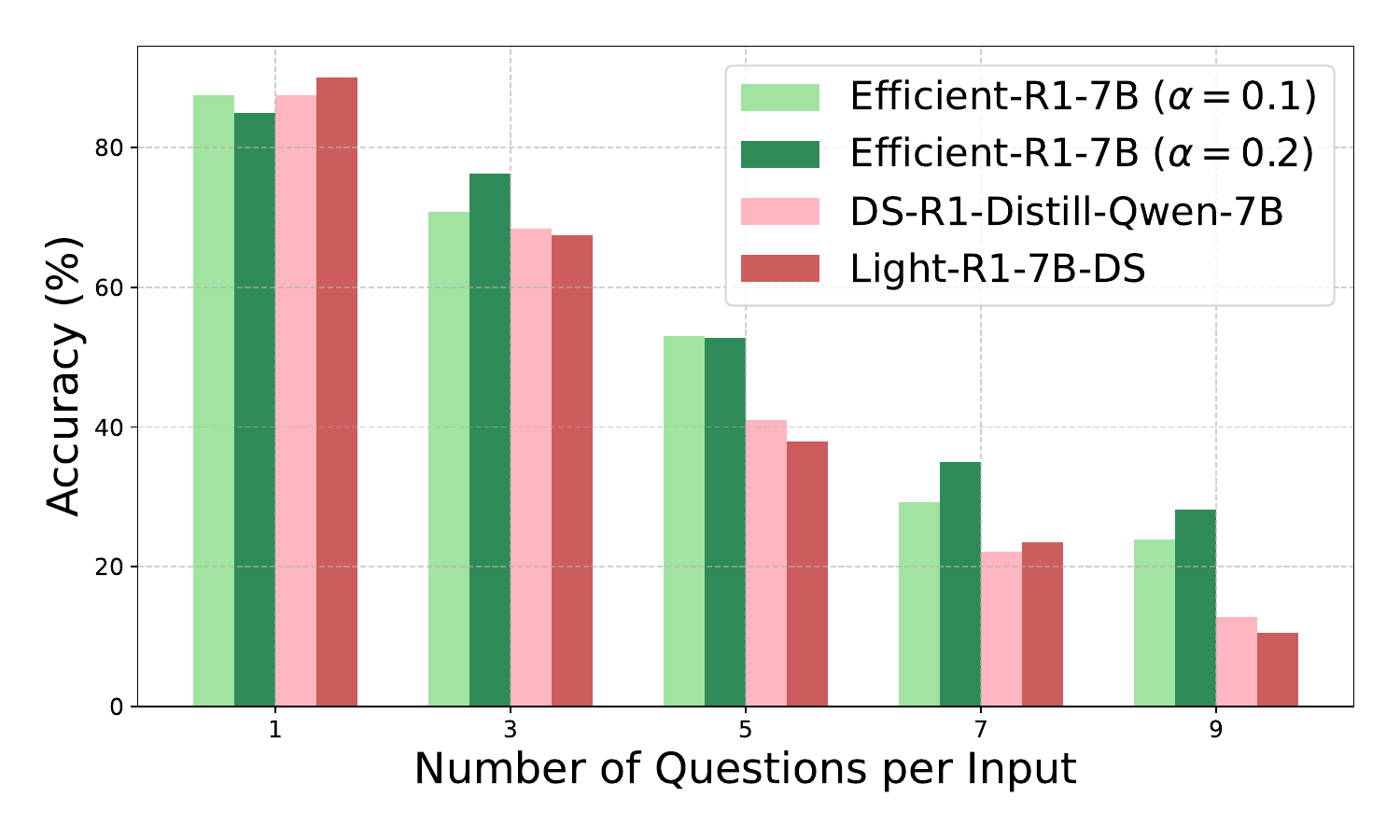}
    \vspace{-6mm}
    \caption{7B Model on AMC23}
    \label{fig:bar_amc23_7b_long2short-acc}
    \end{subfigure}
    \caption{
    The effect of Long2Short training on \sysname. 
    Long2Short training mitigates the performance degradation under higher stress levels.
    }
    \label{fig:long2short}
    \vspace{-3mm}
\end{figure*}

\paragraph{Potential of Long2Short Training.}
Despite their impressive performance on reasoning tasks, LRMs often struggle with excessively verbose reasoning processes~\citep{sui2025stop,chen2024not,wang2025thoughts}. To address this issue, recent studies~\citep{aggarwal2025l1,arora2025training,team2025kimi} propose the Long2Short technique, which introduces a length reward in the training objective to promote more concise and efficient reasoning.
Our experiments demonstrate that Long2Short training significantly enhances reasoning robustness under \sysname. As illustrated in Fig.~\ref{fig:long2short}, \begin{wraptable}{r}{0.5\linewidth}
    \centering
    \def\arraystretch{1,2}
    \vspace{-3mm}
    \caption{Evaluation results on \textit{GPQA Diamond} and \textit{LiveCodeBench}. The ``Stress'' column reports the average performance across four distinct stress levels, as described in Sec.~\ref{sec:evaluation_setup}. Detailed accuracies for each stress level are in Tab.~\ref{tab:main_detail1} of the Appendix. Models perform best under ``Stress'' are highlighted in \textcolor{gray}{gray}.}
    \label{tab:main_other}
    \resizebox{1.0\linewidth}{!}{%
    \begin{tabular}{c|c>{\cellcolor{cyan!5}}c|c>{\cellcolor{cyan!5}}c}
    \toprule[0.8pt]
       \multirow{2}{*}{\diagbox{Model}{Bench}} & \multicolumn{2}{c}{GQPA Diamond} & \multicolumn{2}{c}{LiveCodeBench} \\
       & Single & Stress & Single & Stress \\
       \hline
       \rowcolor{blue!15}
       \multicolumn{5}{c}{\textit{1.5B LRMs}} \\
       DS-R1-Distill-Qwen-1.5B~\citep{guo2025deepseek} & \textbf{37.37} & 22.11 & 15.05 & 0.48 \\
       DeepScaleR-1.5B~\citep{deepscaler2025} & 31.82 & 27.90 & \textbf{21.15} & 1.83 \\
       L1-Qwen-1.5B-Exact~\citep{aggarwal2025l1} & 33.84 & 31.01 & 18.28 & 2.70 \\
       \cellcolor{lightgray}L1-Qwen-1.5B-Max~\citep{aggarwal2025l1} & 36.87 & \cellcolor{lightgray}\textbf{32.03} & 19.35 & \cellcolor{lightgray}\textbf{2.45} \\
       Qwen2.5-1.5B-Inst~\citep{yang2024qwen25} & 26.26 & 21.52 & 1.79 & 0.37 \\
       \rowcolor{blue!15}
       \multicolumn{5}{c}{\textit{$7\sim8$B LRMs}} \\
       DS-R1-Distill-Qwen-7B~\citep{guo2025deepseek} & \textbf{51.01} & 31.67 & 37.63 & 2.89 \\
       DS-R1-Distill-LLaMA-8B~\citep{guo2025deepseek} & 50.00 & 33.99 & 39.43 & 11.25 \\
       Efficient-R1-7B ($\alpha=0.2$)~\citep{arora2025training} & 47.97 & 34.37 & 38.71 & 2.50 \\
       Nemotron-Nano-8B~\citep{bercovich2025llamanemotronefficientreasoningmodels} & \textbf{51.01} & 34.06 & \textbf{50.90} & 8.56 \\
       AReaL-boba-RL-7B~\citep{mei2025real} & 48.98 & 29.13 & 37.99 & 4.73 \\
       Light-R1-7B-DS~\citep{wen2025light} & 41.91 & 30.32 & 39.07 & 2.85 \\
       OpenR1-Qwen-7B~\citep{openr1} & 38.38 & 36.04 & 4.66 & 1.27 \\
       \cellcolor{lightgray}OpenThinker2-7B~\citep{OpenThoughts2025} & 49.49 & \cellcolor{lightgray}\textbf{40.60} & 39.43 & \cellcolor{lightgray}\textbf{14.51} \\
       SimpleRL-Zoo-Qwen-7B~\citep{zeng2025simplerlzooinvestigatingtamingzero} & 33.84 & 35.49 & 5.73 & 0.16 \\
       Open-Reasoner-Zero-7B~\citep{hu2025open} & 37.37 & 34.25 & 16.13 & 0.64 \\
       Marco-O1-7B~\citep{zhao2024marcoo1openreasoningmodels} & 30.81 & 28.32 & 9.32 & 7.24 \\
       Qwen2.5-7B-Inst~\citep{yang2024qwen25} & 35.86 & 35.15 & 13.98 & 10.19 \\
       \rowcolor{blue!15}
       \multicolumn{5}{c}{\textit{32B LRMs}} \\
       DS-R1-Distill-Qwen-32B~\citep{guo2025deepseek} & 60.10 & 53.73 & 55.56 & 26.71 \\
       \cellcolor{lightgray}Qwen-QwQ-32B~\cite{qwq32b} & 63.64 & \cellcolor{lightgray}\textbf{60.03} & \textbf{62.37} & 32.16 \\
       AReaL-boba-SFT-32B~\citep{mei2025real} & 63.13 & 50.59 & 60.93 & 31.74 \\
       Light-R1-32B-DS~\citep{wen2025light} & \textbf{65.66} & 50.11 & 60.93 & 28.22 \\ 
       S1.1-32B~\citep{muennighoff2025s1simpletesttimescaling} & 61.62 & 54.54 & 25.45 & 24.46 \\
       \cellcolor{lightgray}OpenThinker2-32B~\citep{OpenThoughts2025} & 62.12 & 57.79 & 56.27 & \cellcolor{lightgray}\textbf{37.53} \\
       SimpleRL-Zoo-Qwen-32B~\citep{zeng2025simplerlzooinvestigatingtamingzero} & 46.46 & 46.21 & 26.52 & 23.95 \\
       Open-Reasoner-Zero-32B~\citep{hu2025open} & 60.10 & 49.57 & 35.13 & 13.01 \\
       Qwen2.5-32B-Inst~\citep{yang2024qwen25} & 42.93 & 40.04 & 26.88 & 24.00 \\
       \rowcolor{blue!15}
       \multicolumn{5}{c}{\textit{API-Based LRMs}} \\
       DeepSeek-R1~\citep{guo2025deepseek} & 70.20 & 64.63 & 63.44 & 40.83 \\
       O3-mini~\citep{openai2025o3mini} & 71.21 & 67.39 & 60.21 & 48.36 \\
       \cellcolor{lightgray}O4-mini~\citep{openai2025o3mini} & 76.26 & \cellcolor{lightgray}\bf 73.11 & \bf 70.61 & \cellcolor{lightgray}\bf 63.07 \\
       Gemini-2.5-Flash-Thinking~\citep{team2023gemini} & \bf 78.79 & 68.00 & 61.65 & 48.34 \\
    \bottomrule[0.8pt]
    \end{tabular}
    }
\vspace{-20mm}
\end{wraptable}models trained with Long2Short, such as L1-Qwen-1.5B-Exact and L1-Qwen-1.5B-Max~\citep{aggarwal2025l1}, exhibit notable performance advantages at high stress levels. For instance, as shown in Tab.~\ref{tab:main_detail1} of appendix, L1-Qwen-1.5B-Max surpasses R1-1.5B by a substantial 44.71\% accuracy margin at stress level $s = 9$ on MATH500. A similar trend is observed among 7B models. For example, the Efficient-R1-7B ($\alpha = 0.2$) model~\citep{arora2025training}, despite having a lower single-question accuracy (88.20\%) compared to R1-7B (93.00\%), achieves 66.02\% accuracy at stress level $s = 9$, outperforming the R1-7B's 49.42\% by a significant 16.60\% accuracy margin.
\section{Further Analysis}

\subsection{Error Analysis}

\paragraph{Maximum output length overflow is not the only cause of performance degradation.}
To uncover the underlying reasons for LRMs' failures in \sysname, we analyze the primary factors contributing to performance degradation under stress conditions. A straightforward hypothesis is that performance drops primarily due to models exceeding their maximum output length when handling multiple problems. However, our findings indicate that this is not the sole cause—\textit{models frequently fail even when their responses remain well within the maximum output token limit.}

\begin{figure*}[t]
    \centering
    \hfill
    \begin{subfigure}{0.24\textwidth}
    \includegraphics[width=\linewidth]{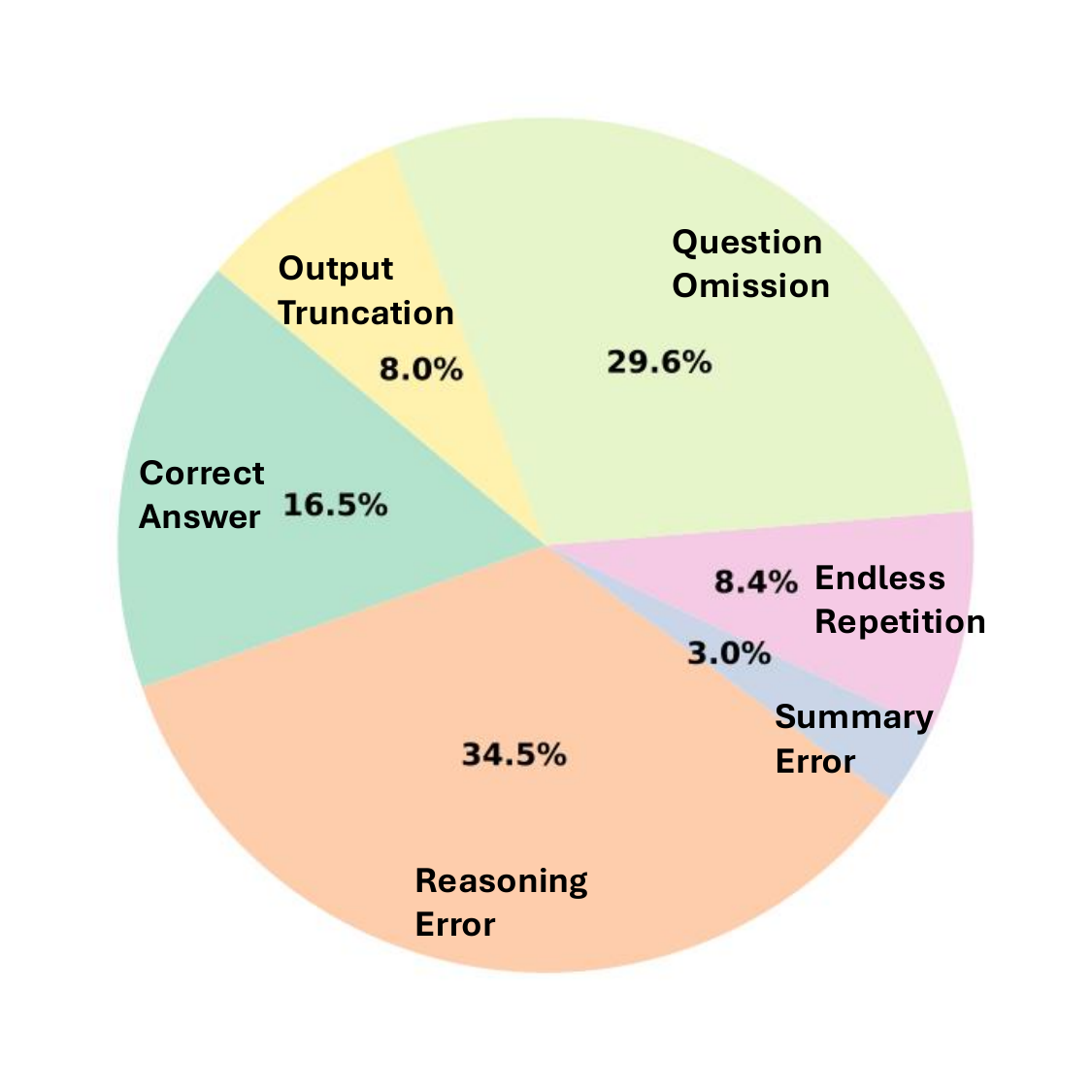}
    \vspace{-6mm}
    \caption{R1-7B~\citep{guo2025deepseek}}
    \label{fig:error_type_r1-7b}
    \end{subfigure}
    \hfill
    \begin{subfigure}{0.24\textwidth}
    \includegraphics[width=\linewidth]{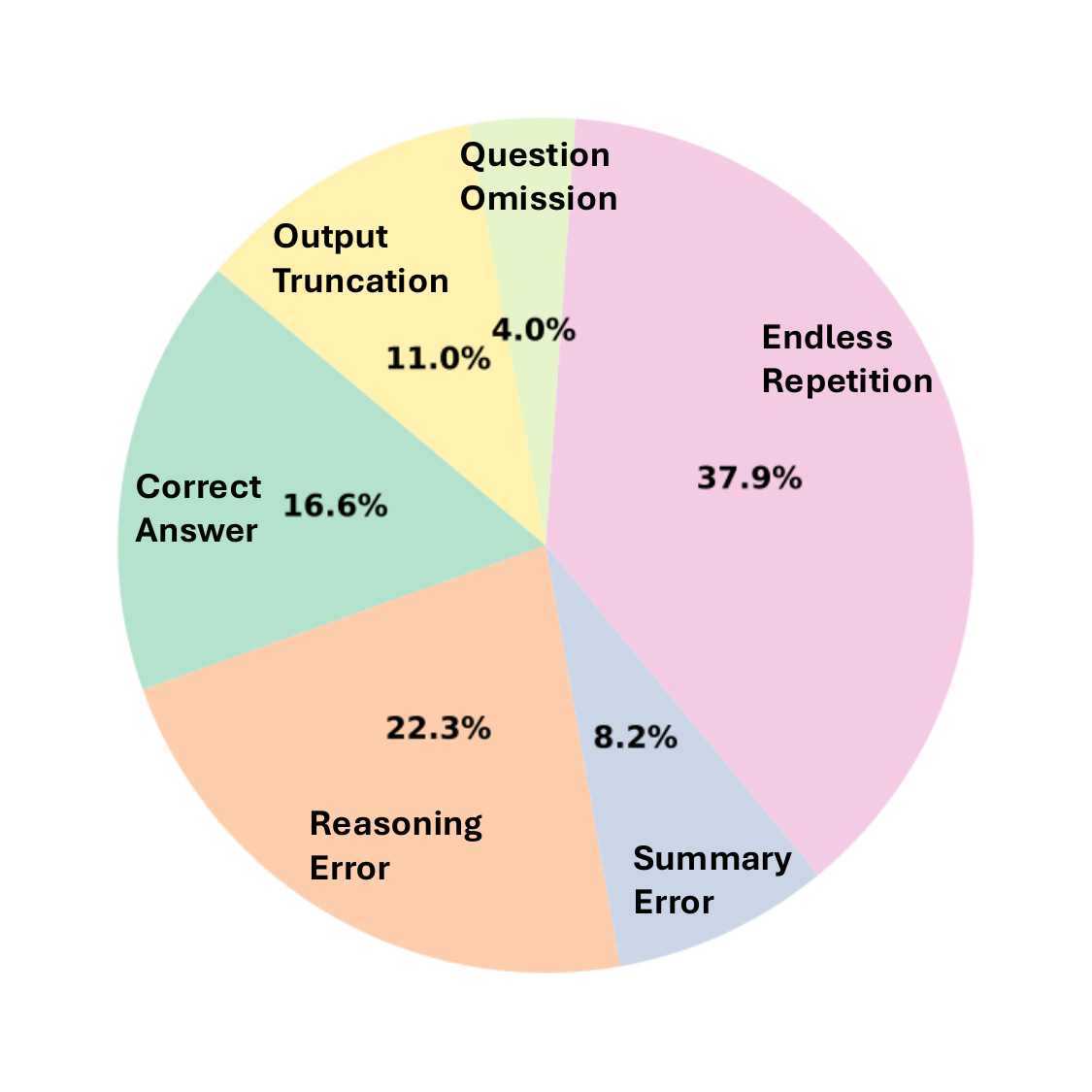}
    \vspace{-6mm}
    \caption{Light-R1-7B~\citep{wen2025light}}
    \label{fig:error_type_light-r1}
    \end{subfigure}
    \hfill
    \begin{subfigure}{0.24\textwidth}
    \includegraphics[width=\linewidth]{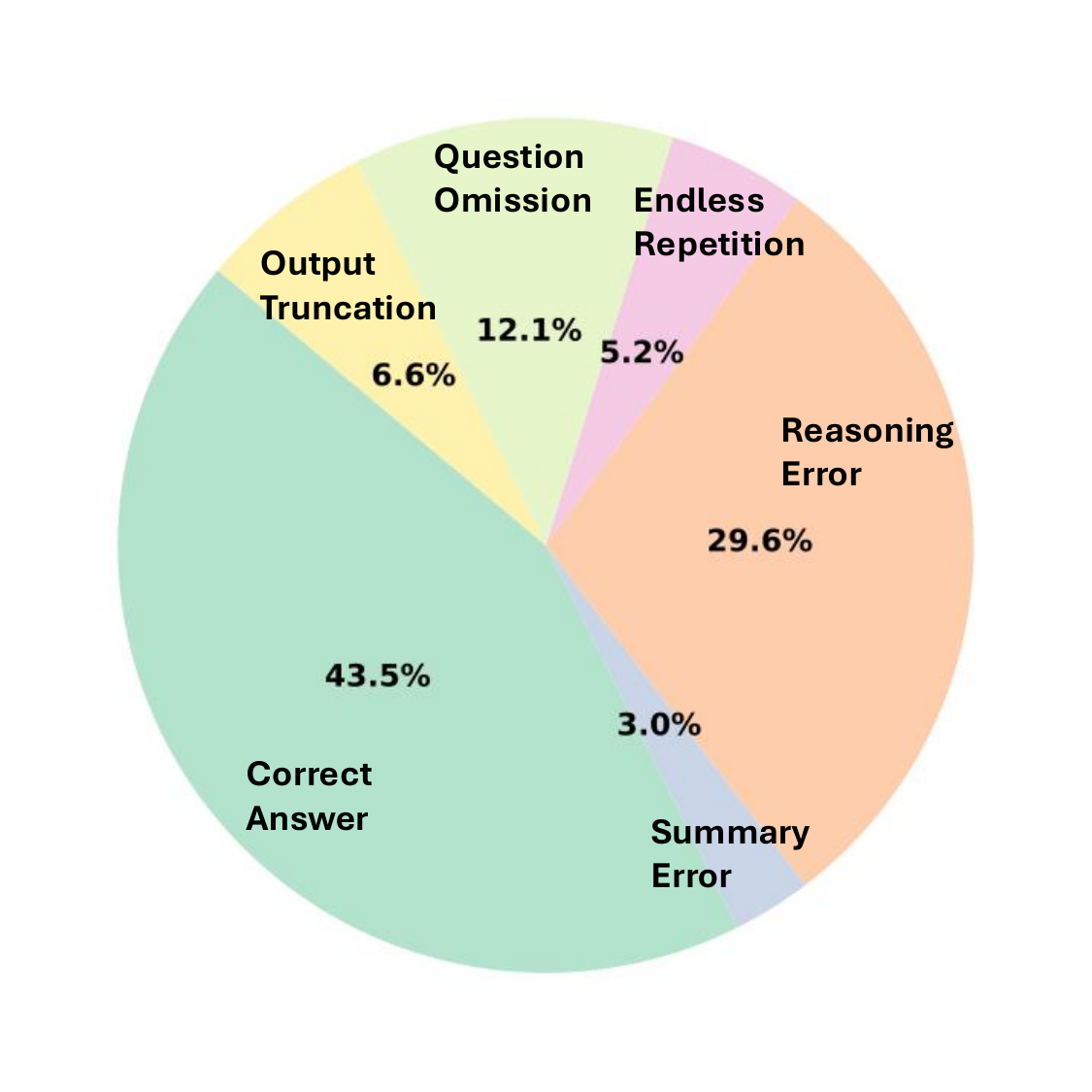}
    \vspace{-6mm}
    \caption{Nemotron-7B~\citep{bercovich2025llamanemotronefficientreasoningmodels}}
    \label{fig:error_type_nemotron}
    \end{subfigure}
    \hfill
    \begin{subfigure}{0.24\textwidth}
    \includegraphics[width=\linewidth]{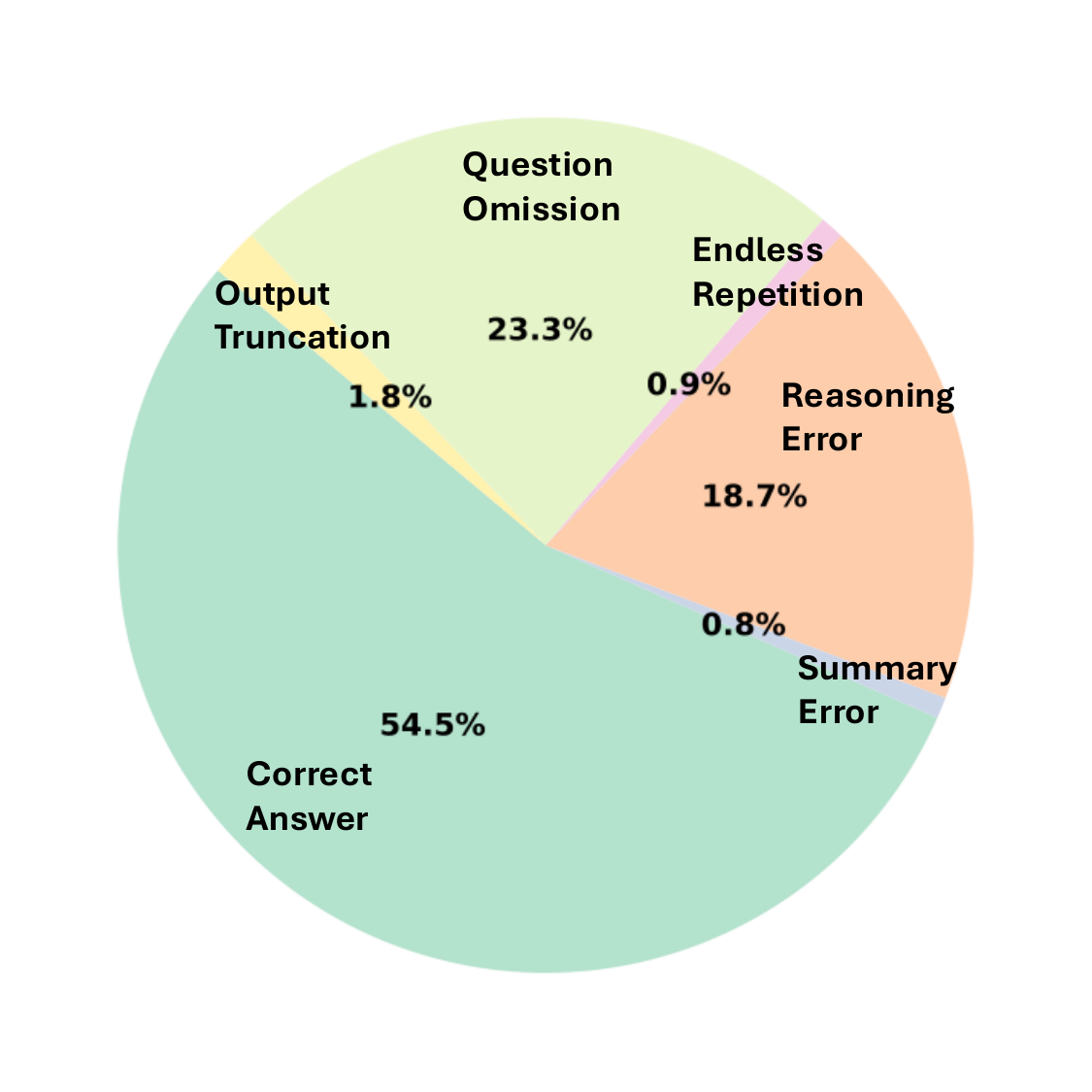}
    \vspace{-6mm}
    \caption{DeepSeek-R1~\citep{guo2025deepseek}}
    \label{fig:error_type_r1}
    \end{subfigure}
    \hfill
    \caption{
    The error type distribution for various LRMs on AIME24 under \sysname. 
    }
    \label{fig:error_type}
\end{figure*}

To investigate the cause of errors, we categorize the observed mistakes into six distinct types, as summarized in Tab.~\ref{tab:error_taxonomy}. A detailed distribution of these error types for four representative LRMs on the AIME24 benchmark is illustrated in Fig.~\ref{fig:error_type}. The analysis reveals that Question Omission (QO) and Reasoning Error (RE) are the dominant error types for DeepSeek-R1 (in subfigure~\ref{fig:error_type_r1}), collectively accounting for a substantial 42.00\% of the performance degradation under stress test. The high frequency of Question Omission (QO) suggests that DeepSeek-R1 suffers from a pronounced question position bias, often focusing disproportionately on the first question in multi-problem prompts, despite explicit instructions to address each question sequentially.
In contrast, error causes like Summary Error (SE), Output Truncation (OT), and Endless Repetition (ER) are rare in such high-capacity models.

\begin{table}[t]
    \centering
    \vspace{-3mm}
    \caption{Six common cause of the performance degradation in \sysname.}
    \resizebox{\textwidth}{!}{%
    \begin{tabular}{l|l}
        \toprule[0.15em]
        \textbf{Error Cause} & \textbf{Definition} \\
        \midrule
        \textbf{Output Truncation (OT)} & The output is trucated due to reaching the maximum output length (32K tokens for LRMs). \\
        \textbf{Endless Repetition (ER)} & The model repeatedly generates the same phrase or sentence in its output. \\
        \textbf{Format Violation (FV)} & The model fails to comply with the required output format (\textit{e.g.,} not putting the answer whtin ``\textbackslash boxed\{\}''). \\
        \textbf{Question Omission (QO)} & The model fails to respond to all the given questions in the thinking process. (\textit{e.g.,} only answers the first question) \\
        \textbf{Summary Error (SE)} & The model fails to summarize all the answers generated in the thinking process. \\
        \textbf{Reasoning Error (RE)} & Error occurs in the reasoning process, such as calculation mistakes or concept misunderstandings. \\
    
        \bottomrule[0.15em]
    \end{tabular}%
    }
    \label{tab:error_taxonomy}
    \vspace{-2mm}
\end{table}

However, for smaller models like Light-R1-7B (in subfigure~\ref{fig:error_type_light-r1}), the error profile shifts significantly, with Endless Repetition (ER) and Output Truncation (OT) becoming more prominent, collectively accounting for 48.90\%. This shift underscores smaller models' unique challenges, whose limited capacity significantly exacerbates performance degradation under the stress imposed by \sysname.
Finally, it is important to note that no Format Violation (FV) errors were found for any of the four models, which is why they are not included in Fig.~\ref{fig:error_type}. This observation supports the robustness of our prompt design and answer extraction strategy, confirming that \textit{the observed performance degradation arises primarily from the intrinsic limitations of the LRMs themselves, rather than artifacts introduced by the evaluation process.}

\begin{figure*}[!b]
    \centering
    \hfill
    \begin{subfigure}{0.3\textwidth}
    \includegraphics[width=\linewidth]{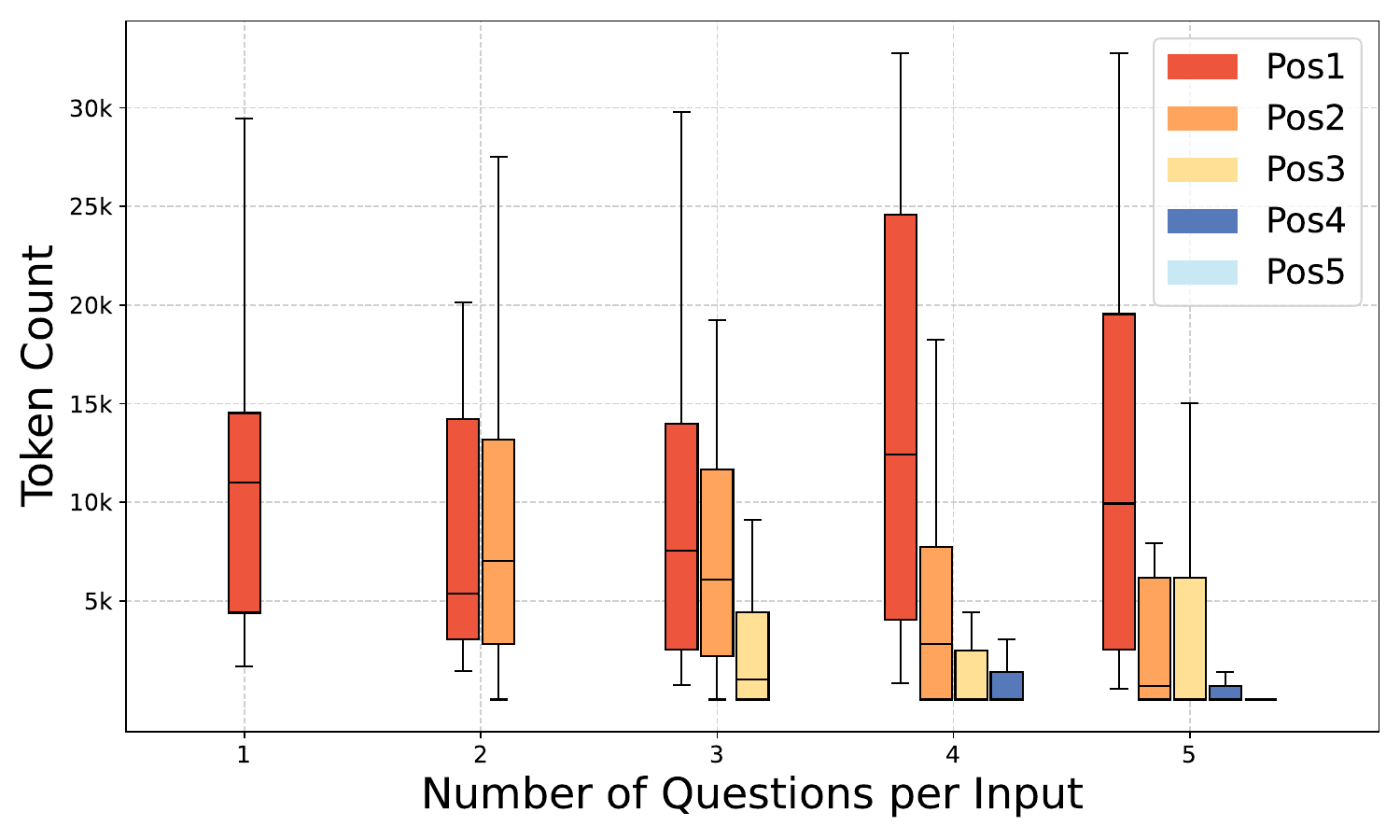}
    \vspace{-6mm}
    \caption{DS-R1-Distill-Qwen-7B~\citep{guo2025deepseek}}
    \label{fig:ds-r1-distill-qwen-7b}
    \end{subfigure}
    \hfill
    \begin{subfigure}{0.3\textwidth}
    \includegraphics[width=\linewidth]{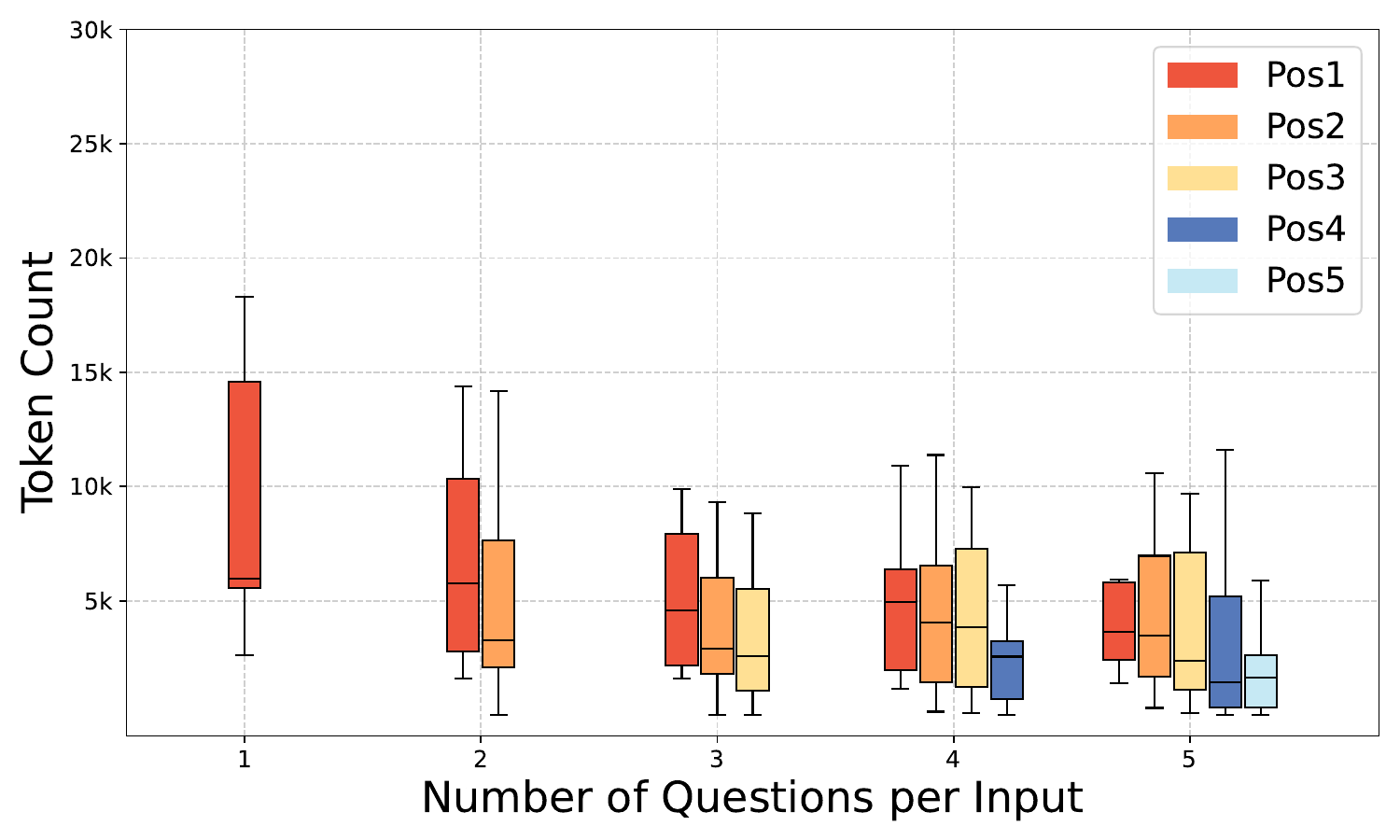}
    \vspace{-6mm}
    \caption{Nemotron-nano-7B~\citep{bercovich2025llamanemotronefficientreasoningmodels}}
    \label{fig:nemotron-nano-7b}
    \end{subfigure}
    \hfill
    \begin{subfigure}{0.3\textwidth}
    \includegraphics[width=\linewidth]{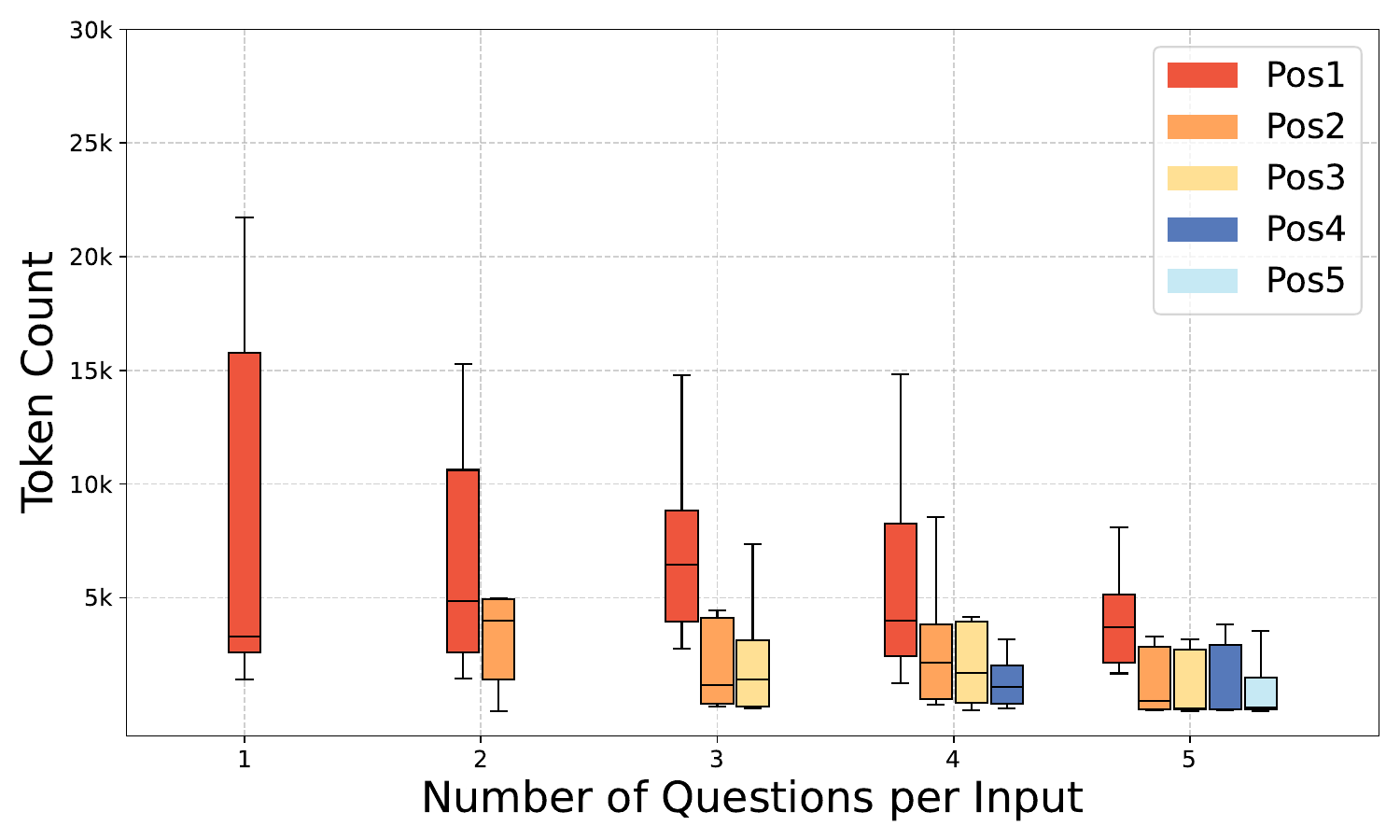}
    \vspace{-6mm}
    \caption{DeepSeek-R1~\citep{guo2025deepseek}}
    \label{fig:deepseek-r1}
    \end{subfigure}
    \hfill
    \vspace{1mm}
    \caption{
    The reasoning token count for questions at different positions on AIME24.
    }
    \label{fig:boxplot_token}
\end{figure*}
\paragraph{LRMs that perform better on \sysname\ exhibit concise reasoning for earlier questions.}
\label{sec:reason_behind_rest}
In this paragraph, we investigate why certain models exhibit substantial performance differences under \sysname, despite achieving similar results on traditional single-question benchmarks. 
To analyze this, on AIME24, we manually split the entire response into chunks corresponding to each question and calculate the reasoning token count for each chunk. The results are presented in Fig.~\ref{fig:boxplot_token}. It is evident that high-performing models under \sysname\ (e.g., \textit{Nemotron-nano-7B} and \textit{DeepSeek-R1}) tend to use fewer reasoning tokens for the first question when the stress level exceeds 1. In contrast, models with lower \sysname\ performance, such as \textit{R1-7B}, often overthink the first question, consuming a substantial portion of their reasoning tokens and leaving insufficient space for subsequent questions.
This observation reveals that \textit{LRMs with superior performance in \sysname\ tend to employ more concise reasoning for earlier questions under stress}, leaving sufficient space to address subsequent questions.
We refer to this ability as ``adaptive reasoning effort allocation,'' which we believe is a critical factor for achieving robust performance under \sysname.

\subsection{The Impact of Question Position and Question Order}
\begin{figure*}[!h]
    \centering
    \begin{subfigure}{0.24\textwidth}
    \includegraphics[width=\linewidth]{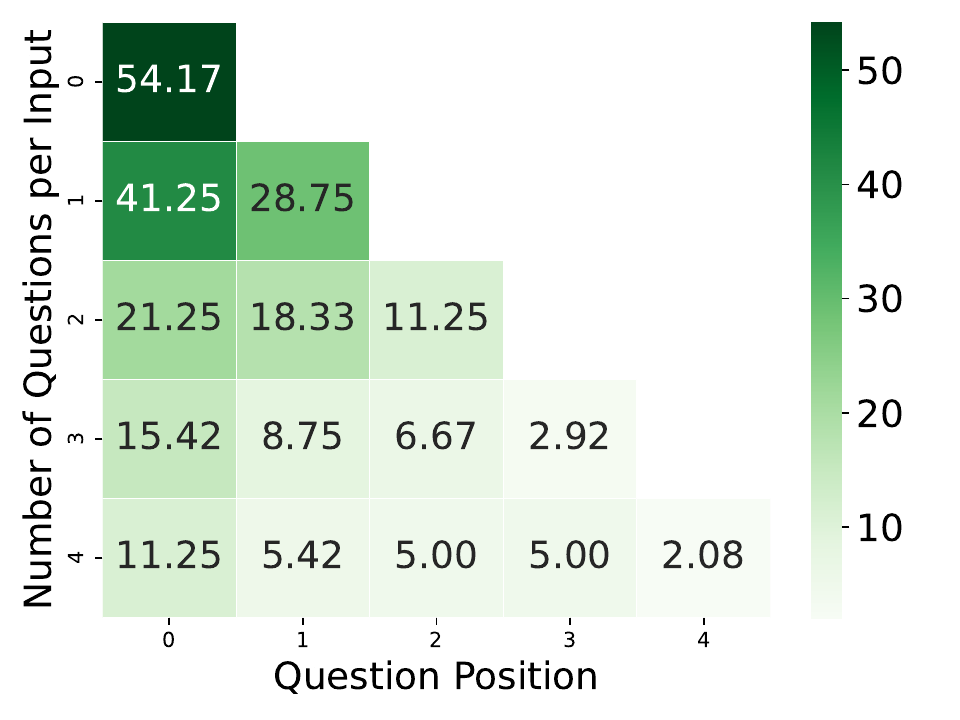}
    \vspace{-6mm}
    \caption{7B on AIME24}
    \label{fig:heatmap_aime24_7b_position}
    \end{subfigure}
    \hfill
    \begin{subfigure}{0.24\textwidth}
    \includegraphics[width=\linewidth]{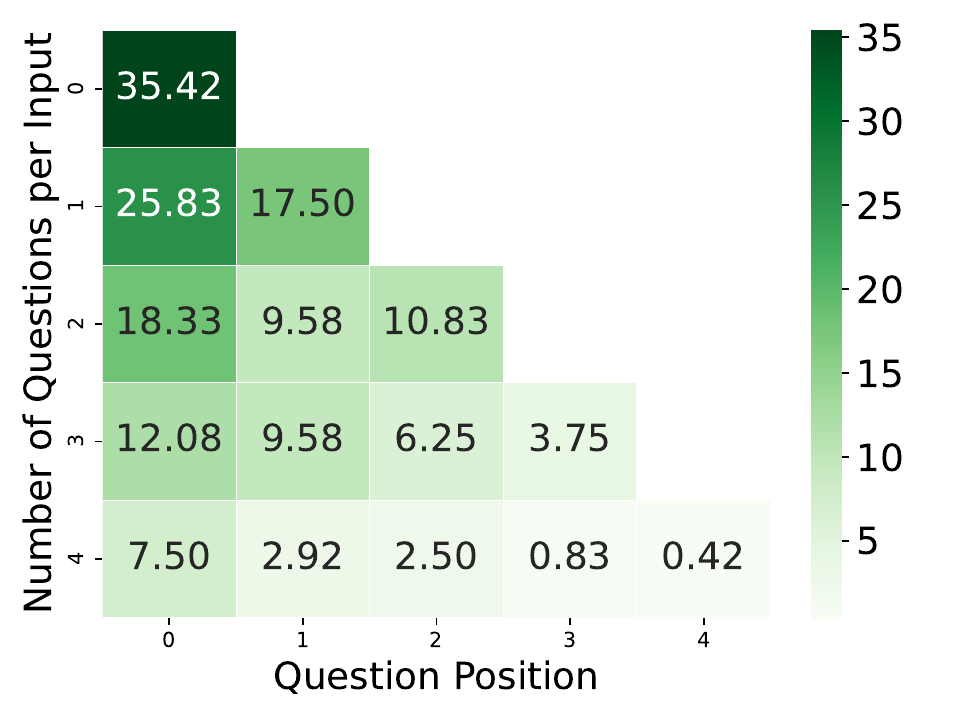}
    \vspace{-6mm}
    \caption{7B on AIME25}
    \label{fig:heatmap_aime25_7b_position}
    \end{subfigure}
    \hfill
    \begin{subfigure}{0.24\textwidth}
    \includegraphics[width=\linewidth]{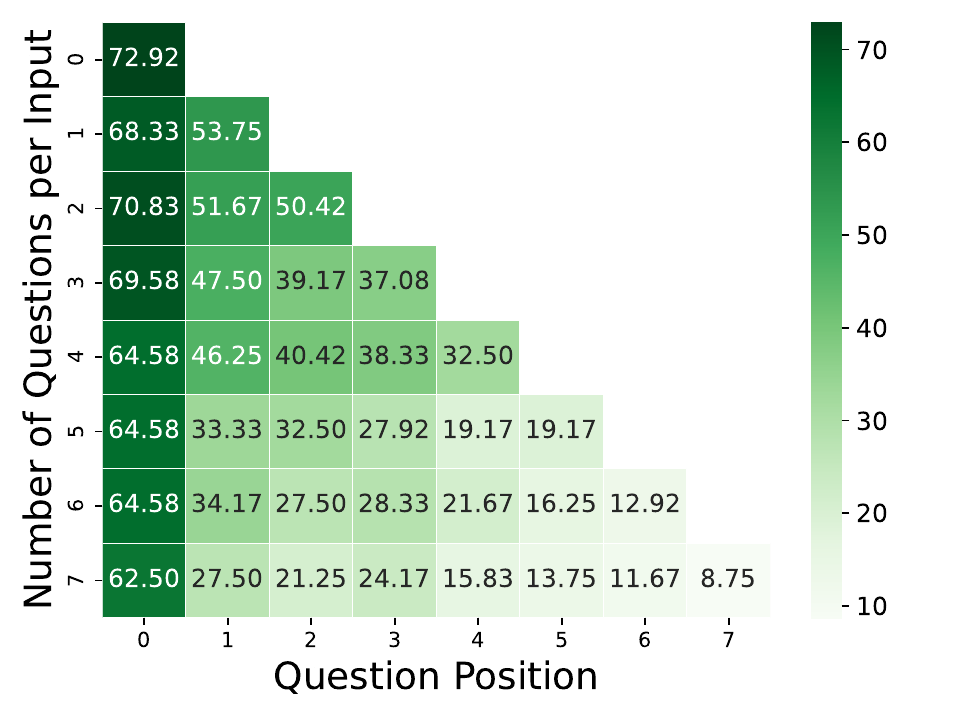}
    \vspace{-6mm}
    \caption{32B on AIME24}
    \label{fig:heatmap_aime25_32b_sorted}
    \end{subfigure}
    \hfill
    \begin{subfigure}{0.24\textwidth}
    \includegraphics[width=\linewidth]{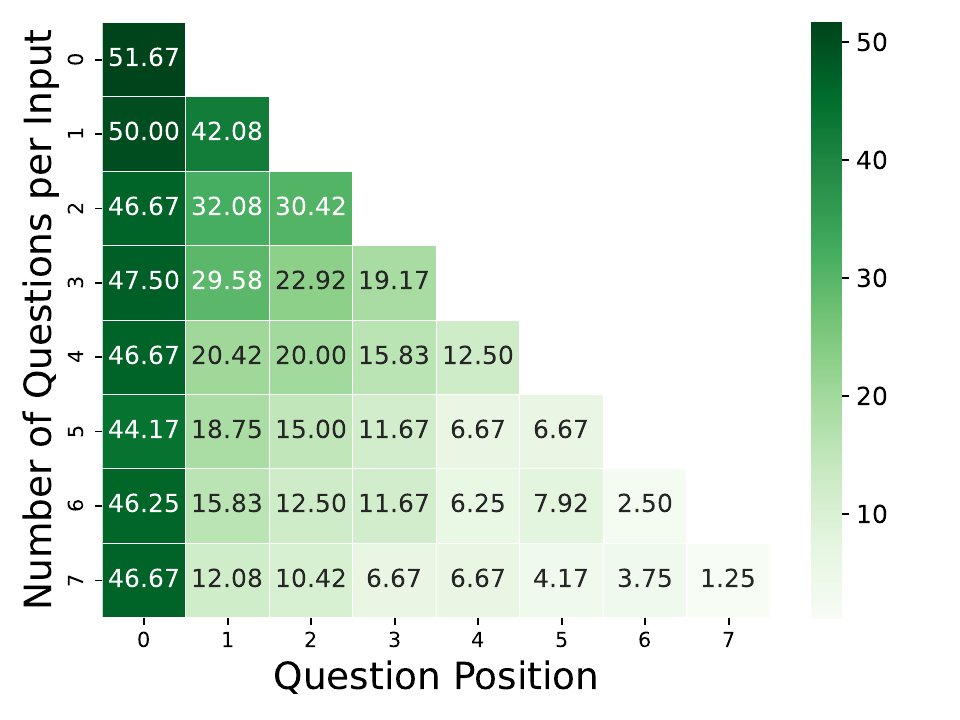}
    \vspace{-6mm}
    \caption{32B on AIME25}
    \label{fig:heatmap_aime25_32b_sorted}
    \end{subfigure}
    \vspace{1mm}
    \caption{
The effect of question position under stress tests. The performance at each position is calculated by averaging the accuracy of questions appearing at the same position across the full benchmark. LRMs generally achieve higher accuracy for earlier questions, while their performance declines for subsequent ones.
    }
    \label{fig:heatmap_acc_position}
\end{figure*}

\paragraph{Question position affects performance under stress test.}
Recent studies~\citep{yu2024mitigate,liu2024lost} have identified positional biases of LLMs over long inputs. In this paragraph, we investigate whether the position of questions within the input can influence response accuracy. Fig.~\ref{fig:heatmap_acc_position} presents a heatmap illustrating the accuracy for questions placed at different positions. The results indicate that \textbf{\textit{questions appearing earlier consistently achieve higher accuracy}}, even under the highest stress levels, while the accuracy for later questions tends to decline as the stress level increases.
This observation contrasts with earlier findings~\citep{wang2024evaluating}, which suggest that question position has a negligible effect on performance because previous analyses primarily focused on simpler text classification tasks. In contrast, complex reasoning tasks exhibit a stronger positional dependence.
We attribute this position-dependent performance degradation to three primary factors. First, our error analysis reveals that Question Omission (QO) is a prevalent failure mode, even for SOTA LRMs like DeepSeek-R1,
leading to high accuracy for the first question but significantly lower accuracy for the later ones. 
Second, the extensive reasoning required for earlier questions can introduce substantial noise, potentially disrupting the logical processing needed for subsequent responses. Finally, as shown in Fig.~\ref{fig:boxplot_token}, answering earlier questions may consume a considerable portion of the output tokens, reducing the model's available reasoning tokens for later questions.

\begin{figure*}[!t]
    \centering
    \begin{subfigure}{0.3\textwidth}
    \includegraphics[width=\linewidth]{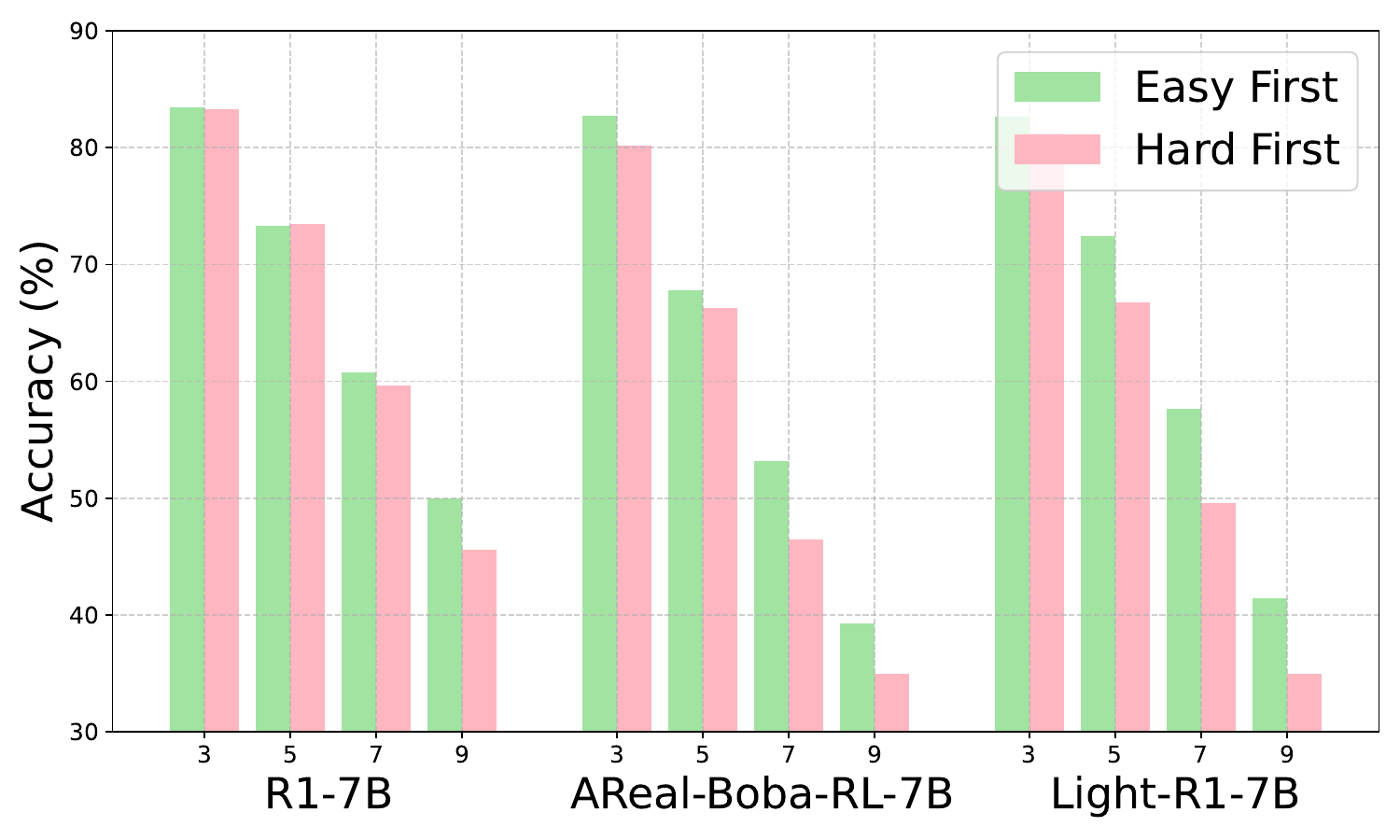}
    \vspace{-6mm}
    \caption{MATH500}
    \label{fig:bar_math500_sorted}
    \end{subfigure}
    \hfill
    \begin{subfigure}{0.3\textwidth}
    \includegraphics[width=\linewidth]{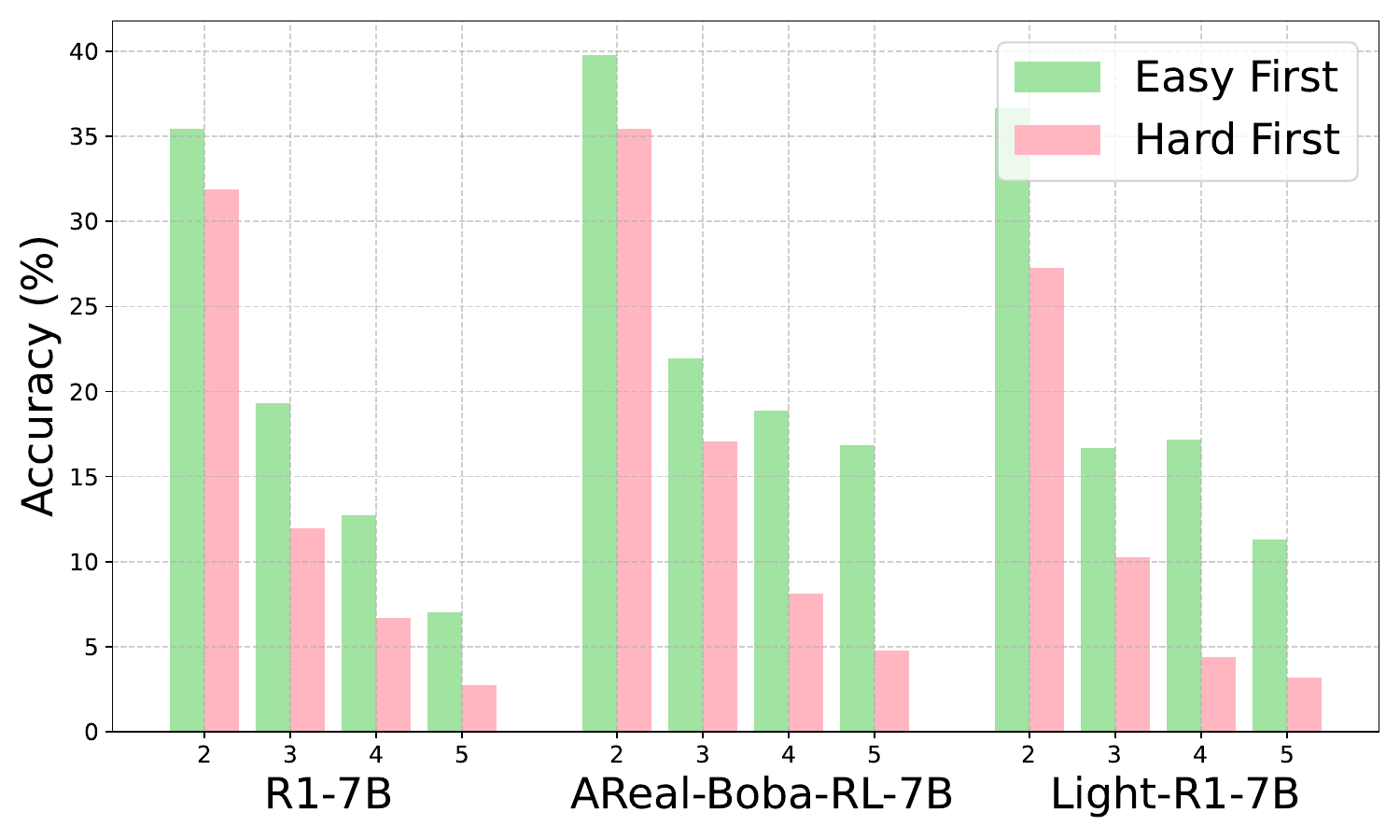}
    \vspace{-6mm}
    \caption{AIME24}
    \label{fig:bar_aime24_sorted}
    \end{subfigure}
    \hfill
    \begin{subfigure}{0.3\textwidth}
    \includegraphics[width=\linewidth]{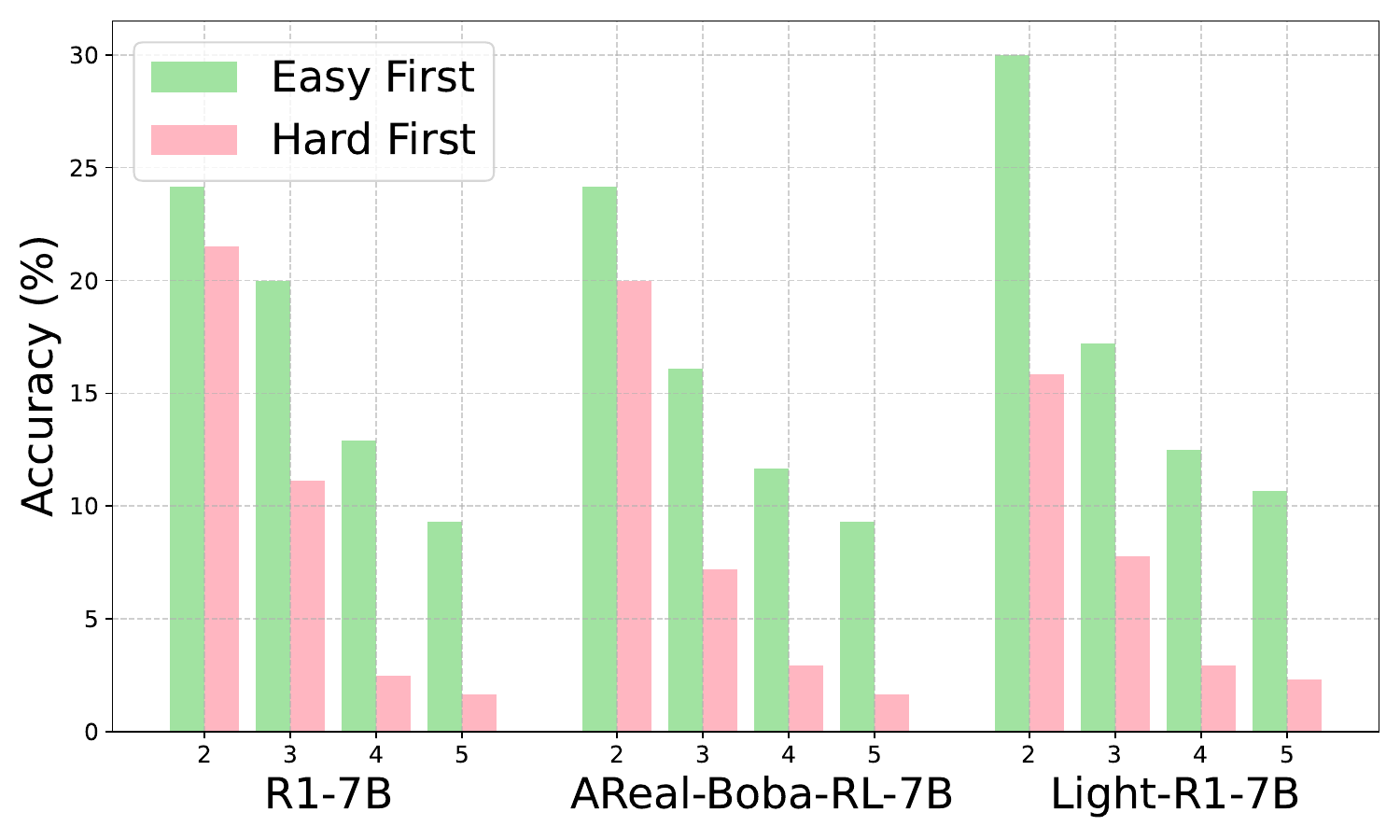}
    \vspace{-6mm}
    \caption{AIME25}
    \label{fig:bar_aime25_sorted}
    \end{subfigure}
    \vspace{1mm}
    \caption{
The effect of question order on overall performance under stress tests. LRMs consistently perform worse when problems are arranged from hard to easy.
    }
    \label{fig:bar_acc_sorted}
\end{figure*}

\paragraph{Question order affects performance under stress test.}
Building on the observation that question position impacts performance, we further investigate the effect of question order within each prompt on overall benchmark accuracy. Specifically, we assess this by arranging the questions within each instruction in either descending (hard-first) or ascending (easy-last) order of difficulty, using the fail rate of \textit{R1-7B} as a proxy for difficulty for problems in \textit{AIME24} and \textit{AIME25}.
As shown in Fig.~\ref{fig:bar_acc_sorted}, \textit{presenting questions from easy to hard consistently yields better overall accuracy compared to the reverse order.} This discrepancy can be attributed to the tendency of LRMs to engage in lengthy and redundant reasoning processes when confronted with difficult questions early in the input. In contrast, placing simpler questions first encourages more concise and straightforward reasoning chains, thereby minimizing the cognitive load and reducing the interference with subsequent questions.

\section{Conclusion}

In conclusion, we present \sysname, a stress testing framework designed to evaluate Large Reasoning Models (LRMs) in simultaneous multi-problem solving scenarios. \sysname\ significantly enhances the discriminative power of existing benchmarks that often suffer from performance saturation, revealing substantial performance differences among models that appear comparable in traditional single-question evaluations. Notably, even SOTA models like DeepSeek-R1 exhibit marked performance degradation under stress testing, challenging the common assumption that ``LLMs are multi-problem solvers.''
Additionally, \sysname\ uncovers a few reasoning misbehaviors, such as question omission and summary error, which remain obscure in single-question assessments. Critical findings, such as the identification of question position bias and the demonstrated effectiveness of long2short training, offer valuable insights for the future development of more robust and capable LRMs.


\clearpage
\newpage

\clearpage
\newpage
\beginappendix
\section{Detailed Evaluation Result}
\label{sec:detailed_result}

We present the detailed accuracy statistics for each stress level of various LRMs in Table~\ref{tab:main_detail1}, Table~\ref{tab:main_detail2} and Table~\ref{tab:main_detail3}. The stress degree $\{\uppercase\expandafter{\romannumeral1}, \uppercase\expandafter{\romannumeral2}, \uppercase\expandafter{\romannumeral3}, \uppercase\expandafter{\romannumeral4}, \uppercase\expandafter{\romannumeral5}\}$ corresponds to different stress levels in each dataset. For GSM8K: $\{1, 3, 6, 9, 12\}$; for MATH500 and AMC23: $\{1, 3, 5, 7, 9\}$; for AIME24, AIME25, GPQA and LiveCodeBench: $\{1, 2, 3, 4, 5\}$. 

\section{Prompt}

We present the prompt for mathematical tasks, \textit{GPQA Diamond} and \textit{LiveCodeBench} in Figure~\ref{fig:math_prompt}, Figure~\ref{fig:gpqa_prompt} and Figure~\ref{fig:code_prompt}, respectively. 

\begin{figure}[h]
\begin{prompt}{\textit{Prompt for Math Tasks}}{math_prompt}
\small

$\{\text{Questions}\}$. Answer the above questions one by one. 
Remember to put your final answer within \textbackslash boxed\{\}.

\end{prompt}
\caption{Evaluation prompt for math tasks.}
\label{fig:math_prompt}
\end{figure}
\begin{figure}[h]
\begin{prompt}{\textit{Prompt for GQPA Diamond}}{gpqa_prompt}
\small

$\{\text{Questions}\}$. Answer the above multiple-choice question one by one. Remember to give each answer in the following format: `ANSWER: \textbackslash boxed\{LETTER\}' (without quotes) where LETTER is one of ABCD.

\end{prompt}
\caption{Evaluation prompt for GPQA Diamond.}
\label{fig:gpqa_prompt}
\end{figure}
\begin{figure}[h]
\begin{prompt}{\textit{Prompt for Code Generation}}{code_prompt}
\small

$\{\text{Questions}\}$. 
Answer the above questions one by one. Enclose the code for each question within delimiters as follows.

\textasciigrave\textasciigrave\textasciigrave python \#YOUR CODE HERE
\textasciigrave\textasciigrave\textasciigrave.

\#\#\# Answer: (use the provided format with backticks)

\end{prompt}
\caption{Evaluation prompt for code generation tasks.}
\label{fig:code_prompt}
\end{figure}

\begin{figure}[h]
\begin{prompt}{\textit{Answer Extraction Promopt}}{answer_extraction_prompt}
\small

Extract the final answers from the given predictions. Here are some extraction criteria:

1. Don't try to answer the original question. Your task is to extract the final answer from the prediction as it is, even if it is incorrect.\\
2. Prediction sometimes involves lengthy thinking processes, you don't need to consider these, just extract the final answer. \\
3. If there is question that has not been answered, don't answer it yourself. You should set the final answer to None (e.g., \textbackslash boxed\{None\}).\\
4. Ensure the number of final answers you extract is exactly the same as the number of the given questions. \\
5. Extract the final answer for each question one by one and enclose each final answer within an \textbackslash boxed{}. For example, if there are three questions, the output should be \\
Answer to Q1: \textbackslash boxed\{answer 1\}\\
Answer to Q2: \textbackslash boxed\{answer 2\}\\
Answer to Q3: \textbackslash boxed\{answer 3\}.

Here is your task. Simply extract the final answers from the given predictions. Don't apologize or correct yourself if there was a mistake in the predictions; we are just trying to extract the final answer.

<Questions Begin>: 

\{question\}

<Questions End>

<Predictions Begin>: 

\{prediction\}

<Predictions End>
    
Extract the final answers from the given predictions.
    
\end{prompt}
\caption{Prompt for extracting answers from the response.}
\label{fig:answer_extracgtion_prompt}
\end{figure}

\section{Comparison between Rule-based and LLM-based Answer Extraction}
\label{sec:answer_extraction}

\begin{table}[!h]
    \centering
    \def\arraystretch{1,2}
    \resizebox{0.7\linewidth}{!}{%
    \begin{tabular}{c|ccccc}
    \toprule[0.8pt]
    Extraction Method & GSM8K & MATH500 & AMC23
    & AIME24 & AIME25 \\
    \hline
    \rowcolor{lightgray}
    \multicolumn{6}{c}{\textit{DS-R1-Distill-Qwen-7B}} \\
    Rule-Based & 89.06 & 66.75 & 36.06 & 16.53 & 11.37 \\
    LLM-Based & 90.00 & 70.59 & 41.05 & 18.32 & 14.16 \\
    \rowcolor{lightgray}
    \multicolumn{6}{c}{\textit{DS-R1-Distill-Qwen-32B}} \\
    Rule-Based & 95.50 & 88.97 & 86.24 & 52.51 & 33.83 \\
    LLM-Based & 95.83 & 92.18 & 87.48 & 53.63 & 34.39 \\
    
    \bottomrule[0.8pt]
    \end{tabular}
    }
    \vspace{2mm}
    \caption{
Comparison of rule-based answer extraction and LLM-based answer extraction. 
    }
    \label{tab:answer_extraction}
\end{table}

In this section, we compare the performance of several representative LRMs under \sysname\ when using rule-based versus LLM-based answer extraction. For rule-based extraction, we search the specific pattern sequentially using code modified from Qwen-Math evaluation tool~\footnote{\url{https://github.com/QwenLM/Qwen2.5-Math}}. For the LLM-based approach, we employ \textit{gemma-3-27b-it}~\citep{team2025gemma} with prompt shown in Figure~\ref{fig:answer_extracgtion_prompt}. The results in Table~\ref{tab:answer_extraction} indicate that LLM-based answer extraction can lead to a modest performance improvement. However, this approach significantly increases evaluation costs and may introduce additional instability, making it less practical for large-scale assessments. Therefore, we adopt rule-based answer extraction as the default approach in our experiments.

\section{The Impact of the maximum output length}
\label{sec:impact_of_max-out-len}
\paragraph{Extending the maximum output length can not provide substantial improvement.}
As our error analysis identified output truncation as a significant error type, it is questionable whether extending the maximum output length could improve performance under stress tests. To evaluate this, we extended the maximum output length from 32K to 128K tokens for top-tier 7B$\sim$32B reasoning models. 
As shown in Tab.~\ref{tab:output_128k}, this extension has a negligible impact on stress test accuracy. 

\begin{table}[!h]
    \centering
    \def\arraystretch{1,2}
    \resizebox{0.7\linewidth}{!}{%
    \begin{tabular}{c|ccccc}
    \toprule[0.8pt]
    Max Output Token & GSM8K & MATH500 & AMC23
    & AIME24 & AIME25 \\
    \hline
    \rowcolor{lightgray}
    \multicolumn{6}{c}{\textit{DS-R1-Distill-Qwen-7B}} \\
    32K & 89.06 & 66.75 & 36.06 & 16.53 & 11.37 \\
    128K & 89.37 & 65.83 & 36.66 & 16.17 & 11.73 \\
    \rowcolor{lightgray}
    \multicolumn{6}{c}{\textit{DS-R1-Distill-Qwen-32B}} \\
    32K & 95.50 & 88.97 & 86.24 & 52.51 & 33.83 \\
    128K & 95.62 & 89.71 & 86.77 & 52.75 & 33.87 \\
    
    \bottomrule[0.8pt]
    \end{tabular}
    }
    \vspace{2mm}
    \caption{
The effect of extending the maximum output length. Extending the maximum output length has almost negligible impact on both 7B and 32B models.
    }
    \label{tab:output_128k}
\end{table}

\FloatBarrier

\section{Limitation}
\label{sec:limitation}

While \sysname\ provides a powerful and discriminative evaluation framework, several limitations remain. First, it increases evaluation costs, as multi-problem prompts generally require longer reasoning sequences, leading to increased computational overhead. Additionally, given the rapid pace of advancement in reasoning models, there is always a risk that our experiments may not capture all the latest state-of-the-art models released after our initial evaluations. To address this, we plan to continuously monitor newly released models and maintain an up-to-date leaderboard to reflect the latest progress in the field.

\begin{table}[t!]
\fontsize{9}{10} \selectfont
    \centering
    \def\arraystretch{1,2}
    \resizebox{\textwidth}{!}{%
    \begin{tabular}{ccccccccc}
    \toprule[0.8pt]
       \diagbox{Model}{Bench} & \makecell[c]{Stress\\Degree} & GSM8K & MATH500 & AMC23 & AIME24 & AIME25 & \makecell[c]{GPQA\\Diamond} & \makecell[c]{LiveCode\\Bench} \\
       \hline
       \rowcolor{brown!10}
       \multicolumn{9}{c}{\textit{1.5B LRMs}} \\
        \multirow{5}{*}{DS-R1-Distill-Qwen-1.5B} & \uppercase\expandafter{\romannumeral1} & 84.62 & 83.40 & 62.50 & 29.17 & 25.00 & 37.37 & 15.05 \\
       & \cellcolor{gray!10}\uppercase\expandafter{\romannumeral2} & \cellcolor{gray!10}82.21 & \cellcolor{gray!10}65.53 & \cellcolor{gray!10}28.75 & \cellcolor{gray!10}9.17 & \cellcolor{gray!10}13.54 & \cellcolor{gray!10}23.74 & \cellcolor{gray!10}1.08 \\
       & \cellcolor{gray!15}\uppercase\expandafter{\romannumeral3} & \cellcolor{gray!15}76.21 & \cellcolor{gray!15}47.52 & \cellcolor{gray!15}13.75 & \cellcolor{gray!15}6.11 & \cellcolor{gray!15}5.83 & \cellcolor{gray!15}23.06 & \cellcolor{gray!15}0.60 \\
       & \cellcolor{gray!20}\uppercase\expandafter{\romannumeral4} & \cellcolor{gray!20}66.30 & \cellcolor{gray!20}34.43 & \cellcolor{gray!20}6.61 & \cellcolor{gray!20}2.08 & \cellcolor{gray!20}2.60 & \cellcolor{gray!20}21.34 & \cellcolor{gray!20}0.18 \\
       & \cellcolor{gray!25}\uppercase\expandafter{\romannumeral5} & \cellcolor{gray!25}56.13 & \cellcolor{gray!25}22.40 & \cellcolor{gray!25}6.81 & \cellcolor{gray!25}2.50 & \cellcolor{gray!25}1.67 & \cellcolor{gray!25}20.30 & \cellcolor{gray!25}0.07 \\

       \multirow{5}{*}{DeepScaleR-1.5B} & \uppercase\expandafter{\romannumeral1} & 84.84 & 87.60 & 76.25 & 38.75 & 31.25 & 31.82 & 21.15 \\
       & \cellcolor{gray!10}\uppercase\expandafter{\romannumeral2} & \cellcolor{gray!10} 83.37 & \cellcolor{gray!10} 79.67 & \cellcolor{gray!10} 54.58 & \cellcolor{gray!10} 21.88 & \cellcolor{gray!10} 21.67 & \cellcolor{gray!10} 31.31 & \cellcolor{gray!10} 3.76 \\
       & \cellcolor{gray!15}\uppercase\expandafter{\romannumeral3} & \cellcolor{gray!15} 73.81 & \cellcolor{gray!15} 66.88 & \cellcolor{gray!15} 37.25 & \cellcolor{gray!15} 13.19 & \cellcolor{gray!15} 15.97 & \cellcolor{gray!15} 27.27 & \cellcolor{gray!15} 1.67 \\
       & \cellcolor{gray!20}\uppercase\expandafter{\romannumeral4} & \cellcolor{gray!20} 61.10 & \cellcolor{gray!20} 50.31 & \cellcolor{gray!20} 21.79 & \cellcolor{gray!20} 9.79 & \cellcolor{gray!20} 11.77 & \cellcolor{gray!20} 26.77 & \cellcolor{gray!20} 1.61 \\
       & \cellcolor{gray!25}\uppercase\expandafter{\romannumeral5} & \cellcolor{gray!25} 48.04 & \cellcolor{gray!25} 42.20 & \cellcolor{gray!25} 14.58 & \cellcolor{gray!25} 6.42 & \cellcolor{gray!25} 7.50 & \cellcolor{gray!25} 26.26 & \cellcolor{gray!25} 0.29 \\
       \multirow{5}{*}{L1-Qwen-1.5B-Exact} & \uppercase\expandafter{\romannumeral1} & 84.87 & 84.00 & 71.25 & 21.25 & 18.33 & 33.84 & 18.28 \\
        & \cellcolor{gray!10}\uppercase\expandafter{\romannumeral2} & \cellcolor{gray!10} 84.25 & \cellcolor{gray!10} 78.07 & \cellcolor{gray!10} 57.92 & \cellcolor{gray!10} 13.33 & \cellcolor{gray!10} 18.33 & \cellcolor{gray!10} 31.57 & \cellcolor{gray!10} 5.73 \\
       & \cellcolor{gray!15}\uppercase\expandafter{\romannumeral3} & \cellcolor{gray!15} 83.19 & \cellcolor{gray!15} 72.96 & \cellcolor{gray!15} 50.00 & \cellcolor{gray!15} 14.72 & \cellcolor{gray!15} 12.08 & \cellcolor{gray!15} 32.83 & \cellcolor{gray!15} 1.79 \\
       & \cellcolor{gray!20}\uppercase\expandafter{\romannumeral4} &\cellcolor{gray!20} 76.62 & \cellcolor{gray!20} 70.46 & \cellcolor{gray!20} 41.96 & \cellcolor{gray!20} 12.19 & \cellcolor{gray!20} 11.25 & \cellcolor{gray!20} 30.43 & \cellcolor{gray!20} 2.33 \\
       & \cellcolor{gray!25}\uppercase\expandafter{\romannumeral5} & \cellcolor{gray!25} 69.99 & \cellcolor{gray!25} 66.78 & \cellcolor{gray!25} 39.58 & \cellcolor{gray!25} 10.25 & \cellcolor{gray!25} 10.17 & \cellcolor{gray!25} 29.19 & \cellcolor{gray!25} 0.93 \\
       
       \multirow{5}{*}{L1-Qwen-1.5B-Max} & \uppercase\expandafter{\romannumeral1} & 84.17 & 83.40 & 77.50 & 20.00 & 22.92 & 36.87 & 19.35 \\
        & \cellcolor{gray!10}\uppercase\expandafter{\romannumeral2} & \cellcolor{gray!10} 83.92 & \cellcolor{gray!10} 78.93 & \cellcolor{gray!10} 61.67 & \cellcolor{gray!10} 16.25 & \cellcolor{gray!10} 18.33 & \cellcolor{gray!10} 32.32 & \cellcolor{gray!10} 4.48 \\
       & \cellcolor{gray!15}\uppercase\expandafter{\romannumeral3} & \cellcolor{gray!15} 81.98 & \cellcolor{gray!15} 75.28 & \cellcolor{gray!15} 52.00 & \cellcolor{gray!15} 16.11 & \cellcolor{gray!15} 16.53 & \cellcolor{gray!15} 32.49 & \cellcolor{gray!15} 2.27 \\
       & \cellcolor{gray!20}\uppercase\expandafter{\romannumeral4} & \cellcolor{gray!20} 77.37 & \cellcolor{gray!20} 71.60 & \cellcolor{gray!20} 41.61 & \cellcolor{gray!20} 15.00 & \cellcolor{gray!20} 13.44 & \cellcolor{gray!20} 30.68 & \cellcolor{gray!20} 1.61 \\
       & \cellcolor{gray!25}\uppercase\expandafter{\romannumeral5} & \cellcolor{gray!25} 67.89 & \cellcolor{gray!25} 67.11 & \cellcolor{gray!25} 38.19 & \cellcolor{gray!25} 13.17 & \cellcolor{gray!25} 11.50 & \cellcolor{gray!25} 32.63 & \cellcolor{gray!25} 1.43 \\
       
       \multirow{5}{*}{Qwen2.5-Math-1.5B-Inst} & \uppercase\expandafter{\romannumeral1} & 85.37 & 73.00 & 57.50 & 10.83 & 10.83 & 26.77 & 1.79 \\
        & \cellcolor{gray!10}\uppercase\expandafter{\romannumeral2} & \cellcolor{gray!10} 81.32 & \cellcolor{gray!10} 65.00 & \cellcolor{gray!10} 37.50 & \cellcolor{gray!10} 9.38 & \cellcolor{gray!10} 8.75 & \cellcolor{gray!10} - & \cellcolor{gray!10} -\\
       & \cellcolor{gray!15}\uppercase\expandafter{\romannumeral3} & \cellcolor{gray!15} 75.74 & \cellcolor{gray!15} 58.36 & \cellcolor{gray!15} 24.50 & \cellcolor{gray!15} 5.97 & \cellcolor{gray!15} 0.14 & \cellcolor{gray!15} - & \cellcolor{gray!15} - \\
       & \cellcolor{gray!20}\uppercase\expandafter{\romannumeral4} & \cellcolor{gray!20} 64.67 & \cellcolor{gray!20} 52.09 & \cellcolor{gray!20} 15.36 & \cellcolor{gray!20} 6.15 & \cellcolor{gray!20} 1.77 & \cellcolor{gray!20} - & \cellcolor{gray!20} -\\
       & \cellcolor{gray!25}\uppercase\expandafter{\romannumeral5} & \cellcolor{gray!25} 48.21 & \cellcolor{gray!25} 40.29 & \cellcolor{gray!25} 11.53 & \cellcolor{gray!25} 3.17 & \cellcolor{gray!25} 0.67 & \cellcolor{gray!25} - & \cellcolor{gray!25} -\\

       \multirow{5}{*}{Qwen2.5-1.5B-Inst} & \uppercase\expandafter{\romannumeral1} & 65.13 & 53.40 & 30.00 & 2.50 & 0.00 & 26.26 & 1.79 \\
        & \cellcolor{gray!10}\uppercase\expandafter{\romannumeral2} & \cellcolor{gray!10} 25.52 & \cellcolor{gray!10} 37.00 & \cellcolor{gray!10} 14.17 & \cellcolor{gray!10} 0 & \cellcolor{gray!10} 1.67 & \cellcolor{gray!10} 22.80 & \cellcolor{gray!10} 0.90\\
       & \cellcolor{gray!15}\uppercase\expandafter{\romannumeral3} & \cellcolor{gray!15} 12.94 & \cellcolor{gray!15} 28.60 & \cellcolor{gray!15} 11.00 & \cellcolor{gray!15} 4.58 & \cellcolor{gray!15} 0 & \cellcolor{gray!15} 21.96 & \cellcolor{gray!15} 0.36\\
       & \cellcolor{gray!20}\uppercase\expandafter{\romannumeral4} & \cellcolor{gray!20} 7.65 & \cellcolor{gray!20} 23.29 & \cellcolor{gray!20} 10.89 & \cellcolor{gray!20} 2.92 & \cellcolor{gray!20} 0 & \cellcolor{gray!20} 21.03 & \cellcolor{gray!20} 0.09 \\
       & \cellcolor{gray!25}\uppercase\expandafter{\romannumeral5} & \cellcolor{gray!25} 6.19 & \cellcolor{gray!25} 19.31 & \cellcolor{gray!25} 6.53 & \cellcolor{gray!25} 0.67 & \cellcolor{gray!25} 0 & \cellcolor{gray!25} 20.27 & \cellcolor{gray!25} 0.14 \\  
    
    \bottomrule[0.8pt]
    \end{tabular}
    }
    \vspace{1mm}
    \caption{Detailed accuracy statistics for each stress level of various LRMs. Some models are only evaluated on math-related benchmarks, as their official guidelines do not recommend their use for other task domains.
}
    \label{tab:main_detail1}
\end{table}

\begin{table}[t!]
\fontsize{9}{10} \selectfont
    \centering
    \def\arraystretch{1,2}
    \resizebox{\textwidth}{!}{%
    \begin{tabular}{ccccccccc}
    \toprule[0.8pt]
       \diagbox{Model}{Bench} & \makecell[c]{Stress\\Degree} & GSM8K & MATH500 & AMC23 & AIME24 & AIME25 & \makecell[c]{GPQA\\Diamond} & \makecell[c]{LiveCode\\Bench} \\
       \hline

       \rowcolor{brown!10}
       \multicolumn{9}{c}{\textit{$7\sim8$B LRMs}} \\       
       \multirow{5}{*}{DS-R1-Distill-Qwen-7B} & \uppercase\expandafter{\romannumeral1} & 89.49 & 93.00 & 87.50 & 54.17 & 35.42 & 51.01 & 37.63 \\
       & \cellcolor{gray!10}\uppercase\expandafter{\romannumeral2} & \cellcolor{gray!10} 89.01 & \cellcolor{gray!10} 83.67 & \cellcolor{gray!10} 68.33 & \cellcolor{gray!10} 35.00 & \cellcolor{gray!10} 21.67 & \cellcolor{gray!10} 38.38 & \cellcolor{gray!10} 7.35 \\
       & \cellcolor{gray!15}\uppercase\expandafter{\romannumeral3} & \cellcolor{gray!15} 90.03 & \cellcolor{gray!15} 72.16 & \cellcolor{gray!15} 41.00 & \cellcolor{gray!15} 16.94 & \cellcolor{gray!15} 13.06 & \cellcolor{gray!15} 31.31 & \cellcolor{gray!15} 2.15 \\
       & \cellcolor{gray!20}\uppercase\expandafter{\romannumeral4} & \cellcolor{gray!20} 89.37 & \cellcolor{gray!20} 61.74 & \cellcolor{gray!20} 22.14 & \cellcolor{gray!20} 8.44 & \cellcolor{gray!20} 7.92 & \cellcolor{gray!20} 29.80 & \cellcolor{gray!20} 0.90\\
       & \cellcolor{gray!25}\uppercase\expandafter{\romannumeral5} & \cellcolor{gray!25} 87.83 & \cellcolor{gray!25} 49.42 & \cellcolor{gray!25} 12.78 & \cellcolor{gray!25} 5.75 & \cellcolor{gray!25} 2.83 & \cellcolor{gray!25} 27.17 & \cellcolor{gray!25} 1.15 \\  

        \multirow{5}{*}{Efficient-R1-7B ($\alpha=0.1$)} & \uppercase\expandafter{\romannumeral1} &  88.63 &  90.00 &  87.50 &  54.58 &  35.42 &  48.98 &  39.07 \\
       & \cellcolor{gray!10}\uppercase\expandafter{\romannumeral2} &\cellcolor{gray!10} 85.49 & \cellcolor{gray!10} 86.80 & \cellcolor{gray!10} 70.83 & \cellcolor{gray!10} 40.62 & \cellcolor{gray!10} 26.46 & \cellcolor{gray!10} 39.65 & \cellcolor{gray!10} 7.35\\
       & \cellcolor{gray!15}\uppercase\expandafter{\romannumeral3} &\cellcolor{gray!15} 84.52 & \cellcolor{gray!15} 79.12 & \cellcolor{gray!15} 53.00 & \cellcolor{gray!15} 21.11 & \cellcolor{gray!15} 16.94 & \cellcolor{gray!15} 38.38 & \cellcolor{gray!15} 2.75 \\
       & \cellcolor{gray!20}\uppercase\expandafter{\romannumeral4} & \cellcolor{gray!20} 84.85 & \cellcolor{gray!20} 72.86 & \cellcolor{gray!20} 29.29 & \cellcolor{gray!20} 14.58 & \cellcolor{gray!20} 12.19 & \cellcolor{gray!20} 33.84 & \cellcolor{gray!20} 0.72\\
       & \cellcolor{gray!25}\uppercase\expandafter{\romannumeral5} & \cellcolor{gray!25} 84.16 & \cellcolor{gray!25} 61.18 & \cellcolor{gray!25} 23.89 & \cellcolor{gray!25} 9.50 & \cellcolor{gray!25} 7.42 & \cellcolor{gray!25} 31.72 & \cellcolor{gray!25} 0.65 \\  
       
       \multirow{5}{*}{Efficient-R1-7B ($\alpha=0.2$)} & \uppercase\expandafter{\romannumeral1} & 87.95 & 88.20 & 85.00 & 50.42 & 33.75 & 47.97 & 38.71 \\
       & \cellcolor{gray!10}\uppercase\expandafter{\romannumeral2} & \cellcolor{gray!10} 84.10 & \cellcolor{gray!10} 85.20 & \cellcolor{gray!10} 76.25 & \cellcolor{gray!10} 38.54 & \cellcolor{gray!10} 27.29 & \cellcolor{gray!10} 38.13 & \cellcolor{gray!10} 6.63 \\
       & \cellcolor{gray!15}\uppercase\expandafter{\romannumeral3} & \cellcolor{gray!15} 81.49 & \cellcolor{gray!15} 81.28 & \cellcolor{gray!15} 52.75 & \cellcolor{gray!15} 24.86 & \cellcolor{gray!15} 19.31 & \cellcolor{gray!15} 33.67 & \cellcolor{gray!15} 2.15 \\
       & \cellcolor{gray!20}\uppercase\expandafter{\romannumeral4} & \cellcolor{gray!20} 78.92 & \cellcolor{gray!20} 73.14 & \cellcolor{gray!20} 35.00 & \cellcolor{gray!20} 16.67 & \cellcolor{gray!20} 14.06 & \cellcolor{gray!20} 33.46 & \cellcolor{gray!20} 0.63 \\
       & \cellcolor{gray!25}\uppercase\expandafter{\romannumeral5} & \cellcolor{gray!25} 77.02 & \cellcolor{gray!25} 66.02 & \cellcolor{gray!25} 28.19 & \cellcolor{gray!25} 8.42 & \cellcolor{gray!25} 8.33 & \cellcolor{gray!25} 32.22 & \cellcolor{gray!25} 0.57 \\  
       
       \multirow{5}{*}{DS-R1-Distill-LLaMA-8B} & \uppercase\expandafter{\romannumeral1} & 90.45 & 89.80 & 87.50 & 55.00 & 28.33 & 50.00 & 39.43 \\
       & \cellcolor{gray!10}\uppercase\expandafter{\romannumeral2} & \cellcolor{gray!10} 89.13 & \cellcolor{gray!10} 85.20 & \cellcolor{gray!10} 81.25 & \cellcolor{gray!10} 38.33 & \cellcolor{gray!10} 25.00 & \cellcolor{gray!10} 39.14 & \cellcolor{gray!10} 12.54 \\
       & \cellcolor{gray!15}\uppercase\expandafter{\romannumeral3} & \cellcolor{gray!15} 86.59 & \cellcolor{gray!15} 82.92 & \cellcolor{gray!15} 75.25 & \cellcolor{gray!15} 36.67 & \cellcolor{gray!15} 22.78 & \cellcolor{gray!15} 34.85 & \cellcolor{gray!15} 14.34 \\
       & \cellcolor{gray!20}\uppercase\expandafter{\romannumeral4} & \cellcolor{gray!20} 84.33 & \cellcolor{gray!20} 80.69 & \cellcolor{gray!20} 64.29 & \cellcolor{gray!20} 27.92 & \cellcolor{gray!20} 24.17 & \cellcolor{gray!20} 32.07 & \cellcolor{gray!20} 10.75 \\
       & \cellcolor{gray!25}\uppercase\expandafter{\romannumeral5} & \cellcolor{gray!25} 80.67 & \cellcolor{gray!25} 76.53 & \cellcolor{gray!25} 62.22 & \cellcolor{gray!25} 22.00 & \cellcolor{gray!25} 18.67 & \cellcolor{gray!25} 29.90 & \cellcolor{gray!25} 7.38 \\
       
       \multirow{5}{*}{Nemotron-Nano-8B} & \uppercase\expandafter{\romannumeral1} & 91.36 & 94.40 & 90.00 & 63.33 & 50.00 & 51.01 & 50.90 \\
       & \cellcolor{gray!10}\uppercase\expandafter{\romannumeral2} & \cellcolor{gray!10} 90.37 & \cellcolor{gray!10} 89.67 & \cellcolor{gray!10} 88.75 & \cellcolor{gray!10} 52.50 & \cellcolor{gray!10} 40.83 & \cellcolor{gray!10} 31.57 & \cellcolor{gray!10} 15.41 \\
       & \cellcolor{gray!15}\uppercase\expandafter{\romannumeral3} & \cellcolor{gray!15} 69.27 & \cellcolor{gray!15} 87.48 & \cellcolor{gray!15} 80.00 & \cellcolor{gray!15} 47.78 & \cellcolor{gray!15} 36.11 & \cellcolor{gray!15} 34.18 & \cellcolor{gray!15} 9.08 \\
       & \cellcolor{gray!20}\uppercase\expandafter{\romannumeral4} & \cellcolor{gray!20} 34.93 & \cellcolor{gray!20} 84.83 & \cellcolor{gray!20} 70.36 & \cellcolor{gray!20} 41.25 & \cellcolor{gray!20} 27.50 & \cellcolor{gray!20} 38.76 & \cellcolor{gray!20} 6.09 \\
       & \cellcolor{gray!25}\uppercase\expandafter{\romannumeral5} & \cellcolor{gray!25} 18.71 & \cellcolor{gray!25} 82.18 & \cellcolor{gray!25} 65.83 & \cellcolor{gray!25} 32.67 & \cellcolor{gray!25} 24.67 & \cellcolor{gray!25} 31.72 & \cellcolor{gray!25} 3.66 \\     
       
       \multirow{5}{*}{AReaL-boba-RL-7B} & \uppercase\expandafter{\romannumeral1} & 91.66 & 95.00 & 91.25 & 61.25 & 45.83 & 48.99 & 37.99 \\
       & \cellcolor{gray!10}\uppercase\expandafter{\romannumeral2} & \cellcolor{gray!10} 91.08 & \cellcolor{gray!10} 83.00 & \cellcolor{gray!10} 60.42 & \cellcolor{gray!10} 37.08 & \cellcolor{gray!10} 22.08 & \cellcolor{gray!10} 38.64 & \cellcolor{gray!10} 8.60\\
       & \cellcolor{gray!15}\uppercase\expandafter{\romannumeral3} & \cellcolor{gray!15} 85.72 & \cellcolor{gray!15} 68.56 & \cellcolor{gray!15} 37.00 & \cellcolor{gray!15} 22.22 & \cellcolor{gray!15} 12.64 & \cellcolor{gray!15} 27.61 & \cellcolor{gray!15} 3.46 \\
       & \cellcolor{gray!20}\uppercase\expandafter{\romannumeral4} & \cellcolor{gray!20} 72.48 & \cellcolor{gray!20} 53.46 & \cellcolor{gray!20} 20.71 & \cellcolor{gray!20} 15.00 & \cellcolor{gray!20} 8.33 & \cellcolor{gray!20} 25.00 & \cellcolor{gray!20} 3.41\\
       & \cellcolor{gray!25}\uppercase\expandafter{\romannumeral5} & \cellcolor{gray!25} 61.91 & \cellcolor{gray!25} 38.07 & \cellcolor{gray!25} 13.61 & \cellcolor{gray!25} 11.42 & \cellcolor{gray!25} 6.25 & \cellcolor{gray!25} 25.25 & \cellcolor{gray!25} 3.44 \\
       
        \multirow{5}{*}{Light-R1-7B-DS} & \uppercase\expandafter{\romannumeral1} & 88.05 & 93.20 & 90.00 & 55.83 & 45.83 & 51.01 & 39.07 \\
       & \cellcolor{gray!10}\uppercase\expandafter{\romannumeral2} & \cellcolor{gray!10} 83.90 & \cellcolor{gray!10} 83.47 & \cellcolor{gray!10} 67.50 & \cellcolor{gray!10} 31.87 & \cellcolor{gray!10} 24.17 & \cellcolor{gray!10} 36.87 & \cellcolor{gray!10} 8.06 \\
       & \cellcolor{gray!15}\uppercase\expandafter{\romannumeral3} & \cellcolor{gray!15} 84.96 & \cellcolor{gray!15} 69.16 & \cellcolor{gray!15} 38.00 & \cellcolor{gray!15} 18.19 & \cellcolor{gray!15} 14.17 & \cellcolor{gray!15} 32.15 & \cellcolor{gray!15} 1.43 \\
       & \cellcolor{gray!20}\uppercase\expandafter{\romannumeral4} & \cellcolor{gray!20} 83.56 & \cellcolor{gray!20} 55.17 & \cellcolor{gray!20} 23.57 & \cellcolor{gray!20} 10.21 & \cellcolor{gray!20} 7.81 & \cellcolor{gray!20} 28.41 & \cellcolor{gray!20} 0.81 \\
       & \cellcolor{gray!25}\uppercase\expandafter{\romannumeral5} & \cellcolor{gray!25} 78.34 & \cellcolor{gray!25} 39.11 & \cellcolor{gray!25} 10.56 & \cellcolor{gray!25} 6.25 & \cellcolor{gray!25} 5.67 & \cellcolor{gray!25} 23.84 & \cellcolor{gray!25} 1.08 \\
        \bottomrule[0.8pt]
    \end{tabular}
    }
    \vspace{2mm}
    \caption[]{Detailed accuracy statistics for each stress level of various LRMs. Some models are only evaluated on math-related benchmarks, as their official guidelines do not recommend their use for other task domains.
}
    \label{tab:main_detail2}
\end{table}

\begin{table}[t!]
\fontsize{9}{10} \selectfont
    \centering
    \def\arraystretch{1,2}
    \resizebox{\textwidth}{!}{%
    \begin{tabular}{ccccccccc}
    \toprule[0.8pt]
       \diagbox{Model}{Bench} & \makecell[c]{Stress\\Degree} & GSM8K & MATH500 & AMC23 & AIME24 & AIME25 & \makecell[c]{GPQA\\Diamond} & \makecell[c]{LiveCode\\Bench} \\
       \hline
       
       \rowcolor{brown!10}
       \multicolumn{9}{c}{\textit{$7\sim8$B LRMs}} \\
       \multirow{5}{*}{OpenR1-Qwen-7B} & \uppercase\expandafter{\romannumeral1} & 95.60 & 92.20 & 83.75 & 47.50 & 32.92 & 38.38 & 4.66 \\
       & \cellcolor{gray!10}\uppercase\expandafter{\romannumeral2} & \cellcolor{gray!10} 91.43 & \cellcolor{gray!10} 86.40 & \cellcolor{gray!10} 73.75 & \cellcolor{gray!10} 37.08 & \cellcolor{gray!10} 27.50 & \cellcolor{gray!10} 35.86 & \cellcolor{gray!10} 2.33 \\
       & \cellcolor{gray!15}\uppercase\expandafter{\romannumeral3} & \cellcolor{gray!15} 92.67 & \cellcolor{gray!15} 83.64 & \cellcolor{gray!15} 61.00 & \cellcolor{gray!15} 26.11 & \cellcolor{gray!15} 25.56 & \cellcolor{gray!15} 36.70 & \cellcolor{gray!15} 0.72 \\
       & \cellcolor{gray!20}\uppercase\expandafter{\romannumeral4} & \cellcolor{gray!20} 90.38 & \cellcolor{gray!20} 80.11 & \cellcolor{gray!20} 44.46 & \cellcolor{gray!20} 26.15 & \cellcolor{gray!20} 18.12 & \cellcolor{gray!20} 38.76 & \cellcolor{gray!20} 0.72 \\
       & \cellcolor{gray!25}\uppercase\expandafter{\romannumeral5} & \cellcolor{gray!25} 86.38 & \cellcolor{gray!25} 76.42 & \cellcolor{gray!25} 37.22 & \cellcolor{gray!25} 17.75 & \cellcolor{gray!25} 13.58 & \cellcolor{gray!25} 32.83 & \cellcolor{gray!25} 1.29 \\
       
       \multirow{5}{*}{OpenThinker2-7B} & \uppercase\expandafter{\romannumeral1} & 94.39 & 93.80 & 85.00 & 54.58 & 41.67 & 44.44 & 39.43 \\
       & \cellcolor{gray!10}\uppercase\expandafter{\romannumeral2} & \cellcolor{gray!10} 93.78 & \cellcolor{gray!10} 89.20 & \cellcolor{gray!10} 78.75 & \cellcolor{gray!10} 45.83 & \cellcolor{gray!10} 30.21 & \cellcolor{gray!10} 41.67 & \cellcolor{gray!10} 17.38 \\
       & \cellcolor{gray!15}\uppercase\expandafter{\romannumeral3} & \cellcolor{gray!15} 92.85 & \cellcolor{gray!15} 83.48 & \cellcolor{gray!15} 63.00 & \cellcolor{gray!15} 36.67 & \cellcolor{gray!15} 27.36 & \cellcolor{gray!15} 40.91 & \cellcolor{gray!15} 16.37 \\
       & \cellcolor{gray!20}\uppercase\expandafter{\romannumeral4} & \cellcolor{gray!20} 91.31 & \cellcolor{gray!20} 82.23 & \cellcolor{gray!20} 58.39 & \cellcolor{gray!20} 31.15 & \cellcolor{gray!20} 20.31 & \cellcolor{gray!20} 42.05 & \cellcolor{gray!20} 12.81 \\
       & \cellcolor{gray!25}\uppercase\expandafter{\romannumeral5} & \cellcolor{gray!25} 90.01 & \cellcolor{gray!25} 78.29 & \cellcolor{gray!25} 52.78 & \cellcolor{gray!25} 24.33 & \cellcolor{gray!25} 16.75 & \cellcolor{gray!25} 37.78 & \cellcolor{gray!25} 11.47 \\

       \multirow{5}{*}{SimpleRL-Zoo-Qwen-7B} & \uppercase\expandafter{\romannumeral1} & 90.52 & 77.80 & 62.50 & 26.67 & 10.00 & 33.84 & 5.73 \\
       & \cellcolor{gray!10}\uppercase\expandafter{\romannumeral2} & \cellcolor{gray!10} 81.20 & \cellcolor{gray!10} 70.53 & \cellcolor{gray!10} 41.67 & \cellcolor{gray!10} 6.67 & \cellcolor{gray!10} 10.00 & \cellcolor{gray!10} 36.62 & \cellcolor{gray!10} 0.18 \\
       & \cellcolor{gray!15}\uppercase\expandafter{\romannumeral3} & \cellcolor{gray!15} 86.57 & \cellcolor{gray!15} 66.76 & \cellcolor{gray!15} 29.50 & \cellcolor{gray!15} 12.08 & \cellcolor{gray!15} 5.42 & \cellcolor{gray!15} 33.16 & \cellcolor{gray!15} 0.36 \\
       & \cellcolor{gray!20}\uppercase\expandafter{\romannumeral4} & \cellcolor{gray!20} 86.31 & \cellcolor{gray!20} 59.54 & \cellcolor{gray!20} 31.79 & \cellcolor{gray!20} 7.29 & \cellcolor{gray!20} 5.21 & \cellcolor{gray!20} 36.11 & \cellcolor{gray!20} 0.09 \\
       & \cellcolor{gray!25}\uppercase\expandafter{\romannumeral5} & \cellcolor{gray!25} 81.97 & \cellcolor{gray!25} 52.82 & \cellcolor{gray!25} 21.94 & \cellcolor{gray!25} 4.17 & \cellcolor{gray!25} 5.25 & \cellcolor{gray!25} 36.06 & \cellcolor{gray!25} 0\\
       
       \multirow{5}{*}{Open-Reasoner-Zero-7B} & \uppercase\expandafter{\romannumeral1} & 92.87 & 83.00 & 60.00 & 17.92 & 16.25 & 37.37 & 16.13 \\
       & \cellcolor{gray!10}\uppercase\expandafter{\romannumeral2} & \cellcolor{gray!10} 87.19 & \cellcolor{gray!10} 52.93 & \cellcolor{gray!10} 30.83 & \cellcolor{gray!10} 9.79 & \cellcolor{gray!10} 7.50 & \cellcolor{gray!10} 33.33 & \cellcolor{gray!10} 0.72 \\
       & \cellcolor{gray!15}\uppercase\expandafter{\romannumeral3} & \cellcolor{gray!15} 68.46 & \cellcolor{gray!15} 34.84 & \cellcolor{gray!15} 15.75 & \cellcolor{gray!15} 7.92 & \cellcolor{gray!15} 4.31 & \cellcolor{gray!15} 35.69 & \cellcolor{gray!15} 0.48 \\
       & \cellcolor{gray!20}\uppercase\expandafter{\romannumeral4} & \cellcolor{gray!20} 55.10 & \cellcolor{gray!20} 26.46 & \cellcolor{gray!20} 10.00 & \cellcolor{gray!20} 3.54 & \cellcolor{gray!20} 1.56 & \cellcolor{gray!20} 35.86 & \cellcolor{gray!20} 0.63 \\
       & \cellcolor{gray!25}\uppercase\expandafter{\romannumeral5} & \cellcolor{gray!25} 49.80 & \cellcolor{gray!25} 15.82 & \cellcolor{gray!25} 8.06 & \cellcolor{gray!25} 3.25 & \cellcolor{gray!25} 2.17 & \cellcolor{gray!25} 32.12 & \cellcolor{gray!25} 0.72 \\

       \multirow{5}{*}{Marco-O1-7B} & \uppercase\expandafter{\romannumeral1} & 89.08 & 72.40 & 47.50 & 10.00 & 10.83 & 30.81 & 9.32 \\
       & \cellcolor{gray!10}\uppercase\expandafter{\romannumeral2} & \cellcolor{gray!10} 83.62 & \cellcolor{gray!10} 57.80 & \cellcolor{gray!10} 26.67 & \cellcolor{gray!10} 5.42 & \cellcolor{gray!10} 5.62 & \cellcolor{gray!10} 28.54 & \cellcolor{gray!10} 8.24 \\
       & \cellcolor{gray!15}\uppercase\expandafter{\romannumeral3} & \cellcolor{gray!15} 76.06 & \cellcolor{gray!15} 51.08 & \cellcolor{gray!15} 17.00 & \cellcolor{gray!15} 5.69 & \cellcolor{gray!15} 3.47 & \cellcolor{gray!15} 30.64 & \cellcolor{gray!15} 6.69\\
       & \cellcolor{gray!20}\uppercase\expandafter{\romannumeral4} & \cellcolor{gray!20} 77.95 & \cellcolor{gray!20} 46.66 & \cellcolor{gray!20} 13.57 & \cellcolor{gray!20} 4.27 & \cellcolor{gray!20} 4.79 & \cellcolor{gray!20} 29.67 & \cellcolor{gray!20} 7.08\\
       & \cellcolor{gray!25}\uppercase\expandafter{\romannumeral5} & \cellcolor{gray!25} 80.61 & \cellcolor{gray!25} 37.20 & \cellcolor{gray!25} 11.67 & \cellcolor{gray!25} 2.00 & \cellcolor{gray!25} 0.67 & \cellcolor{gray!25} 24.44 & \cellcolor{gray!25} 6.95\\

       \multirow{5}{*}{MathFusion-Qwen-7B} & \uppercase\expandafter{\romannumeral1} & 89.46 & 74.00 & 52.50 & 9.58 & 5.83 & - & - \\
       & \cellcolor{gray!10}\uppercase\expandafter{\romannumeral2} & \cellcolor{gray!10} 87.47 & \cellcolor{gray!10} 72.27 & \cellcolor{gray!10} 50.00 & \cellcolor{gray!10} 10.42 & \cellcolor{gray!10} 3.12 & \cellcolor{gray!10}- & \cellcolor{gray!10} -\\
       & \cellcolor{gray!15}\uppercase\expandafter{\romannumeral3} & \cellcolor{gray!15} 84.76 & \cellcolor{gray!15} 69.32 & \cellcolor{gray!15} 36.00 & \cellcolor{gray!15} 7.78 & \cellcolor{gray!15} 4.03 & \cellcolor{gray!15} -& \cellcolor{gray!15} -\\
       & \cellcolor{gray!20}\uppercase\expandafter{\romannumeral4} & \cellcolor{gray!20} 82.92 & \cellcolor{gray!20} 67.17 & \cellcolor{gray!20} 32.86 & \cellcolor{gray!20} 6.67 & \cellcolor{gray!20} 1.98 & \cellcolor{gray!20} -& \cellcolor{gray!20}- \\
       & \cellcolor{gray!25}\uppercase\expandafter{\romannumeral5} & \cellcolor{gray!25} 79.97 & \cellcolor{gray!25} 63.84 & \cellcolor{gray!25} 26.11 & \cellcolor{gray!25} 6.67 & \cellcolor{gray!25} 0.25 & \cellcolor{gray!25}- & \cellcolor{gray!25}- \\

       \multirow{5}{*}{Eurus-2-7B-PRIME} & \uppercase\expandafter{\romannumeral1} &  92.72 &  81.40 &  62.50 &  20.83 &  14.58 &- & - \\
       & \cellcolor{gray!10}\uppercase\expandafter{\romannumeral2} & \cellcolor{gray!10} 90.12 & \cellcolor{gray!10} 70.60 & \cellcolor{gray!10} 49.17 & \cellcolor{gray!10} 14.17 & \cellcolor{gray!10} 6.46 & \cellcolor{gray!10} - & \cellcolor{gray!10} - \\
       & \cellcolor{gray!15}\uppercase\expandafter{\romannumeral3} & \cellcolor{gray!15} 88.38 & \cellcolor{gray!15} 64.72 & \cellcolor{gray!15} 42.75 & \cellcolor{gray!15} 13.75 & \cellcolor{gray!15} 4.31& \cellcolor{gray!15} - & \cellcolor{gray!15} - \\
       & \cellcolor{gray!20}\uppercase\expandafter{\romannumeral4} & \cellcolor{gray!20} 86.83 & \cellcolor{gray!20} 63.80 & \cellcolor{gray!20} 35.18 & \cellcolor{gray!20} 7.60 & \cellcolor{gray!20} 4.17 & \cellcolor{gray!20} - & \cellcolor{gray!20} - \\
       & \cellcolor{gray!25}\uppercase\expandafter{\romannumeral5} & \cellcolor{gray!25} 86.70 & \cellcolor{gray!25} 59.64 & \cellcolor{gray!25} 27.22 & \cellcolor{gray!25} 7.83 & \cellcolor{gray!25} 3.00 & \cellcolor{gray!25} - & \cellcolor{gray!25} - \\
       
       \multirow{5}{*}{Qwen2.5-Math-7B-Inst} & \uppercase\expandafter{\romannumeral1} & 95.53 & 83.60 & 60.00 & 14.17 & 11.67 & 35.35 & 4.30 \\
       & \cellcolor{gray!10}\uppercase\expandafter{\romannumeral2} & \cellcolor{gray!10} 92.14 & \cellcolor{gray!10} 73.27 & \cellcolor{gray!10} 45.00 & \cellcolor{gray!10} 8.12 & \cellcolor{gray!10} 9.17 & \cellcolor{gray!10} 33.58 & \cellcolor{gray!10} 2.87 \\
       & \cellcolor{gray!15}\uppercase\expandafter{\romannumeral3} & \cellcolor{gray!15} 86.62 & \cellcolor{gray!15} 64.28 & \cellcolor{gray!15} 29.00 & \cellcolor{gray!15} 7.78 & \cellcolor{gray!15} 3.75 & \cellcolor{gray!15} & \cellcolor{gray!15} 1.55\\
       & \cellcolor{gray!20}\uppercase\expandafter{\romannumeral4} & \cellcolor{gray!20} 76.03 & \cellcolor{gray!20} 49.89 & \cellcolor{gray!20} 21.79 & \cellcolor{gray!20} 6.35 & \cellcolor{gray!20} 4.58 & \cellcolor{gray!20} 34.34 & \cellcolor{gray!20} 0.81 \\
       & \cellcolor{gray!25}\uppercase\expandafter{\romannumeral5} & \cellcolor{gray!25} 59.32 & \cellcolor{gray!25} 38.91 & \cellcolor{gray!25} 18.06 & \cellcolor{gray!25} 3.33 & \cellcolor{gray!25} 3.83 & \cellcolor{gray!25} & \cellcolor{gray!25}0.57 \\

       \multirow{5}{*}{Qwen2.5-7B-Instruct} & \uppercase\expandafter{\romannumeral1} & 92.27 & 77.60 &42.50 & 10.00 & 3.75 & 35.86 & 13.98 \\
       & \cellcolor{gray!10}\uppercase\expandafter{\romannumeral2} & \cellcolor{gray!10} 88.07 & \cellcolor{gray!10} 68.80 & \cellcolor{gray!10} 41.67 & \cellcolor{gray!10} 7.71 & \cellcolor{gray!10} 0 & \cellcolor{gray!10} 33.08 & \cellcolor{gray!10} 11.83 \\
       & \cellcolor{gray!15}\uppercase\expandafter{\romannumeral3} & \cellcolor{gray!15} 85.37 & \cellcolor{gray!15} 67.60 & \cellcolor{gray!15} 34.00 & \cellcolor{gray!15} 6.53 & \cellcolor{gray!15} 3.89 & \cellcolor{gray!15} 35.69 & \cellcolor{gray!15} 11.47 \\
       & \cellcolor{gray!20}\uppercase\expandafter{\romannumeral4} & \cellcolor{gray!20} 83.89 & \cellcolor{gray!20} 64.89 & \cellcolor{gray!20} 30.36 & \cellcolor{gray!20} 5.42 & \cellcolor{gray!20} 4.38 & \cellcolor{gray!20} 36.99 & \cellcolor{gray!20} 9.05 \\
       & \cellcolor{gray!25}\uppercase\expandafter{\romannumeral5} & \cellcolor{gray!25} 83.16 & \cellcolor{gray!25} 61.84 & \cellcolor{gray!25} 31.81 & \cellcolor{gray!25} 8.42 & \cellcolor{gray!25} 5.00 & \cellcolor{gray!25} 34.85 & \cellcolor{gray!25} 8.39 \\

        \bottomrule[0.8pt]
    \end{tabular}
    }
    \vspace{2mm}
    \caption[]{Detailed accuracy statistics for each stress level of various LRMs. Some models are only evaluated on math-related benchmarks, as their official guidelines do not recommend their use for other task domains.
}
    \label{tab:main_detail3}
\end{table}

\begin{table}[t!]
\fontsize{9}{10} \selectfont
    \centering
    \def\arraystretch{1,2}
    \resizebox{\textwidth}{!}{%
    \begin{tabular}{ccccccccc}
    \toprule[0.8pt]
       \diagbox{Model}{Bench} & \makecell[c]{Stress\\Degree} & GSM8K & MATH500 & AMC23 & AIME24 & AIME25 & \makecell[c]{GPQA\\Diamond} & \makecell[c]{LiveCode\\Bench} \\
       \hline
       \rowcolor{brown!10}
       \multicolumn{9}{c}{\textit{32B LRMs}} \\
       \multirow{5}{*}{DS-R1-Distill-Qwen-32B} & \uppercase\expandafter{\romannumeral1} & 95.54 & 94.60 & 94.75 & 72.92 & 51.67 & 60.10 & 55.56 \\
       & \cellcolor{gray!10}\uppercase\expandafter{\romannumeral2} & \cellcolor{gray!10} 95.73 & \cellcolor{gray!10} 91.47 & \cellcolor{gray!10} 91.25 & \cellcolor{gray!10} 63.33 & \cellcolor{gray!10} 46.04 & \cellcolor{gray!10} 57.83 & \cellcolor{gray!10} 37.63 \\
       & \cellcolor{gray!15}\uppercase\expandafter{\romannumeral3} & \cellcolor{gray!15} 95.44 & \cellcolor{gray!15} 90.72 & \cellcolor{gray!15} 89.00 & \cellcolor{gray!15} 56.25 & \cellcolor{gray!15} 36.39 & \cellcolor{gray!15} 55.22 & \cellcolor{gray!15} 26.64 \\
       & \cellcolor{gray!20}\uppercase\expandafter{\romannumeral4} & \cellcolor{gray!20} 95.43 & \cellcolor{gray!20} 88.69 & \cellcolor{gray!20} 82.50 & \cellcolor{gray!20} 48.54 & \cellcolor{gray!20} 29.79 & \cellcolor{gray!20} 51.14 & \cellcolor{gray!20} 23.30 \\
       & \cellcolor{gray!25}\uppercase\expandafter{\romannumeral5} & \cellcolor{gray!25} 95.39 & \cellcolor{gray!25} 85.00 & \cellcolor{gray!25} 82.22 & \cellcolor{gray!25} 41.92 & \cellcolor{gray!25} 23.08 & \cellcolor{gray!25} 50.71 & \cellcolor{gray!25} 19.28 \\
       
       \multirow{5}{*}{Qwen-QwQ-32B} & \uppercase\expandafter{\romannumeral1} & 95.83 & 96.2 & 95.00 & 78.75 & 69.58 & 63.64 & 62.37 \\
       & \cellcolor{gray!10}\uppercase\expandafter{\romannumeral2} & \cellcolor{gray!10} 96.21 & \cellcolor{gray!10} 93.80 & \cellcolor{gray!10} 94.17 & \cellcolor{gray!10} 66.88 & \cellcolor{gray!10} 56.04 & \cellcolor{gray!10} 60.35 & \cellcolor{gray!10} 45.34 \\
       & \cellcolor{gray!15}\uppercase\expandafter{\romannumeral3} & \cellcolor{gray!15} 95.99 & \cellcolor{gray!15} 92.48 & \cellcolor{gray!15} 87.25 & \cellcolor{gray!15} 59.86 & \cellcolor{gray!15} 45.83 & \cellcolor{gray!15} 59.43 & \cellcolor{gray!15} 32.97 \\
       & \cellcolor{gray!20}\uppercase\expandafter{\romannumeral4} & \cellcolor{gray!20} 95.71 & \cellcolor{gray!20} 92.49 & \cellcolor{gray!20} 79.29 & \cellcolor{gray!20} 49.58 & \cellcolor{gray!20} 37.50 & \cellcolor{gray!20} 61.36 & \cellcolor{gray!20} 26.52\\
       & \cellcolor{gray!25}\uppercase\expandafter{\romannumeral5} & \cellcolor{gray!25} 95.20 & \cellcolor{gray!25} 91.20 & \cellcolor{gray!25} 70.83 & \cellcolor{gray!25} 42.83 & \cellcolor{gray!25} 26.75 & \cellcolor{gray!25} 58.99 & \cellcolor{gray!25} 23.80 \\
       
       \multirow{5}{*}{AReaL-boba-SFT-32B} & \uppercase\expandafter{\romannumeral1} & 93.33 & 95.00 & 97.50 & 77.50 & 60.00 & 63.13 & 60.93 \\
       & \cellcolor{gray!10}\uppercase\expandafter{\romannumeral2} & \cellcolor{gray!10} 93.56 & \cellcolor{gray!10} 93.00 & \cellcolor{gray!10} 95.00 & \cellcolor{gray!10} 63.12 & \cellcolor{gray!10} 50.83 & \cellcolor{gray!10} 54.80 & \cellcolor{gray!10} 45.34 \\
       & \cellcolor{gray!15}\uppercase\expandafter{\romannumeral3} & \cellcolor{gray!15} 95.01 & \cellcolor{gray!15} 89.64 & \cellcolor{gray!15} 83.50 & \cellcolor{gray!15} 55.42 & \cellcolor{gray!15} 37.36 & \cellcolor{gray!15} 54.21 & \cellcolor{gray!15} 36.32 \\
       & \cellcolor{gray!20}\uppercase\expandafter{\romannumeral4} & \cellcolor{gray!20} 95.09 & \cellcolor{gray!20} 87.26 & \cellcolor{gray!20} 75.54 & \cellcolor{gray!20} 37.60 & \cellcolor{gray!20} 28.02 & \cellcolor{gray!20} 47.98 & \cellcolor{gray!20} 27.24 \\
       & \cellcolor{gray!25}\uppercase\expandafter{\romannumeral5} & \cellcolor{gray!25} 95.32 & \cellcolor{gray!25} 85.78 & \cellcolor{gray!25} 61.81 & \cellcolor{gray!25} 27.00 & \cellcolor{gray!25} 18.00 & \cellcolor{gray!25} 45.35 & \cellcolor{gray!25} 18.06 \\

       \multirow{5}{*}{Light-R1-32B-DS} & \uppercase\expandafter{\romannumeral1} & 95.83 & 95.60 & 96.25 & 77.50 & 66.67 & 65.66 & 60.93 \\
       & \cellcolor{gray!10}\uppercase\expandafter{\romannumeral2} & \cellcolor{gray!10} 95.68 & \cellcolor{gray!10} 91.53 & \cellcolor{gray!10} 91.67 & \cellcolor{gray!10} 57.50 & \cellcolor{gray!10} 52.71 & \cellcolor{gray!10} 57.58 & \cellcolor{gray!10} 39.96 \\
       & \cellcolor{gray!15}\uppercase\expandafter{\romannumeral3} & \cellcolor{gray!15} 95.40 & \cellcolor{gray!15} 85.52 & \cellcolor{gray!15} 75.25 & \cellcolor{gray!15} 49.17 & \cellcolor{gray!15} 36.67 & \cellcolor{gray!15} 52.19 & \cellcolor{gray!15} 32.97 \\
       & \cellcolor{gray!20}\uppercase\expandafter{\romannumeral4} & \cellcolor{gray!20} 94.01 & \cellcolor{gray!20} 79.69 & \cellcolor{gray!20} 56.07 & \cellcolor{gray!20} 32.60 & \cellcolor{gray!20} 26.56 & \cellcolor{gray!20} 48.23 & \cellcolor{gray!20} 22.31 \\
       & \cellcolor{gray!25}\uppercase\expandafter{\romannumeral5} & \cellcolor{gray!25} 94.07 & \cellcolor{gray!25} 77.91 & \cellcolor{gray!25} 52.22 & \cellcolor{gray!25} 25.75 & \cellcolor{gray!25} 19.25 & \cellcolor{gray!25} 42.42 & \cellcolor{gray!25} 17.63 \\

       \multirow{5}{*}{S1.1-32B} & \uppercase\expandafter{\romannumeral1} & 89.84 & 90.40 & 90.00 & 55.83 & 45.42 & 61.62 & 25.45 \\
       & \cellcolor{gray!10}\uppercase\expandafter{\romannumeral2} & \cellcolor{gray!10} 70.71 & \cellcolor{gray!10} 65.13 & \cellcolor{gray!10} 48.75 & \cellcolor{gray!10} 36.87 & \cellcolor{gray!10} 29.79 & \cellcolor{gray!10} 55.30 & \cellcolor{gray!10} 25.09\\
       & \cellcolor{gray!15}\uppercase\expandafter{\romannumeral3} & \cellcolor{gray!15} 61.04 & \cellcolor{gray!15} 54.92 & \cellcolor{gray!15} 35.00 & \cellcolor{gray!15} 25.97 & \cellcolor{gray!15} 19.86 & \cellcolor{gray!15} 58.08 & \cellcolor{gray!15} 25.09 \\
       & \cellcolor{gray!20}\uppercase\expandafter{\romannumeral4} & \cellcolor{gray!20} 56.81 & \cellcolor{gray!20} 47.91 & \cellcolor{gray!20} 26.96 & \cellcolor{gray!20} 20.10 & \cellcolor{gray!20} 14.69 & \cellcolor{gray!20} 52.15 & \cellcolor{gray!20} 24.37\\
       & \cellcolor{gray!25}\uppercase\expandafter{\romannumeral5} & \cellcolor{gray!25} 55.82 & \cellcolor{gray!25} 47.44 & \cellcolor{gray!25} 18.33 & \cellcolor{gray!25} 14.75 & \cellcolor{gray!25} 12.17 & \cellcolor{gray!25} 52.63 & \cellcolor{gray!25} 23.30 \\
       
       \multirow{5}{*}{OpenThinker2-32B} & \uppercase\expandafter{\romannumeral1} & 96.44 & 96.20 & 95.00 & 68.33 & 52.50 & 62.12 & 56.27 \\
       & \cellcolor{gray!10}\uppercase\expandafter{\romannumeral2} & \cellcolor{gray!10} 95.48 & \cellcolor{gray!10} 92.80 & \cellcolor{gray!10} 94.58 & \cellcolor{gray!10} 61.46 & \cellcolor{gray!10} 50.62 & \cellcolor{gray!10} 59.09 & \cellcolor{gray!10} 45.34 \\
       & \cellcolor{gray!15}\uppercase\expandafter{\romannumeral3} & \cellcolor{gray!15} 95.27 & \cellcolor{gray!15} 90.48 & \cellcolor{gray!15} 80.25 & \cellcolor{gray!15} 61.94 & \cellcolor{gray!15} 41.53 & \cellcolor{gray!15} 56.73 & \cellcolor{gray!15} 39.19 \\
       & \cellcolor{gray!20}\uppercase\expandafter{\romannumeral4} & \cellcolor{gray!20} 95.19 & \cellcolor{gray!20} 88.63 & \cellcolor{gray!20} 78.04 & \cellcolor{gray!20} 50.21 & \cellcolor{gray!20} 33.96 & \cellcolor{gray!20} 58.46 & \cellcolor{gray!20} 35.39 \\
       & \cellcolor{gray!25}\uppercase\expandafter{\romannumeral5} & \cellcolor{gray!25} 94.74 & \cellcolor{gray!25} 88.49 & \cellcolor{gray!25} 71.11 & \cellcolor{gray!25} 38.42 & \cellcolor{gray!25} 26.67 & \cellcolor{gray!25} 56.87 & \cellcolor{gray!25} 30.18 \\
       
       \multirow{5}{*}{SimpleRL-Zoo-Qwen-32B} & \uppercase\expandafter{\romannumeral1} & 96.06 & 83.20 & 67.50 & 27.20 & 16.67 & 46.46 & 26.52 \\
       & \cellcolor{gray!10}\uppercase\expandafter{\romannumeral2} & \cellcolor{gray!10} 95.40 & \cellcolor{gray!10} 79.80 & \cellcolor{gray!10} 61.67 & \cellcolor{gray!10} 16.67 & \cellcolor{gray!10} 8.12 & \cellcolor{gray!10} 44.19 & \cellcolor{gray!10} 25.81 \\
       & \cellcolor{gray!15}\uppercase\expandafter{\romannumeral3} & \cellcolor{gray!15} 94.35 & \cellcolor{gray!15} 79.52 & \cellcolor{gray!15} 57.00 & \cellcolor{gray!15} 16.67 & \cellcolor{gray!15} 7.92 & \cellcolor{gray!15} 48.48 & \cellcolor{gray!15} 25.33 \\
       & \cellcolor{gray!20}\uppercase\expandafter{\romannumeral4} & \cellcolor{gray!20} 92.79 & \cellcolor{gray!20} 79.37 & \cellcolor{gray!20} 56.07 & \cellcolor{gray!20} 19.17 & \cellcolor{gray!20} 8.75 & \cellcolor{gray!20} 45.58 & \cellcolor{gray!20} 22.49 \\
       & \cellcolor{gray!25}\uppercase\expandafter{\romannumeral5} & \cellcolor{gray!25} 91.43 & \cellcolor{gray!25} 76.89 & \cellcolor{gray!25} 53.33 & \cellcolor{gray!25} 14.67 & \cellcolor{gray!25} 10.67 & \cellcolor{gray!25} 46.57 & \cellcolor{gray!25} 22.15 \\
       
       \multirow{5}{*}{Open-Reasoner-Zero-32B} & \uppercase\expandafter{\romannumeral1} & 95.83 & 92.00 & 83.75 & 46.67 & 36.67 & 60.10 & 35.13 \\
       & \cellcolor{gray!10}\uppercase\expandafter{\romannumeral2} & \cellcolor{gray!10} 93.86 & \cellcolor{gray!10} 87.13 & \cellcolor{gray!10} 80.00 & \cellcolor{gray!10} 38.33 & \cellcolor{gray!10} 26.67 & \cellcolor{gray!10} 51.77 & \cellcolor{gray!10} 14.16 \\
       & \cellcolor{gray!15}\uppercase\expandafter{\romannumeral3} & \cellcolor{gray!15} 92.86 & \cellcolor{gray!15} 83.28 & \cellcolor{gray!15} 68.50 & \cellcolor{gray!15} 35.00 & \cellcolor{gray!15} 26.11 & \cellcolor{gray!15} 50.51 & \cellcolor{gray!15} 13.50 \\
       & \cellcolor{gray!20}\uppercase\expandafter{\romannumeral4} & \cellcolor{gray!20} 91.21 & \cellcolor{gray!20} 80.06 & \cellcolor{gray!20} 68.04 & \cellcolor{gray!20} 27.92 & \cellcolor{gray!20} 22.08 & \cellcolor{gray!20} 47.60 & \cellcolor{gray!20} 11.92 \\
       & \cellcolor{gray!25}\uppercase\expandafter{\romannumeral5} & \cellcolor{gray!25} 89.27 & \cellcolor{gray!25} 81.13 & \cellcolor{gray!25} 63.61 & \cellcolor{gray!25} 25.33 & \cellcolor{gray!25} 19.67 & \cellcolor{gray!25} 48.38 & \cellcolor{gray!25} 12.47\\

       \multirow{5}{*}{Qwen2.5-32B-Inst} & \uppercase\expandafter{\romannumeral1} & 95.53 & 82.20 & 60.00 & 20.00 & 16.67 & 42.93 & 26.88 \\
       & \cellcolor{gray!10}\uppercase\expandafter{\romannumeral2} & \cellcolor{gray!10} 94.36 & \cellcolor{gray!10} 77.07 & \cellcolor{gray!10} 60.00 & \cellcolor{gray!10} 10.00 & \cellcolor{gray!10} 11.67 & \cellcolor{gray!10} 39.39 & \cellcolor{gray!10} 25.45 \\
       & \cellcolor{gray!15}\uppercase\expandafter{\romannumeral3} & \cellcolor{gray!15} 94.17 & \cellcolor{gray!15} 74.76 & \cellcolor{gray!15} 50.00 & \cellcolor{gray!15} 11.11 & \cellcolor{gray!15} 6.39 & \cellcolor{gray!15} 39.06 & \cellcolor{gray!15} 24.25 \\
       & \cellcolor{gray!20}\uppercase\expandafter{\romannumeral4} & \cellcolor{gray!20} 93.37 & \cellcolor{gray!20} 72.20 & \cellcolor{gray!20} 48.04 & \cellcolor{gray!20} 8.33 & \cellcolor{gray!20} 4.17 & \cellcolor{gray!20} 41.79 & \cellcolor{gray!20} 23.30 \\
       & \cellcolor{gray!25}\uppercase\expandafter{\romannumeral5} & \cellcolor{gray!25} 93.19 & \cellcolor{gray!25} 69.51 & \cellcolor{gray!25} 40.83 & \cellcolor{gray!25} 9.00 & \cellcolor{gray!25} 4.67 & \cellcolor{gray!25} 39.90 & \cellcolor{gray!25} 23.01 \\
        \bottomrule[0.8pt]
    \end{tabular}
    }
    \vspace{2mm}
    \caption[]{Detailed accuracy statistics for each stress level of various LRMs.
    }
    \label{tab:main_detail4}
\end{table}

\begin{table}[t!]
\fontsize{9}{10} \selectfont
    \centering
    \def\arraystretch{1,2}
    \resizebox{\textwidth}{!}{%
    \begin{tabular}{ccccccccc}
    \toprule[0.8pt]
       \diagbox{Model}{Bench} & \makecell[c]{Stress\\Degree} & GSM8K & MATH500 & AMC23 & AIME24 & AIME25 & \makecell[c]{GPQA\\Diamond} & \makecell[c]{LiveCode\\Bench} \\
       \hline
       \rowcolor{brown!10}
       \multicolumn{9}{c}{\textit{API-based LRMs}} \\
       \multirow{5}{*}{DeepSeek-R1} & \uppercase\expandafter{\romannumeral1} &  96.20 &  97.00 &  93.75 &  81.66 &  68.75 &  70.20 &  63.44\\
       & \cellcolor{gray!10}\uppercase\expandafter{\romannumeral2} & \cellcolor{gray!10} 96.10 & \cellcolor{gray!10} 94.20 & \cellcolor{gray!10} 88.75 & \cellcolor{gray!10} 66.67 & \cellcolor{gray!10} 55.00 & \cellcolor{gray!10} 70.45 & \cellcolor{gray!10} 50.54\\
       & \cellcolor{gray!15}\uppercase\expandafter{\romannumeral3} & \cellcolor{gray!15} 96.20 & \cellcolor{gray!15} 93.40 & \cellcolor{gray!15} 84.75 & \cellcolor{gray!15} 57.78 & \cellcolor{gray!15} 40.00 & \cellcolor{gray!15} 64.48 & \cellcolor{gray!15} 44.44 \\
       & \cellcolor{gray!20}\uppercase\expandafter{\romannumeral4} & \cellcolor{gray!20} 96.18 & \cellcolor{gray!20} 91.94 & \cellcolor{gray!20} 84.11 & \cellcolor{gray!20} 54.17 & \cellcolor{gray!20} 31.67 & \cellcolor{gray!20} 62.88 & \cellcolor{gray!20} 34.59 \\
       & \cellcolor{gray!25}\uppercase\expandafter{\romannumeral5} & \cellcolor{gray!25} 95.75 & \cellcolor{gray!25} 88.80 & \cellcolor{gray!25} 69.58 & \cellcolor{gray!25} 31.33 & \cellcolor{gray!25} 22.00 & \cellcolor{gray!25} 60.71 & \cellcolor{gray!25} 33.76 \\

       \multirow{5}{*}{O3-Mini} & \uppercase\expandafter{\romannumeral1} &  95.83 &  95.00 &  90.00 &  79.16 &  71.66 &  71.21 &  60.21 \\
       & \cellcolor{gray!10}\uppercase\expandafter{\romannumeral2} & \cellcolor{gray!10} 95.17 & \cellcolor{gray!10} 93.87 & \cellcolor{gray!10} 70.75 & \cellcolor{gray!10} 49.00 & \cellcolor{gray!10} 58.33 & \cellcolor{gray!10} 68.43 & \cellcolor{gray!10} 52.87 \\
       & \cellcolor{gray!15}\uppercase\expandafter{\romannumeral3} & \cellcolor{gray!15} 94.35 & \cellcolor{gray!15} 80.60 & \cellcolor{gray!15} 59.58 & \cellcolor{gray!15} 40.56 & \cellcolor{gray!15} 38.33 & \cellcolor{gray!15} 67.51 & \cellcolor{gray!15} 51.73 \\
       & \cellcolor{gray!20}\uppercase\expandafter{\romannumeral4} & \cellcolor{gray!20} 93.69 & \cellcolor{gray!20} 88.94 & \cellcolor{gray!20} 58.75 & \cellcolor{gray!20} 30.83 & \cellcolor{gray!20} 27.50 & \cellcolor{gray!20} 66.54 & \cellcolor{gray!20} 47.67 \\
       & \cellcolor{gray!25}\uppercase\expandafter{\romannumeral5} & \cellcolor{gray!25} 92.17 & \cellcolor{gray!25} 90.31 & \cellcolor{gray!25} 56.25 & \cellcolor{gray!25} 30.83 & \cellcolor{gray!25} 13.75 & \cellcolor{gray!25} 67.07 & \cellcolor{gray!25} 41.15 \\

        \multirow{5}{*}{O4-Mini} & 
\uppercase\expandafter{\romannumeral1} &93.71 & 90.00 & 96.25 & 73.33 & 80.00 & 76.26 & 70.61 \\
& \cellcolor{gray!10}\uppercase\expandafter{\romannumeral2} &\cellcolor{gray!10} 93.25 & \cellcolor{gray!10} 85.67 & \cellcolor{gray!10} 90.42 & \cellcolor{gray!10} 62.50 & \cellcolor{gray!10} 52.50 & \cellcolor{gray!10} 74.24 & \cellcolor{gray!10} 67.20 \\
& \cellcolor{gray!15}\uppercase\expandafter{\romannumeral3} &\cellcolor{gray!15} 92.96 & \cellcolor{gray!15} 83.40 & \cellcolor{gray!15} 80.75 & \cellcolor{gray!15} 53.89 & \cellcolor{gray!15} 53.33 & \cellcolor{gray!15} 73.40 & \cellcolor{gray!15} 62.96 \\
& \cellcolor{gray!20}\uppercase\expandafter{\romannumeral4} &\cellcolor{gray!20} 92.97 & \cellcolor{gray!20} 82.83 & \cellcolor{gray!20} 84.82 & \cellcolor{gray!20} 52.50 & \cellcolor{gray!20} 49.17 & \cellcolor{gray!20} 71.97 & \cellcolor{gray!20} 61.56 \\
& \cellcolor{gray!25}\uppercase\expandafter{\romannumeral5} &\cellcolor{gray!25} 93.11 & \cellcolor{gray!25} 80.96 & \cellcolor{gray!25} 75.83 & \cellcolor{gray!25} 42.67 & \cellcolor{gray!25} 33.67 & \cellcolor{gray!25} 72.83 & \cellcolor{gray!25} 60.57 \\
        
        \multirow{5}{*}{Gemini-2.5-Flash-Thinking} &
\uppercase\expandafter{\romannumeral1} &89.23 & 97.20 & 97.50 & 76.67 & 71.67 & 78.79 & 61.65 \\
& \cellcolor{gray!10}\uppercase\expandafter{\romannumeral2} &\cellcolor{gray!10} 89.87 & \cellcolor{gray!10} 87.67 & \cellcolor{gray!10} 75.00 & \cellcolor{gray!10} 65.83 & \cellcolor{gray!10} 47.50 & \cellcolor{gray!10} 72.73 & \cellcolor{gray!10} 53.05 \\
& \cellcolor{gray!15}\uppercase\expandafter{\romannumeral3} &\cellcolor{gray!15} 91.76 & \cellcolor{gray!15} 71.96 & \cellcolor{gray!15} 54.00 & \cellcolor{gray!15} 38.89 & \cellcolor{gray!15} 27.22 & \cellcolor{gray!15} 67.34 & \cellcolor{gray!15} 50.42 \\
& \cellcolor{gray!20}\uppercase\expandafter{\romannumeral4} &\cellcolor{gray!20} 92.12 & \cellcolor{gray!20} 69.60 & \cellcolor{gray!20} 41.25 & \cellcolor{gray!20} 22.92 & \cellcolor{gray!20} 18.75 & \cellcolor{gray!20} 68.18 & \cellcolor{gray!20} 47.94 \\
& \cellcolor{gray!25}\uppercase\expandafter{\romannumeral5} &\cellcolor{gray!25} 91.35 & \cellcolor{gray!25} 68.20 & \cellcolor{gray!25} 41.39 & \cellcolor{gray!25} 18.00 & \cellcolor{gray!25} 14.33 & \cellcolor{gray!25} 63.74 & \cellcolor{gray!25} 41.94 \\

        \bottomrule[0.8pt]
    \end{tabular}
    }
    \vspace{2mm}
    \caption[]{Detailed accuracy statistics for each stress level of various LRMs.
    }
    \label{tab:main_detail5}
\end{table}


\section{The Impact of Question Type}

\begin{figure*}[!b]
    \centering
    \begin{subfigure}{\textwidth}
    \includegraphics[width=\linewidth]{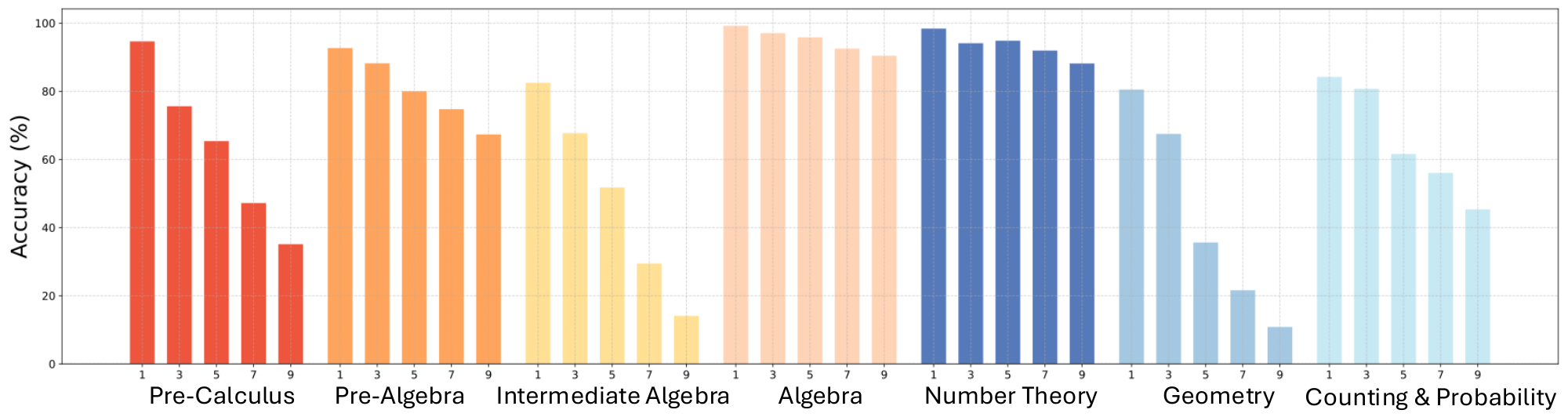}
    \caption{DS-R1-Distill-Qwen-7B}
    \label{fig:bar_math500_category-7b}
    \end{subfigure}\\
    \begin{subfigure}{\textwidth}
    \includegraphics[width=\linewidth]{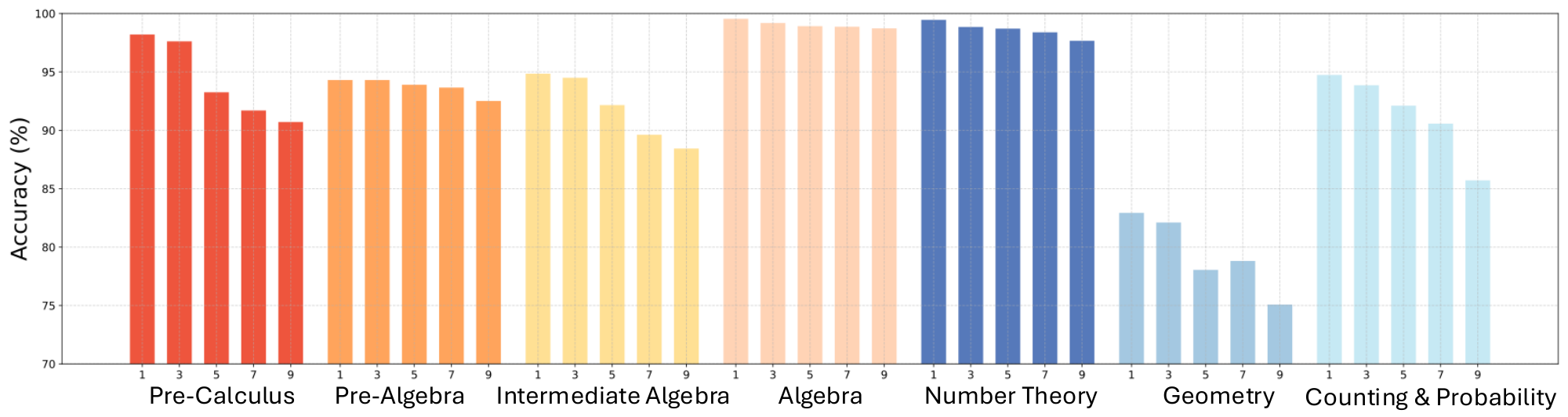}
    \caption{DS-R1-Distill-Qwen-32B}
    \label{fig:bar_math500_category-32b}
    \end{subfigure}\\
    \caption{
    Performance degradation across different question types on MATH500.
    }
    \label{fig:bar_acc_category}
\end{figure*}

To further investigate the factors influencing model performance under stress testing, we analyze the effect of question types. Fig.~\ref{fig:bar_acc_category} shows model performance across different stress levels for various types of math problems. We observe that both R1-7B and R1-32B exhibit similar trends. Specifically, performance degradation under stress is minimal for \textbf{Algebra} and \textbf{Number Theory} problems, indicating that models can handle multiple such questions in a single prompt without significant loss. In contrast, \textbf{Pre-Calculus}, \textbf{Pre-Algebra}, and \textbf{Geometry} questions show substantial drops in accuracy under stress, even for the more capable R1-32B model. Notably, this performance degradation is not always correlated with single-question accuracy. For example, under stress level 1 (\textit{i.e.,} single-question setting), R1-32B achieves 98.21\% accuracy on Pre-Calculus, higher than 94.31\% on Pre-Algebra. However, at stress level 9, its Pre-Calculus accuracy drops to 90.71\%, falling below the 92.51\% accuracy on Pre-Algebra. This indicates that performance degradation under stress is not solely determined by problem difficulty, but also influenced by question type.

\section{Error Type Illustrations}

In this section, we provide a detailed case study of each error type under stress tests in~\crefrange{fig:output_truncatun_case}{fig:reasoning_error_case}. Our findings indicate that \textbf{Output Truncation} is not the only reason for the performance degradation under stress tests, revealing some unexpected misbehaviors like \textbf{Question Omission} and \textbf{Summary Error} that remain hidden in standard evaluation settings.

\FloatBarrier
\begin{figure}[h]
\begin{exmp}{\textit{Example of Output Truncation}}{ot}
\small

\textbf{\textit{Query:}} 

\textit{\textbf{Q1:}} 
Let $O(0,0),A(\tfrac{1}{2},0),$ and $B(0,\tfrac{\sqrt{3}}{2})$ be points in the coordinate plane. Let $\mathcal{F}$ be the family of segments $\overline{PQ}$ of unit length lying in the first quadrant with $P$ on the $x$-axis and $Q$ on the $y$-axis. There is a unique point $C$ on $\overline{AB},$ distinct from $A$ and $B,$ that does not belong to any segment from $\mathcal{F}$ other than $\overline{AB}$. Then $OC^2=\tfrac{p}{q}$, where $p$ and $q$ are relatively prime positive integers. Find $p+q$.

\textit{\textbf{Q2:}}
Find the number of triples of nonnegative integers $(a,b,c)$ satisfying $a+b+c=300$ and $a^2b+a^2c+b^2a+b^2c+c^2a+c^2b = 6,000,000.$

Answer the above questions one by one. Remember to put your final answer within \textbackslash boxed\{\}. \\
 \\
 
\textbf{\textit{Response:}}

Okay, so I have these two problems to solve. Let me start with the first one.

\textit{\textbf{Problem Q1:}}

We have points O(0,0), A(1/2, 0), and B(0, sqrt(3)/2). The family F consists of segments PQ of unit length in the first quadrant, with P on the x-axis and Q on the y-axis. We need to find a unique point C on AB (other than A and B) that doesn't lie on any other segment from F except AB. Then compute OC² and express it as a reduced fraction p/q, and find p+q. Alright, let's break this down. First, let's visualize the points. O is the origin, A is (1/2, 0), which is halfway between (0,0) and (1,0). B is (0, sqrt(3)/2), which is halfway between (0,0) and (0,1) in some sense, but sqrt(3)/2 is approximately 0.866. So, AB is a line segment connecting (1/2, 0) to (0, sqrt(3)/2). Let me compute the equation of line AB. 

$\dots$ \textcolor{red}{(Reasoning content for Q1)} $\dots$

So, to find such a point C, we need to find a point on AB that is not on any other PQ, which are all lines from $(p, 0)$ to $(0, q)$ with $p² + q² = 1$.

\textit{\textbf{Problem Q2:}}

Find the number of triples of nonnegative integers $(a, b, c)$ satisfying $a + b + c =300$ and $a²b + a²c + b²a + b²c + c²a + c²b=6,000,000.$

Okay, so we have two equations:

$\dots$ \textcolor{red}{(Reasoning content for Q2)} $\dots$

Thus, each non-symmetric set is unique, so 6 such sets, each contributing 6 ordered triples\:36.
Plus the symmetric solution\:1.
Total=37.
But, in reality, the equation may have more solutions.
Wait, let me think.
Wait, the equation is:$(a²(b + c) + b²(c + a) + c²(a + b))/2006 = abc.$ So, with \textcolor{red}{(Truncated)}

\end{exmp}
\caption{Example of \textbf{Output Truncation} error on DS-R1-Distill-Qwen-7B.}
\label{fig:output_truncatun_case}
\end{figure}

\begin{figure}[h]
\begin{exmp}{\textit{Example of Endless Repetiion}}{er}
\small

\textbf{\textit{Query:}} 

\textit{\textbf{Q1:}} 
Let $ABCD$ be a tetrahedron such that $AB=CD= \sqrt{41}$, $AC=BD= \sqrt{80}$, and $BC=AD= \sqrt{89}$. There exists a point $I$ inside the tetrahedron such that the distances from $I$ to each of the faces of the tetrahedron are all equal. This distance can be written in the form $\frac{m \sqrt n}{p}$, where $m$, $n$, and $p$ are positive integers, $m$ and $p$ are relatively prime, and $n$ is not divisible by the square of any prime. Find $m+n+p$.

\textit{\textbf{Q2:}}
Let $p$ be the least prime number for which there exists a positive integer $n$ such that $n^{4}+1$ is divisible by $p^{2}$. Find the least positive integer $m$ such that $m^{4}+1$ is divisible by $p^{2}$.

Answer the above questions one by one. Remember to put your final answer within \textbackslash boxed\{\}. \\
\\
 
\textbf{\textit{Response:}}

Problem Q1:
We have a tetrahedron ABCD with some specific edge lengths. The edges AB and CD are both $ \sqrt{41} $, AC and BD are both $ \sqrt{80} $, and BC and AD are both $ \sqrt{89} $. There's a point I inside this tetrahedron such that the distances from I to each face are equal. We need to find this distance, express it in a specific form, and then compute $ m + n + p $ where the distance is $ m\sqrt{n}/p $.  
Hmm, okay. So, first, I remember that in a tetrahedron, the point equidistant from all four faces is called the incenter, similar to the incenter of a triangle. The distance from this point to each face is the inradius.

$\dots$ \textcolor{red}{(Reasoning content for Q1)} $\dots$

Wait, $ 236 $? Hmm, that seems a bit high, but maybe it's correct.  
Wait, let me check the formula for the inradius again.  
Yes, for a tetrahedron, $ r = 3V / A $, where A is the total surface area.  
So, yes, that's correct.  
Alternatively, another way to compute the inradius is to note that in a disphenoid, the inradius can be computed using the formula $ r = (3V)/A_{total} $, which is what I did.  
So, I think $ 236 $ is the answer for Q1.

Problem Q2:
We need to find the least prime number $ p $ such that there exists a positive integer $ n $ where $ p^2 $ divides $ n^4 + 1 $.  
Once we find that prime $ p $, we need to find the least positive integer $ m $ such that $ m^4 + 1 $ is divisible by $ p^2 $.  
So, first, we need to find the smallest prime $ p $ where $ p^2 \mid n^4 + 1 $ for some $ n $.  
To do this, I think we can consider primes $ p $ where $ -1 $ is a quartic residue modulo $ p^2 $.  
Because $ n^4 \equiv -1 \mod p^2 $, which implies that $ n^4 + 1 \equiv 0 \mod p^2 $.  
Therefore, we need to find the smallest prime $ p $ such that the equation $ x^4 \equiv -1 \mod p^2 $ has a solution.  
Alternatively, since if $ x^4 \equiv -1 \mod p^2 $, then $ x^4 \equiv -1 \mod p $, so first, we can check for primes $ p $ where $ -1 $ is a quartic residue modulo $ p $.

$\dots$ \textcolor{red}{(Reasoning content for Q2)} $\dots$

But wait, $ 110 $ is quite large, maybe I can find a smaller one.  
Wait, let me try $ n=55 $.  
$ 55^2 = 3,025 $. $ 3,025 \mod 289 $: $ 289 \cdot 10 = 2,890 $. $ 3,025 - 2,890 = 135 $.  
$ 135^2 = 18,225 $. $ 18,225 \mod 289 $: $ 289 \cdot 62 = 17,918 $. $ 18,225 - 17,918 = 307 $. $ 307 - 289 = 18 $. So, $ 55^4 \equiv 18 \mod 289 $. $ 18 + 1 = 19 \neq 0 $.  
$ n=55 $ doesn't work.  
Wait, $ n=68 $.  
$ 68^2 = 4,624 $. $ 4,624 \mod 289 $: $ 289 \cdot 15 = 4,335 $. $ 4,624 - 4,335 = 289 $. $ 289 \equiv 0 \mod 289 $. So, $ 68^2 \equiv 0 \mod 289 $. So, $ 68^4 \equiv 0 \mod 289 $. Doesn't work.  
$ n=69 $.  
$ 69^2 = 4,761 $. $ 4,761 \mod 289 $: $ 289 \cdot 16 = 4,624 $. $ 4,761 - 4,624 = 137 $.  
$ 137^2 = 18,769 $. $ 18,769 \mod 289 $: $ 289 \cdot 64 = 18,496 $. $ 18,769 - 18,496 = 273 $. $ 273 - 289 = -16 \equiv 273 \mod 289 $.  
So, $ 69^4 \equiv 273 \mod 289 $. $ 273 + 1 = 274 \equiv 274 \neq 0 $.  
$ n=70 $.  
$ 70^2 = 4,900 $. $ 4,900 \mod 289 $: $ 289 \cdot 16 = 4,624 $. $ 4,900 - 4,624 = 276 $.  
$ 276^2 = 76,176 $. $ 76,176 \mod 289 $: $ 289 \cdot 263 = 76,107 $. $ 76,176 - 76,107 = 69 $. So, $ 70^4 \equiv 69 \mod 289 $. $ 69 + 1 = 70 \equiv 70 \neq 0 $.  
$ n=71 $.  
$ 71^2 = 5,041 $. $ 5,041 \mod 289 $: $ 289 \cdot 17 = 4,913 $. $ 5,041 - 4,913 = 128 $.  
$ 128^2 = 16,384 $. $ 16,384 \mod 289 $: $ 289 \cdot 56 = 16,184 $. $ 16,384 - 16,184 = 200 $.  
$ 200 + 1 = 201 \neq 0 $.  
$ n=72 $.  
$ 72^2 = 5,184 $. $ 5,184 \mod 289 $: $ 289 \cdot 17 = 4,913 $. $ 5,184 - 4,913 = 271 $.  
$ 271^2 = 73,441 $. $ 73,441 \mod 289 $: $ 289 \cdot 253 = 73,197 $. $ 73,441 - 73,197 = 244 $. $ 244 - 289 = -45 \equiv 244 \mod 289 $.  
$ 244 + 1 = 245 \neq 0 $.  
$ n=73 $.  
$ 73^2 = 5,329 $. $ 5,329 \mod 289 $: $ 289 \cdot 18 = 5,202 $. $ 5,329 - 5,202 = 127 $.  
$ 127^2 = 16,129 $. $ 16,129 \mod 289 $: $ 289 \cdot 55 = 15,895 \cdot 55=15,895 \cdot 55=15,895 \cdot 55=15,895 \cdot 55=15,895 \cdot 55=15,895 \cdot 55=15,$ \textcolor{red}{(Endless Repeating $289 \cdot 55 = 15,895$)}.

\end{exmp}
\caption{Example of \textbf{Endless Repetition} error on DS-R1-Distill-Qwen-7B.}
\label{fig:endless_repetition_case}
\end{figure}

\begin{figure}[h]
\begin{exmp}{\textit{Example of Question Omission}}{qo}
\small

\textbf{\textit{Query:}} 

Q1: 
Let $N$ be the greatest four-digit positive integer with the property that whenever one of its digits is changed to $1$, the resulting number is divisible by $7$. Let $Q$ and $R$ be the quotient and remainder, respectively, when $N$ is divided by $1000$. Find $Q+R$.

Q2: 
Alice chooses a set $A$ of positive integers. Then Bob lists all finite nonempty sets $B$ of positive integers with the property that the maximum element of $B$ belongs to $A$. Bob's list has 2024 sets. Find the sum of the elements of A.

Answer the above questions one by one. Remember to put your final answer within \textbackslash boxed\{\}. \\
\\

\textbf{\textit{Response:}}

Alright, so I have these two math problems to solve. Let me start with Q1. It says:

``Let N be the greatest four-digit positive integer with the property that whenever one of its digits is changed to 1, the resulting number is divisible by 7. Let Q and R be the quotient and remainder, respectively, when N is divided by 1000. Find Q + R.''

Okay, so N is a four-digit number, and if you change any one of its digits to 1, the new number is divisible by 7. We need to find N, then divide it by 1000 to get Q and R, and then add them together.

First, let me think about how to approach this. Since changing any digit to 1 results in a number divisible by 7, that gives us several conditions. Let's denote N as a four-digit number: N = ABCD, where A, B, C, D are its digits. Then, changing each digit one by one to 1 should result in a number divisible by 7.

So, changing A to 1: 1BCD should be divisible by 7.

Changing B to 1: A1CD should be divisible by 7.

$\dots$ \textcolor{red}{(Reasoning content for Q1)} $\dots$

After checking various possibilities, we find that the valid number $ N $ is 5694.

When $ N = 5694 $ is divided by 100, the quotient $ Q $ is 56 and the remainder $ R $ is 94. Therefore, we compute $ Q + 10R $:

$$
Q + 10R = 56 + 10 \times 94 = 56 + 940 = 996
$$

Thus, the final answer is $\boxed{996}$. \textcolor{red}{(Missing answer for Q2)}

\end{exmp}
\caption{Example of \textbf{Question Omission} error on DS-R1-Distill-Qwen-7B.}
\label{fig:question_omission_case}
\end{figure}

\begin{figure}[h]
\begin{exmp}{\textit{Example of Summary Error}}{se}
\small

\textbf{\textit{Query:}} 

Q1: 
Consider the paths of length $16$ that follow the lines from the lower left corner to the upper right corner on an $8\times 8$ grid. Find the number of such paths that change direction exactly four times, like in the examples shown below. 
\begin{verbatim}
[asy] size(7.5cm); usepackage(\"tikz\");
label(\"\\begin{tikzpicture}[scale=.4] \\draw(0,0)grid(8,8);

\\draw[line width=2,red](0,0)--(2,0)--(2,3)--(5,3)--(5,8)--(8,8);

\\end{tikzpicture}\",origin); 

label(\"\\begin{tikzpicture}[scale=.4]\\draw(0,0)grid(8,8);

\\draw[line width=2,red](0,0)--(0,3)--(3,3)--(3,5)--(8,5)--(8,8);

\\end{tikzpicture}\",E); [/asy]
\end{verbatim}

Q2: 
Rectangles $ABCD$ and $EFGH$ are drawn such that $D,E,C,F$ are collinear. Also, $A,D,H,G$ all lie on a circle. If $BC=16,$ $AB=107,$ $FG=17,$ and $EF=184,$ what is the length of $CE$? 
\begin{verbatim}
[asy] import graph; unitsize(0.1cm);  pair A = (0,0);pair B = (70,0);

pair C = (70,16);pair D = (0,16);pair E = (3,16);pair F = (90,16);pair G = (90,33);

pair H = (3,33); dot(A^^B^^C^^D^^E^^F^^G^^H); label(\"$A$\", A, S);

label(\"$B$\", B, S);label(\"$C$\", C, N);label(\"$D$\", D, N);

label(\"$E$\", E, S);label(\"$F$\", F, S);label(\"$G$\", G, N);

label(\"$H$\", H, N); draw(E--D--A--B--C--E--H--G--F--C); [/asy]
\end{verbatim}

Answer the above questions one by one. Remember to put your final answer within \textbackslash boxed\{\}. \\
\\

\textbf{\textit{Response:}}

Okay, so I have these two questions to solve. Let me start with the first one.

Q1\: Paths on an 8x8 Grid

Alright, so the problem is about finding the number of paths of length 16 on an 8x8 grid that ...

$\dots$ \textcolor{red}{(Reasoning content for Q1)} $\dots$

Similarly, the second example alternates more, but I think the reasoning still holds.
So, yeah, 294 seems correct. Therefore, the answer is 294.

Q2: Rectangles and Circle Problem

Alright, moving on to the second question. It's a geometry problem involving rectangles and a circle.

$\dots$ \textcolor{red}{(Reasoning content for Q2)} $\dots$

Therefore, regardless of the exact value of e, CE is 201.
Wait, that's interesting.
Therefore, $CE=201$.
So, the exact value is 201.
Therefore, the length of CE is 201.
</think>

\textbf{Final Answer}
To solve the first problem, we need to find the number of ways to arrange the letters of the rectangle ROOT such that the letters D and E are adjacent. 
1. The word ROOT has 4 letters with two O's. The distinct letters are R, O, O, T.
2. We treat the pair DE as a single letter, resulting in three units: DE, R, T.
3. These three units can be arranged in 3! = 6 ways.
4. However, DE can also be treated as ED, so we multiply by 2, giving 12 ways.
5. Considering the two O's, the total number of distinct arrangements is 12.
Thus, the number of ways is $\boxed{12}$.

\textcolor{red}{(The answer to Q1 in the final answer is inconsistent with the answer given during the thinking process.)}

For the second problem, we need to find the length of CE in a rectangle ABCD with specific conditions.
$\dots$ \textcolor{red}{(Answer content for Q2)} $\dots$ 

\end{exmp}
\vspace{-2mm}
\caption{Example of \textbf{Summary Error} on DS-R1-Distill-Qwen-7B.}
\label{fig:summary_error_case}
\end{figure}

\begin{figure}[h]
\begin{exmp}{\textit{Example of Reasoning Error}}{re}
\small

\textbf{\textit{Query:}} 

Q1: 
Eight circles of radius $34$ are sequentially tangent, and two of the circles are tangent to $AB$ and $BC$ of triangle $ABC$, respectively. $2024$ circles of radius $1$ can be arranged in the same manner. The inradius of triangle $ABC$ can be expressed as $\frac{m}{n}$, where $m$ and $n$ are relatively prime positive integers. Find $m+n$.
\begin{verbatim} 
[asy] pair A = (2,1); pair B = (0,0); pair C = (3,0); dot(A^^B^^C); 

label(\"$A$\", A, N); label(\"$B$\", B, S); label(\"$C$\", C, S); 

draw(A--B--C--cycle); 

for(real i=0.62; i<2.7; i+=0.29){ draw(circle((i,0.145), 0.145)); } [/asy]
\end{verbatim}

Q2: 
Find the largest possible real part of $(75+117i)z+\frac{96+144i}{z}$ where $z$ is a complex number with $|z|=4$.

Answer the above questions one by one. Remember to put your final answer within \textbackslash boxed\{\}. \\
\\

\textbf{\textit{Response:}}

Okay, so I have these two problems to solve, Q1 and Q2. Let me try to tackle them one by one. I'll start with Q1 because it seems a bit more straightforward, even though I might need to think carefully.

\textbf{Problem Q1:}

Alright, so we have triangle ABC with two circles tangent to sides $AB$ and $BC$ respectively, each of radius 34. Then, there are 2024 circles of radius 1 arranged in the same manner. We need to find the inradius of triangle $ABC$ expressed as a fraction $m/n$, where m and n are coprime, and then find $m + n$.

Hmm, okay. So, I recall that when circles are tangent to sides of a triangle and each other, it often relates to similar triangles or some kind of geometric progression. Maybe it's something to do with the inradius being built up by smaller circles?

$\dots$ \textcolor{red}{(Reasoning content for Q1)} $\dots$

Alternatively, maybe it's the geometric mean, $\sqrt{(34*1)} = \sqrt{34}$, but that's irrational.
But since the problem states that the inradius can be expressed as $m/n$, where m and n are coprime integers, 35 is a reasonable answer.
So, perhaps the answer is 35, so $m + n = 36$.
Therefore, after all this thinking, I think the inradius is 35, so the answer is 36. Final Answer $\boxed{36}$

Now, moving on to problem 2.
We have the function $ f(x) = \frac{75 + 14x - x^2}{1 + 2x} $, and we need to find the maximum value of the expression $ \frac{(75 + 117i - 144j)}{f(i) - 4} $ where $ i $ and $ j $ are complex numbers with $ |i| = |j| = 4 $.
First, I need to understand what is being asked. We have a complex expression, and we need to find its maximum value as $ i $ and $ j $ range over complex numbers with magnitude 4.
So, let me denote $ z = i $, $ w = j $, with $ |z| = |w| = 4 $.
Then, the expression becomes:
$ \frac{75 + 117z - 144w}{f(z) - 4} $

$\dots$ \textcolor{red}{(Reasoning content for Q2)} $\dots$

Wait, maybe the maximum is 15, so the answer is 15, sum is 16.
But to be honest, given that my calculations are not consistent, maybe the answer is 15.
But to end, I think the maximum value is 15, so $15/1$, so $15 +1 =16$. Final Answer Problem 2\: $\boxed{16}$ \textcolor{red}{(Errors occur in the reasoning process of Q1 and Q2.)}

\end{exmp}
\caption{Example of \textbf{Reasoning Error} on DS-R1-Distill-Qwen-7B.}
\label{fig:reasoning_error_case}
\end{figure}

\FloatBarrier


\begin{thebibliography}{63}
\providecommand{\natexlab}[1]{#1}
\providecommand{\url}[1]{\texttt{#1}}
\expandafter\ifx\csname urlstyle\endcsname\relax
  \providecommand{\doi}[1]{doi: #1}\else
  \providecommand{\doi}{doi: \begingroup \urlstyle{rm}\Url}\fi

\bibitem[Aggarwal and Welleck(2025)]{aggarwal2025l1}
Pranjal Aggarwal and Sean Welleck.
\newblock L1: Controlling how long a reasoning model thinks with reinforcement learning.
\newblock \emph{arXiv preprint arXiv:2503.04697}, 2025.

\bibitem[AI-MO()]{aimo_validation_aime}
AI-MO.
\newblock {AIMO Validation AIME Dataset}.

\bibitem[Arora and Zanette(2025)]{arora2025training}
Daman Arora and Andrea Zanette.
\newblock Training language models to reason efficiently.
\newblock \emph{arXiv preprint arXiv:2502.04463}, 2025.

\bibitem[Bercovich et~al.(2025)Bercovich, Levy, Golan, Dabbah, El-Yaniv, Puny, Galil, Moshe, Ronen, Nabwani, Shahaf, Tropp, Karpas, Zilberstein, Zeng, Singhal, Bukharin, Zhang, Konuk, Shen, Mahabaleshwarkar, Kartal, Suhara, Delalleau, Chen, Wang, Mosallanezhad, Renduchintala, Qian, Rekesh, Jia, Majumdar, Noroozi, Ahmad, Narenthiran, Ficek, Samadi, Huang, Jain, Gitman, Moshkov, Du, Toshniwal, Armstrong, Kisacanin, Novikov, Gitman, Bakhturina, Scowcroft, Kamalu, Su, Kong, Kliegl, Karimi, Lin, Satheesh, Parmar, Gundecha, Norick, Jennings, Prabhumoye, Akter, Patwary, Khattar, Narayanan, Waleffe, Zhang, Su, Huang, Kong, Chadha, Jain, Harvey, Segal, Huang, Kashirsky, McQueen, Putterman, Lam, Venkatesan, Wu, Nguyen, Kilaru, Wang, Warno, Somasamudramath, Bhaskar, Dong, Assaf, Mor, Argov, Junkin, Romanenko, Larroy, Katariya, Rovinelli, Balas, Edelman, Bhiwandiwalla, Subramaniam, Ithape, Ramamoorthy, Wu, Velury, Almog, Daw, Fridman, Galinkin, Evans, Luna, Derczynski, Pope, Long, Schneider, Siman, Grzegorzek, Ribalta,
  Katariya, Conway, Saar, Guan, Pawelec, Prayaga, Kuchaiev, Ginsburg, Olabiyi, Briski, Cohen, Catanzaro, Alben, Geifman, Chung, and Alexiuk]{bercovich2025llamanemotronefficientreasoningmodels}
Akhiad Bercovich, Itay Levy, Izik Golan, Mohammad Dabbah, Ran El-Yaniv, Omri Puny, Ido Galil, Zach Moshe, Tomer Ronen, Najeeb Nabwani, Ido Shahaf, Oren Tropp, Ehud Karpas, Ran Zilberstein, Jiaqi Zeng, Soumye Singhal, Alexander Bukharin, Yian Zhang, Tugrul Konuk, Gerald Shen, Ameya~Sunil Mahabaleshwarkar, Bilal Kartal, Yoshi Suhara, Olivier Delalleau, Zijia Chen, Zhilin Wang, David Mosallanezhad, Adi Renduchintala, Haifeng Qian, Dima Rekesh, Fei Jia, Somshubra Majumdar, Vahid Noroozi, Wasi~Uddin Ahmad, Sean Narenthiran, Aleksander Ficek, Mehrzad Samadi, Jocelyn Huang, Siddhartha Jain, Igor Gitman, Ivan Moshkov, Wei Du, Shubham Toshniwal, George Armstrong, Branislav Kisacanin, Matvei Novikov, Daria Gitman, Evelina Bakhturina, Jane~Polak Scowcroft, John Kamalu, Dan Su, Kezhi Kong, Markus Kliegl, Rabeeh Karimi, Ying Lin, Sanjeev Satheesh, Jupinder Parmar, Pritam Gundecha, Brandon Norick, Joseph Jennings, Shrimai Prabhumoye, Syeda~Nahida Akter, Mostofa Patwary, Abhinav Khattar, Deepak Narayanan, Roger Waleffe,
  Jimmy Zhang, Bor-Yiing Su, Guyue Huang, Terry Kong, Parth Chadha, Sahil Jain, Christine Harvey, Elad Segal, Jining Huang, Sergey Kashirsky, Robert McQueen, Izzy Putterman, George Lam, Arun Venkatesan, Sherry Wu, Vinh Nguyen, Manoj Kilaru, Andrew Wang, Anna Warno, Abhilash Somasamudramath, Sandip Bhaskar, Maka Dong, Nave Assaf, Shahar Mor, Omer~Ullman Argov, Scot Junkin, Oleksandr Romanenko, Pedro Larroy, Monika Katariya, Marco Rovinelli, Viji Balas, Nicholas Edelman, Anahita Bhiwandiwalla, Muthu Subramaniam, Smita Ithape, Karthik Ramamoorthy, Yuting Wu, Suguna~Varshini Velury, Omri Almog, Joyjit Daw, Denys Fridman, Erick Galinkin, Michael Evans, Katherine Luna, Leon Derczynski, Nikki Pope, Eileen Long, Seth Schneider, Guillermo Siman, Tomasz Grzegorzek, Pablo Ribalta, Monika Katariya, Joey Conway, Trisha Saar, Ann Guan, Krzysztof Pawelec, Shyamala Prayaga, Oleksii Kuchaiev, Boris Ginsburg, Oluwatobi Olabiyi, Kari Briski, Jonathan Cohen, Bryan Catanzaro, Jonah Alben, Yonatan Geifman, Eric Chung, and Chris
  Alexiuk.
\newblock Llama-nemotron: Efficient reasoning models, 2025.
\newblock URL \url{https://arxiv.org/abs/2505.00949}.

\bibitem[Chen et~al.(2024)Chen, Xu, Liang, He, Pang, Yu, Song, Liu, Zhou, Zhang, et~al.]{chen2024not}
Xingyu Chen, Jiahao Xu, Tian Liang, Zhiwei He, Jianhui Pang, Dian Yu, Linfeng Song, Qiuzhi Liu, Mengfei Zhou, Zhuosheng Zhang, et~al.
\newblock Do not think that much for 2+ 3=? on the overthinking of o1-like llms.
\newblock \emph{arXiv preprint arXiv:2412.21187}, 2024.

\bibitem[Cheng et~al.(2023)Cheng, Kasai, and Yu]{cheng2023batch}
Zhoujun Cheng, Jungo Kasai, and Tao Yu.
\newblock Batch prompting: Efficient inference with large language model apis.
\newblock In \emph{Proceedings of the 2023 Conference on Empirical Methods in Natural Language Processing: Industry Track}, pages 792--810, 2023.

\bibitem[Cobbe et~al.(2021{\natexlab{a}})Cobbe, Kosaraju, Bavarian, Chen, Jun, Kaiser, Plappert, Tworek, Hilton, Nakano, Hesse, and Schulman]{cobbe2021trainingverifierssolvemath}
Karl Cobbe, Vineet Kosaraju, Mohammad Bavarian, Mark Chen, Heewoo Jun, Lukasz Kaiser, Matthias Plappert, Jerry Tworek, Jacob Hilton, Reiichiro Nakano, Christopher Hesse, and John Schulman.
\newblock Training verifiers to solve math word problems, 2021{\natexlab{a}}.
\newblock URL \url{https://arxiv.org/abs/2110.14168}.

\bibitem[Cobbe et~al.(2021{\natexlab{b}})Cobbe, Kosaraju, Bavarian, Chen, Jun, Kaiser, Plappert, Tworek, Hilton, Nakano, et~al.]{cobbe2021gsm8k}
Karl Cobbe, Vineet Kosaraju, Mohammad Bavarian, Mark Chen, Heewoo Jun, Lukasz Kaiser, Matthias Plappert, Jerry Tworek, Jacob Hilton, Reiichiro Nakano, et~al.
\newblock Training verifiers to solve math word problems.
\newblock \emph{arXiv preprint arXiv:2110.14168}, 2021{\natexlab{b}}.

\bibitem[Cui et~al.(2025)Cui, Yuan, Wang, Wang, Li, He, Fan, Yu, Xu, Chen, et~al.]{cui2025process}
Ganqu Cui, Lifan Yuan, Zefan Wang, Hanbin Wang, Wendi Li, Bingxiang He, Yuchen Fan, Tianyu Yu, Qixin Xu, Weize Chen, et~al.
\newblock Process reinforcement through implicit rewards.
\newblock \emph{arXiv preprint arXiv:2502.01456}, 2025.

\bibitem[Face(2025)]{openr1}
Hugging Face.
\newblock Open r1: A fully open reproduction of deepseek-r1, January 2025.
\newblock URL \url{https://github.com/huggingface/open-r1}.

\bibitem[Gandhi et~al.(2025)Gandhi, Chakravarthy, Singh, Lile, and Goodman]{gandhi2025cognitive}
Kanishk Gandhi, Ayush Chakravarthy, Anikait Singh, Nathan Lile, and Noah~D Goodman.
\newblock Cognitive behaviors that enable self-improving reasoners, or, four habits of highly effective stars.
\newblock \emph{arXiv preprint arXiv:2503.01307}, 2025.

\bibitem[Gao et~al.(2024)Gao, Song, Yang, Cai, Miao, Dong, Li, Ma, Chen, Xu, et~al.]{gao2024omni}
Bofei Gao, Feifan Song, Zhe Yang, Zefan Cai, Yibo Miao, Qingxiu Dong, Lei Li, Chenghao Ma, Liang Chen, Runxin Xu, et~al.
\newblock Omni-math: A universal olympiad level mathematic benchmark for large language models.
\newblock \emph{arXiv preprint arXiv:2410.07985}, 2024.

\bibitem[Guo et~al.(2025)Guo, Yang, Zhang, Song, Zhang, Xu, Zhu, Ma, Wang, Bi, et~al.]{guo2025deepseek}
Daya Guo, Dejian Yang, Haowei Zhang, Junxiao Song, Ruoyu Zhang, Runxin Xu, Qihao Zhu, Shirong Ma, Peiyi Wang, Xiao Bi, et~al.
\newblock Deepseek-r1: Incentivizing reasoning capability in llms via reinforcement learning.
\newblock \emph{arXiv preprint arXiv:2501.12948}, 2025.

\bibitem[Han et~al.(2024)Han, Wang, Fang, Zhao, Ma, and Chen]{han2024token}
Tingxu Han, Zhenting Wang, Chunrong Fang, Shiyu Zhao, Shiqing Ma, and Zhenyu Chen.
\newblock Token-budget-aware llm reasoning.
\newblock \emph{arXiv preprint arXiv:2412.18547}, 2024.

\bibitem[He et~al.(2024)He, Luo, Bai, Hu, Thai, Shen, Hu, Han, Huang, Zhang, et~al.]{he2024olympiadbench}
Chaoqun He, Renjie Luo, Yuzhuo Bai, Shengding Hu, Zhen Thai, Junhao Shen, Jinyi Hu, Xu~Han, Yujie Huang, Yuxiang Zhang, et~al.
\newblock Olympiadbench: A challenging benchmark for promoting agi with olympiad-level bilingual multimodal scientific problems.
\newblock In \emph{Proceedings of the 62nd Annual Meeting of the Association for Computational Linguistics (Volume 1: Long Papers)}, pages 3828--3850, 2024.

\bibitem[Hendrycks et~al.(2021{\natexlab{a}})Hendrycks, Burns, Kadavath, Arora, Basart, Tang, Song, and Steinhardt]{hendrycks2021math}
Dan Hendrycks, Collin Burns, Saurav Kadavath, Akul Arora, Steven Basart, Eric Tang, Dawn Song, and Jacob Steinhardt.
\newblock Measuring mathematical problem solving with the math dataset.
\newblock In J.~Vanschoren and S.~Yeung, editors, \emph{Proceedings of the Neural Information Processing Systems Track on Datasets and Benchmarks}, volume~1, 2021{\natexlab{a}}.
\newblock URL \url{https://datasets-benchmarks-proceedings.neurips.cc/paper_files/paper/2021/file/be83ab3ecd0db773eb2dc1b0a17836a1-Paper-round2.pdf}.

\bibitem[Hendrycks et~al.(2021{\natexlab{b}})Hendrycks, Burns, Kadavath, Arora, Basart, Tang, Song, and Steinhardt]{hendrycks2021measuringmathematicalproblemsolving}
Dan Hendrycks, Collin Burns, Saurav Kadavath, Akul Arora, Steven Basart, Eric Tang, Dawn Song, and Jacob Steinhardt.
\newblock Measuring mathematical problem solving with the math dataset, 2021{\natexlab{b}}.
\newblock URL \url{https://arxiv.org/abs/2103.03874}.

\bibitem[Hochlehnert et~al.(2025)Hochlehnert, Bhatnagar, Udandarao, Albanie, Prabhu, and Bethge]{hochlehnert2025sober}
Andreas Hochlehnert, Hardik Bhatnagar, Vishaal Udandarao, Samuel Albanie, Ameya Prabhu, and Matthias Bethge.
\newblock A sober look at progress in language model reasoning: Pitfalls and paths to reproducibility.
\newblock \emph{arXiv preprint arXiv:2504.07086}, 2025.

\bibitem[Hou et~al.(2024)Hou, Luo, Ruan, Wang, Ge, Chen, and Chen]{hou2024compound}
Yutao Hou, Yajing Luo, Zhiwen Ruan, Hongru Wang, Weifeng Ge, Yun Chen, and Guanhua Chen.
\newblock Compound-qa: A benchmark for evaluating llms on compound questions.
\newblock \emph{arXiv preprint arXiv:2411.10163}, 2024.

\bibitem[Hu et~al.(2025)Hu, Zhang, Han, Jiang, Zhang, and Shum]{hu2025open}
Jingcheng Hu, Yinmin Zhang, Qi~Han, Daxin Jiang, Xiangyu Zhang, and Heung-Yeung Shum.
\newblock Open-reasoner-zero: An open source approach to scaling up reinforcement learning on the base model.
\newblock \emph{arXiv preprint arXiv:2503.24290}, 2025.

\bibitem[Hui et~al.(2024)Hui, Yang, Cui, Yang, Liu, Zhang, Liu, Zhang, Yu, Dang, Yang, Men, Huang, Ren, Ren, Zhou, and Lin]{hui2024qwen25codertechnicalreport}
Binyuan Hui, Jian Yang, Zeyu Cui, Jiaxi Yang, Dayiheng Liu, Lei Zhang, Tianyu Liu, Jiajun Zhang, Bowen Yu, Kai Dang, An~Yang, Rui Men, Fei Huang, Xingzhang Ren, Xuancheng Ren, Jingren Zhou, and Junyang Lin.
\newblock Qwen2.5-coder technical report, 2024.
\newblock URL \url{https://arxiv.org/abs/2409.12186}.

\bibitem[Jain et~al.(2024)Jain, Han, Gu, Li, Yan, Zhang, Wang, Solar-Lezama, Sen, and Stoica]{jain2024livecodebench}
Naman Jain, King Han, Alex Gu, Wen-Ding Li, Fanjia Yan, Tianjun Zhang, Sida Wang, Armando Solar-Lezama, Koushik Sen, and Ion Stoica.
\newblock Livecodebench: Holistic and contamination free evaluation of large language models for code.
\newblock \emph{arXiv preprint arXiv:2403.07974}, 2024.

\bibitem[Jimenez et~al.()Jimenez, Yang, Wettig, Yao, Pei, Press, and Narasimhan]{swebench}
Carlos~E Jimenez, John Yang, Alexander Wettig, Shunyu Yao, Kexin Pei, Ofir Press, and Karthik~R Narasimhan.
\newblock Swe-bench: Can language models resolve real-world github issues?
\newblock In \emph{The Twelfth International Conference on Learning Representations}.

\bibitem[Li et~al.(2024)Li, Chen, Wang, Zhao, Liang, Hou, Liu, and Zhou]{li2024mosaic}
Ming Li, Pei Chen, Chenguang Wang, Hongyu Zhao, Yijun Liang, Yupeng Hou, Fuxiao Liu, and Tianyi Zhou.
\newblock Mosaic-it: Free compositional data augmentation improves instruction tuning.
\newblock \emph{arXiv preprint arXiv:2405.13326}, 2024.

\bibitem[Li et~al.(2025)Li, Pei, Sun, Lin, Ming, Gao, Wu, He, and Wu]{li2025cipherbank}
Yu~Li, Qizhi Pei, Mengyuan Sun, Honglin Lin, Chenlin Ming, Xin Gao, Jiang Wu, Conghui He, and Lijun Wu.
\newblock Cipherbank: Exploring the boundary of llm reasoning capabilities through cryptography challenges.
\newblock \emph{arXiv preprint arXiv:2504.19093}, 2025.

\bibitem[Lin et~al.(2025)Lin, Pan, Li, Pei, Gao, Cai, He, and Wu]{lin2025metaladder}
Honglin Lin, Zhuoshi Pan, Yu~Li, Qizhi Pei, Xin Gao, Mengzhang Cai, Conghui He, and Lijun Wu.
\newblock Metaladder: Ascending mathematical solution quality via analogical-problem reasoning transfer.
\newblock \emph{arXiv preprint arXiv:2503.14891}, 2025.

\bibitem[Lin(2025)]{aime2025}
Yen-Ting Lin.
\newblock Aime 2025 dataset, 2025.
\newblock URL \url{https://huggingface.co/datasets/yentinglin/aime_2025}.
\newblock Accessed: 2025-03-29.

\bibitem[Liu et~al.(2024)Liu, Lin, Hewitt, Paranjape, Bevilacqua, Petroni, and Liang]{liu2024lost}
Nelson~F Liu, Kevin Lin, John Hewitt, Ashwin Paranjape, Michele Bevilacqua, Fabio Petroni, and Percy Liang.
\newblock Lost in the middle: How language models use long contexts.
\newblock \emph{Transactions of the Association for Computational Linguistics}, 12, 2024.

\bibitem[Luo et~al.(2025{\natexlab{a}})Luo, Shen, He, Wang, Liu, Li, Tan, Cao, and Tao]{luo2025o1}
Haotian Luo, Li~Shen, Haiying He, Yibo Wang, Shiwei Liu, Wei Li, Naiqiang Tan, Xiaochun Cao, and Dacheng Tao.
\newblock O1-pruner: Length-harmonizing fine-tuning for o1-like reasoning pruning.
\newblock \emph{arXiv preprint arXiv:2501.12570}, 2025{\natexlab{a}}.

\bibitem[Luo et~al.(2025{\natexlab{b}})Luo, Tan, Wong, Shi, Tang, Roongta, Cai, Luo, Li, Popa, and Stoica]{deepscaler2025}
Michael Luo, Sijun Tan, Justin Wong, Xiaoxiang Shi, William~Y. Tang, Manan Roongta, Colin Cai, Jeffrey Luo, Li~Erran Li, Raluca~Ada Popa, and Ion Stoica.
\newblock Deepscaler: Surpassing o1-preview with a 1.5b model by scaling rl, 2025{\natexlab{b}}.
\newblock Notion Blog.

\bibitem[Mei et~al.(2025)Mei, Fu, Li, Wang, Zhang, and Wu]{mei2025real}
Zhiyu Mei, Wei Fu, Kaiwei Li, Guangju Wang, Huanchen Zhang, and Yi~Wu.
\newblock Real: Efficient rlhf training of large language models with parameter reallocation.
\newblock In \emph{Proceedings of the Eighth Conference on Machine Learning and Systems, MLSys 2025, Santa Clara, CA, USA, May 12-15, 2025}. mlsys.org, 2025.

\bibitem[Muennighoff et~al.(2025)Muennighoff, Yang, Shi, Li, Fei-Fei, Hajishirzi, Zettlemoyer, Liang, Candès, and Hashimoto]{muennighoff2025s1simpletesttimescaling}
Niklas Muennighoff, Zitong Yang, Weijia Shi, Xiang~Lisa Li, Li~Fei-Fei, Hannaneh Hajishirzi, Luke Zettlemoyer, Percy Liang, Emmanuel Candès, and Tatsunori Hashimoto.
\newblock s1: Simple test-time scaling, 2025.
\newblock URL \url{https://arxiv.org/abs/2501.19393}.

\bibitem[Nayab et~al.(2024)Nayab, Rossolini, Simoni, Saracino, Buttazzo, Manes, and Giacomelli]{nayab2024concise}
Sania Nayab, Giulio Rossolini, Marco Simoni, Andrea Saracino, Giorgio Buttazzo, Nicolamaria Manes, and Fabrizio Giacomelli.
\newblock Concise thoughts: Impact of output length on llm reasoning and cost.
\newblock \emph{arXiv preprint arXiv:2407.19825}, 2024.

\bibitem[OpenAI(2025)]{openai2025o3mini}
OpenAI.
\newblock Openai o3 and o4-mini system card, Apr 2025.
\newblock URL \url{https://openai.com/index/o3-o4-mini-system-card/}.

\bibitem[Pan et~al.(2025)Pan, Li, Lin, Pei, Tang, Wu, Ming, Zhao, He, and Wu]{pan2025lemma}
Zhuoshi Pan, Yu~Li, Honglin Lin, Qizhi Pei, Zinan Tang, Wei Wu, Chenlin Ming, H~Vicky Zhao, Conghui He, and Lijun Wu.
\newblock Lemma: Learning from errors for mathematical advancement in llms.
\newblock \emph{arXiv preprint arXiv:2503.17439}, 2025.

\bibitem[Pei et~al.(2025)Pei, Wu, Pan, Li, Lin, Ming, Gao, He, and Yan]{pei2025mathfusion}
Qizhi Pei, Lijun Wu, Zhuoshi Pan, Yu~Li, Honglin Lin, Chenlin Ming, Xin Gao, Conghui He, and Rui Yan.
\newblock Mathfusion: Enhancing mathematic problem-solving of llm through instruction fusion.
\newblock \emph{arXiv preprint arXiv:2503.16212}, 2025.

\bibitem[Quan et~al.(2025)Quan, Yang, Yu, Zheng, Liu, Yang, Ren, Gao, Miao, Feng, et~al.]{quan2025codeelo}
Shanghaoran Quan, Jiaxi Yang, Bowen Yu, Bo~Zheng, Dayiheng Liu, An~Yang, Xuancheng Ren, Bofei Gao, Yibo Miao, Yunlong Feng, et~al.
\newblock Codeelo: Benchmarking competition-level code generation of llms with human-comparable elo ratings.
\newblock \emph{arXiv preprint arXiv:2501.01257}, 2025.

\bibitem[Rein et~al.(2024)Rein, Hou, Stickland, Petty, Pang, Dirani, Michael, and Bowman]{rein2024gpqa}
David Rein, Betty~Li Hou, Asa~Cooper Stickland, Jackson Petty, Richard~Yuanzhe Pang, Julien Dirani, Julian Michael, and Samuel~R Bowman.
\newblock Gpqa: A graduate-level google-proof q\&a benchmark.
\newblock In \emph{First Conference on Language Modeling}, 2024.

\bibitem[Rozière et~al.(2024)Rozière, Gehring, Gloeckle, Sootla, Gat, Tan, Adi, Liu, Sauvestre, Remez, Rapin, Kozhevnikov, Evtimov, Bitton, Bhatt, Ferrer, Grattafiori, Xiong, Défossez, Copet, Azhar, Touvron, Martin, Usunier, Scialom, and Synnaeve]{rozière2024codellamaopenfoundation}
Baptiste Rozière, Jonas Gehring, Fabian Gloeckle, Sten Sootla, Itai Gat, Xiaoqing~Ellen Tan, Yossi Adi, Jingyu Liu, Romain Sauvestre, Tal Remez, Jérémy Rapin, Artyom Kozhevnikov, Ivan Evtimov, Joanna Bitton, Manish Bhatt, Cristian~Canton Ferrer, Aaron Grattafiori, Wenhan Xiong, Alexandre Défossez, Jade Copet, Faisal Azhar, Hugo Touvron, Louis Martin, Nicolas Usunier, Thomas Scialom, and Gabriel Synnaeve.
\newblock Code llama: Open foundation models for code, 2024.
\newblock URL \url{https://arxiv.org/abs/2308.12950}.

\bibitem[Semwal et~al.(2018)Semwal, Yenigalla, Mathur, and Nair]{semwal2018practitioners}
Tushar Semwal, Promod Yenigalla, Gaurav Mathur, and Shivashankar~B Nair.
\newblock A practitioners' guide to transfer learning for text classification using convolutional neural networks.
\newblock In \emph{Proceedings of the 2018 SIAM international conference on data mining}, pages 513--521. SIAM, 2018.

\bibitem[Shah et~al.(2025)Shah, Rushton, Singla, Parmar, Smith, Vanjani, Vaswani, Chaluvaraju, Hojel, Ma, et~al.]{shah2025rethinking}
Darsh~J Shah, Peter Rushton, Somanshu Singla, Mohit Parmar, Kurt Smith, Yash Vanjani, Ashish Vaswani, Adarsh Chaluvaraju, Andrew Hojel, Andrew Ma, et~al.
\newblock Rethinking reflection in pre-training.
\newblock \emph{arXiv preprint arXiv:2504.04022}, 2025.

\bibitem[Shao et~al.(2024)Shao, Wang, Zhu, Xu, Song, Bi, Zhang, Zhang, Li, Wu, and Guo]{shao2024deepseekmathpushinglimitsmathematical}
Zhihong Shao, Peiyi Wang, Qihao Zhu, Runxin Xu, Junxiao Song, Xiao Bi, Haowei Zhang, Mingchuan Zhang, Y.~K. Li, Y.~Wu, and Daya Guo.
\newblock Deepseekmath: Pushing the limits of mathematical reasoning in open language models, 2024.
\newblock URL \url{https://arxiv.org/abs/2402.03300}.

\bibitem[Shorten et~al.(2024)Shorten, Pierse, Smith, Cardenas, Sharma, Trengrove, and van Luijt]{shorten2024structuredrag}
Connor Shorten, Charles Pierse, Thomas~Benjamin Smith, Erika Cardenas, Akanksha Sharma, John Trengrove, and Bob van Luijt.
\newblock Structuredrag: Json response formatting with large language models.
\newblock \emph{arXiv preprint arXiv:2408.11061}, 2024.

\bibitem[Son et~al.(2024)Son, Baek, Nam, Jeong, and Kim]{son2024multi}
Guijin Son, SangWon Baek, Sangdae Nam, Ilgyun Jeong, and Seungone Kim.
\newblock Multi-task inference: Can large language models follow multiple instructions at once?
\newblock In \emph{Proceedings of the 62nd Annual Meeting of the Association for Computational Linguistics (Volume 1: Long Papers)}, pages 5606--5627, 2024.

\bibitem[Sui et~al.(2025)Sui, Chuang, Wang, Zhang, Zhang, Yuan, Liu, Wen, Zhong, Chen, et~al.]{sui2025stop}
Yang Sui, Yu-Neng Chuang, Guanchu Wang, Jiamu Zhang, Tianyi Zhang, Jiayi Yuan, Hongyi Liu, Andrew Wen, Shaochen Zhong, Hanjie Chen, et~al.
\newblock Stop overthinking: A survey on efficient reasoning for large language models.
\newblock \emph{arXiv preprint arXiv:2503.16419}, 2025.

\bibitem[Talmor et~al.(2019)Talmor, Herzig, Lourie, and Berant]{talmor2019commonsenseqa}
Alon Talmor, Jonathan Herzig, Nicholas Lourie, and Jonathan Berant.
\newblock Commonsenseqa: A question answering challenge targeting commonsense knowledge.
\newblock In \emph{Proceedings of the 2019 Conference of the North}, page 4149. Association for Computational Linguistics, 2019.

\bibitem[Tam et~al.(2024)Tam, Wu, Tsai, Lin, Lee, and Chen]{tam2024let}
Zhi~Rui Tam, Cheng-Kuang Wu, Yi-Lin Tsai, Chieh-Yen Lin, Hung-yi Lee, and Yun-Nung Chen.
\newblock Let me speak freely? a study on the impact of format restrictions on performance of large language models.
\newblock \emph{arXiv preprint arXiv:2408.02442}, 2024.

\bibitem[Team et~al.(2023)Team, Anil, Borgeaud, Alayrac, Yu, Soricut, Schalkwyk, Dai, Hauth, Millican, et~al.]{team2023gemini}
Gemini Team, Rohan Anil, Sebastian Borgeaud, Jean-Baptiste Alayrac, Jiahui Yu, Radu Soricut, Johan Schalkwyk, Andrew~M Dai, Anja Hauth, Katie Millican, et~al.
\newblock Gemini: a family of highly capable multimodal models.
\newblock \emph{arXiv preprint arXiv:2312.11805}, 2023.

\bibitem[Team et~al.(2025{\natexlab{a}})Team, Kamath, Ferret, Pathak, Vieillard, Merhej, Perrin, Matejovicova, Ram{\'e}, Rivi{\`e}re, et~al.]{team2025gemma}
Gemma Team, Aishwarya Kamath, Johan Ferret, Shreya Pathak, Nino Vieillard, Ramona Merhej, Sarah Perrin, Tatiana Matejovicova, Alexandre Ram{\'e}, Morgane Rivi{\`e}re, et~al.
\newblock Gemma 3 technical report.
\newblock \emph{arXiv preprint arXiv:2503.19786}, 2025{\natexlab{a}}.

\bibitem[Team et~al.(2025{\natexlab{b}})Team, Du, Gao, Xing, Jiang, Chen, Li, Xiao, Du, Liao, et~al.]{team2025kimi}
Kimi Team, Angang Du, Bofei Gao, Bowei Xing, Changjiu Jiang, Cheng Chen, Cheng Li, Chenjun Xiao, Chenzhuang Du, Chonghua Liao, et~al.
\newblock Kimi k1. 5: Scaling reinforcement learning with llms.
\newblock \emph{arXiv preprint arXiv:2501.12599}, 2025{\natexlab{b}}.

\bibitem[Team(2025{\natexlab{a}})]{OpenThoughts2025}
Open~Thoughts Team.
\newblock {Open Thoughts}, January 2025{\natexlab{a}}.
\newblock URL \url{https://www.open-thoughts.ai/}.

\bibitem[Team(2025{\natexlab{b}})]{qwq32b}
Qwen Team.
\newblock Qwq-32b: Embracing the power of reinforcement learning, March 2025{\natexlab{b}}.
\newblock URL \url{https://qwenlm.github.io/blog/qwq-32b/}.

\bibitem[Wang et~al.(2025)Wang, Liu, Xu, Liang, Chen, He, Song, Yu, Li, Zhang, et~al.]{wang2025thoughts}
Yue Wang, Qiuzhi Liu, Jiahao Xu, Tian Liang, Xingyu Chen, Zhiwei He, Linfeng Song, Dian Yu, Juntao Li, Zhuosheng Zhang, et~al.
\newblock Thoughts are all over the place: On the underthinking of o1-like llms.
\newblock \emph{arXiv preprint arXiv:2501.18585}, 2025.

\bibitem[Wang et~al.(2024)Wang, Kodner, and Rambow]{wang2024evaluating}
Zhengxiang Wang, Jordan Kodner, and Owen Rambow.
\newblock Evaluating llms with multiple problems at once: A new paradigm for probing llm capabilities.
\newblock \emph{arXiv e-prints}, pages arXiv--2406, 2024.

\bibitem[Wen et~al.(2025)Wen, Cai, Xiao, He, An, Duan, Du, Liu, Tang, Lv, Zou, Deng, Jia, and Zhang]{wen2025light}
Liang Wen, Yunke Cai, Fenrui Xiao, Xin He, Qi~An, Zhenyu Duan, Yimin Du, Junchen Liu, Lifu Tang, Xiaowei Lv, Haosheng Zou, Yongchao Deng, Shousheng Jia, and Xiangzheng Zhang.
\newblock Light-r1: Curriculum sft, dpo and rl for long cot from scratch and beyond.
\newblock \emph{arXiv preprint arXiv:2503.10460}, 2025.

\bibitem[Yang et~al.(2024{\natexlab{a}})Yang, Yang, Zhang, Hui, Zheng, Yu, Li, Liu, Huang, Wei, et~al.]{yang2024qwen25}
An~Yang, Baosong Yang, Beichen Zhang, Binyuan Hui, Bo~Zheng, Bowen Yu, Chengyuan Li, Dayiheng Liu, Fei Huang, Haoran Wei, et~al.
\newblock Qwen2.5 technical report.
\newblock \emph{arXiv preprint arXiv:2412.15115}, 2024{\natexlab{a}}.

\bibitem[Yang et~al.(2024{\natexlab{b}})Yang, Zhang, Hui, Gao, Yu, Li, Liu, Tu, Zhou, Lin, Lu, Xue, Lin, Liu, Ren, and Zhang]{yang2024qwen25mathtechnicalreportmathematical}
An~Yang, Beichen Zhang, Binyuan Hui, Bofei Gao, Bowen Yu, Chengpeng Li, Dayiheng Liu, Jianhong Tu, Jingren Zhou, Junyang Lin, Keming Lu, Mingfeng Xue, Runji Lin, Tianyu Liu, Xingzhang Ren, and Zhenru Zhang.
\newblock Qwen2.5-math technical report: Toward mathematical expert model via self-improvement, 2024{\natexlab{b}}.
\newblock URL \url{https://arxiv.org/abs/2409.12122}.

\bibitem[Yang et~al.(2024{\natexlab{c}})Yang, Zhang, Hui, Gao, Yu, Li, Liu, Tu, Zhou, Lin, et~al.]{yang2024qwen2}
An~Yang, Beichen Zhang, Binyuan Hui, Bofei Gao, Bowen Yu, Chengpeng Li, Dayiheng Liu, Jianhong Tu, Jingren Zhou, Junyang Lin, et~al.
\newblock Qwen2.5-math technical report: Toward mathematical expert model via self-improvement.
\newblock \emph{arXiv preprint arXiv:2409.12122}, 2024{\natexlab{c}}.

\bibitem[Ye et~al.(2025)Ye, Xiao, Mi, and Liu]{ye2025aimepreview}
Yixin Ye, Yang Xiao, Tiantian Mi, and Pengfei Liu.
\newblock Aime-preview: A rigorous and immediate evaluation framework for advanced mathematical reasoning.
\newblock \url{https://github.com/GAIR-NLP/AIME-Preview}, 2025.
\newblock GitHub repository.

\bibitem[Yu et~al.()Yu, Jiang, Luo, Wu, Lin, Li, Yang, Huang, and Qiu]{yu2024mitigate}
Yijiong Yu, Huiqiang Jiang, Xufang Luo, Qianhui Wu, Chin-Yew Lin, Dongsheng Li, Yuqing Yang, Yongfeng Huang, and Lili Qiu.
\newblock Mitigate position bias in large language models via scaling a single dimension.
\newblock In \emph{First Workshop on Long-Context Foundation Models@ ICML 2024}.

\bibitem[Yue et~al.(2025)Yue, Chen, Lu, Zhao, Wang, Song, and Huang]{yue2025does}
Yang Yue, Zhiqi Chen, Rui Lu, Andrew Zhao, Zhaokai Wang, Shiji Song, and Gao Huang.
\newblock Does reinforcement learning really incentivize reasoning capacity in llms beyond the base model?
\newblock \emph{arXiv preprint arXiv:2504.13837}, 2025.

\bibitem[Zeng et~al.(2025)Zeng, Huang, Liu, Liu, He, Ma, and He]{zeng2025simplerlzooinvestigatingtamingzero}
Weihao Zeng, Yuzhen Huang, Qian Liu, Wei Liu, Keqing He, Zejun Ma, and Junxian He.
\newblock Simplerl-zoo: Investigating and taming zero reinforcement learning for open base models in the wild, 2025.
\newblock URL \url{https://arxiv.org/abs/2503.18892}.

\bibitem[Zhao et~al.(2024)Zhao, Yin, Zeng, Wang, Shi, Lyu, Wang, Luo, and Zhang]{zhao2024marcoo1openreasoningmodels}
Yu~Zhao, Huifeng Yin, Bo~Zeng, Hao Wang, Tianqi Shi, Chenyang Lyu, Longyue Wang, Weihua Luo, and Kaifu Zhang.
\newblock Marco-o1: Towards open reasoning models for open-ended solutions, 2024.
\newblock URL \url{https://arxiv.org/abs/2411.14405}.

\end{thebibliography}
\end{document}